\documentclass[10pt,twocolumn,letterpaper]{article}

\usepackage{cvpr}
\usepackage{times}
\usepackage{epsfig}
\usepackage{graphicx}
\usepackage{amsmath}
\usepackage{amssymb}
\usepackage{color}
\usepackage{subcaption}

\newcommand{\Eqnref}[1]{Equation~(\ref{#1})}
\newcommand{\figref}[1]{Fig.~\ref{#1}}
\newcommand{\tabref}[1]{Table~\ref{#1}}
\newcommand{\eqnref}[1]{Eq.~(\ref{#1})}
\newcommand{\secref}[1]{Sec.~\ref{#1}}

\newcommand\inv[1]{#1\raisebox{1.15ex}{$\scriptscriptstyle-\!1$}}
\newcommand{\RN}[1]{%
	\textup{\uppercase\expandafter{\romannumeral#1}}%
}

\newcommand{\argmin}{\operatornamewithlimits{argmin}}
\renewcommand{\paragraph}[1]{\noindent{\bf #1.}}

\usepackage[breaklinks=true,bookmarks=false]{hyperref}

\cvprfinalcopy 

\def\cvprPaperID{2443} 

\ifcvprfinal\fi
\begin{document}

\title{Robust Depth Estimation from Auto Bracketed Images}

\author{Sunghoon Im,~Hae-Gon Jeon,~In So Kweon\\
	Korea Advanced Institute of Science and Technology (KAIST), Republic of Korea\\
	{\tt\small \{dlarl8927, earboll, iskweon77\}@kaist.ac.kr}	
}

\maketitle

\begin{abstract}
	As demand for advanced photographic applications on hand-held devices grows, these electronics require the capture of high quality depth. However, under low-light conditions, most devices still suffer from low imaging quality and inaccurate depth acquisition. To address the problem, we present a robust depth estimation method from a short burst shot with varied intensity (\ie, Auto Bracketing) or strong noise (\ie, High ISO). We introduce a geometric transformation between flow and depth tailored for burst images, enabling our learning-based multi-view stereo matching to be performed effectively. We then describe our depth estimation pipeline that incorporates the geometric transformation into our residual-flow network. It allows our framework to produce an accurate depth map even with a bracketed image sequence. We demonstrate that our method outperforms state-of-the-art methods for various datasets captured by a smartphone and a DSLR camera. Moreover, we show that the estimated depth is applicable for image quality enhancement and photographic editing. 
\end{abstract}

\section{Introduction}

\begin{figure}[t]
	\centering
	\begin{tabular}{c@{\hspace{1mm}}c@{\hspace{1mm}}}
		\subcaptionbox{\label{input} Input: Exposure bracketed images}{\includegraphics[height=0.255\linewidth]{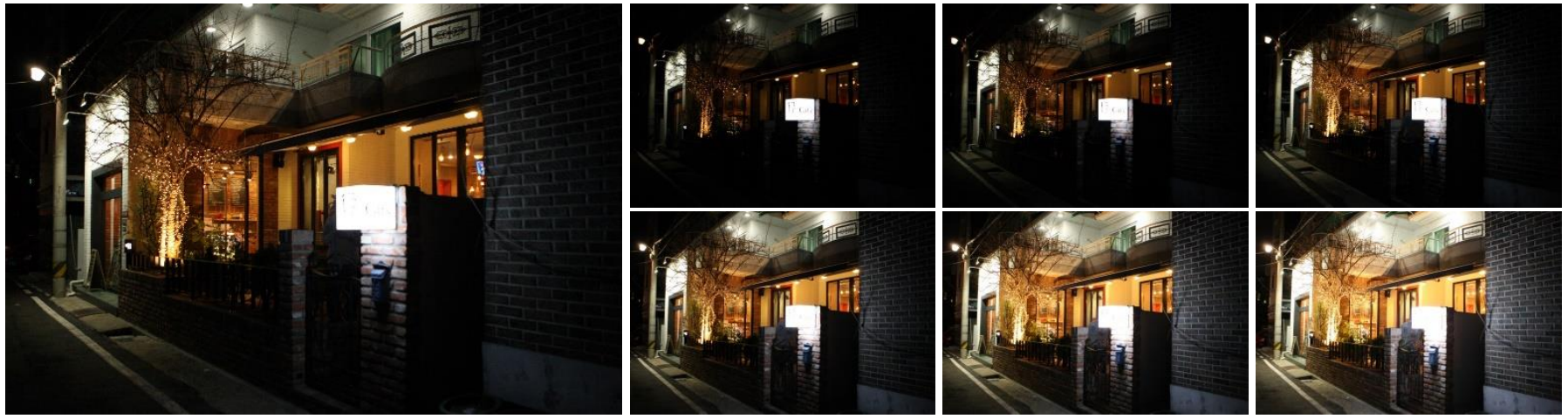}}\\
		\subcaptionbox{\label{campose} Camera pose \& 3D points}{\includegraphics[height=0.31\linewidth]{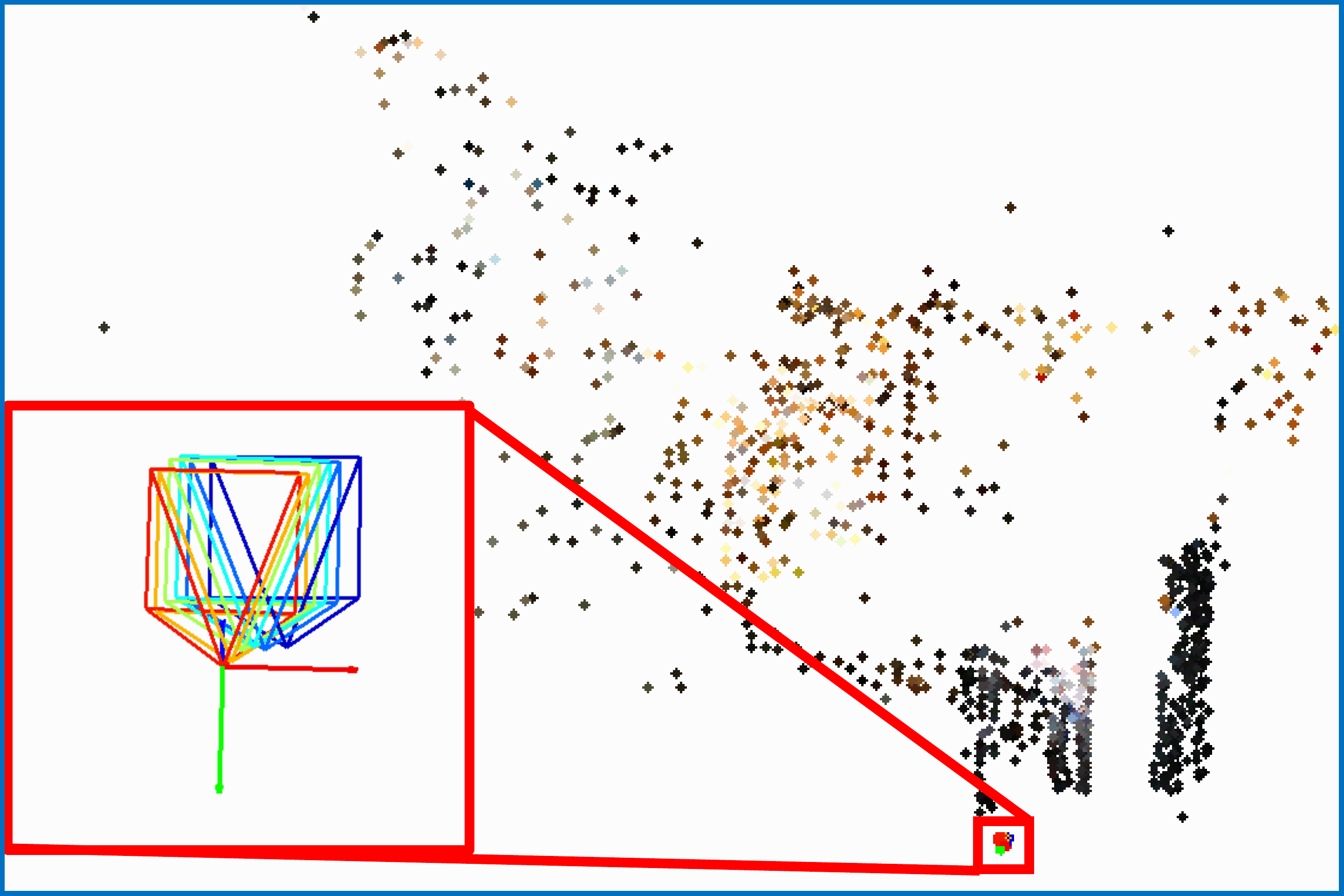}}				
		\subcaptionbox{\label{dcnn} Depth map result}{\includegraphics[height=0.31\linewidth]{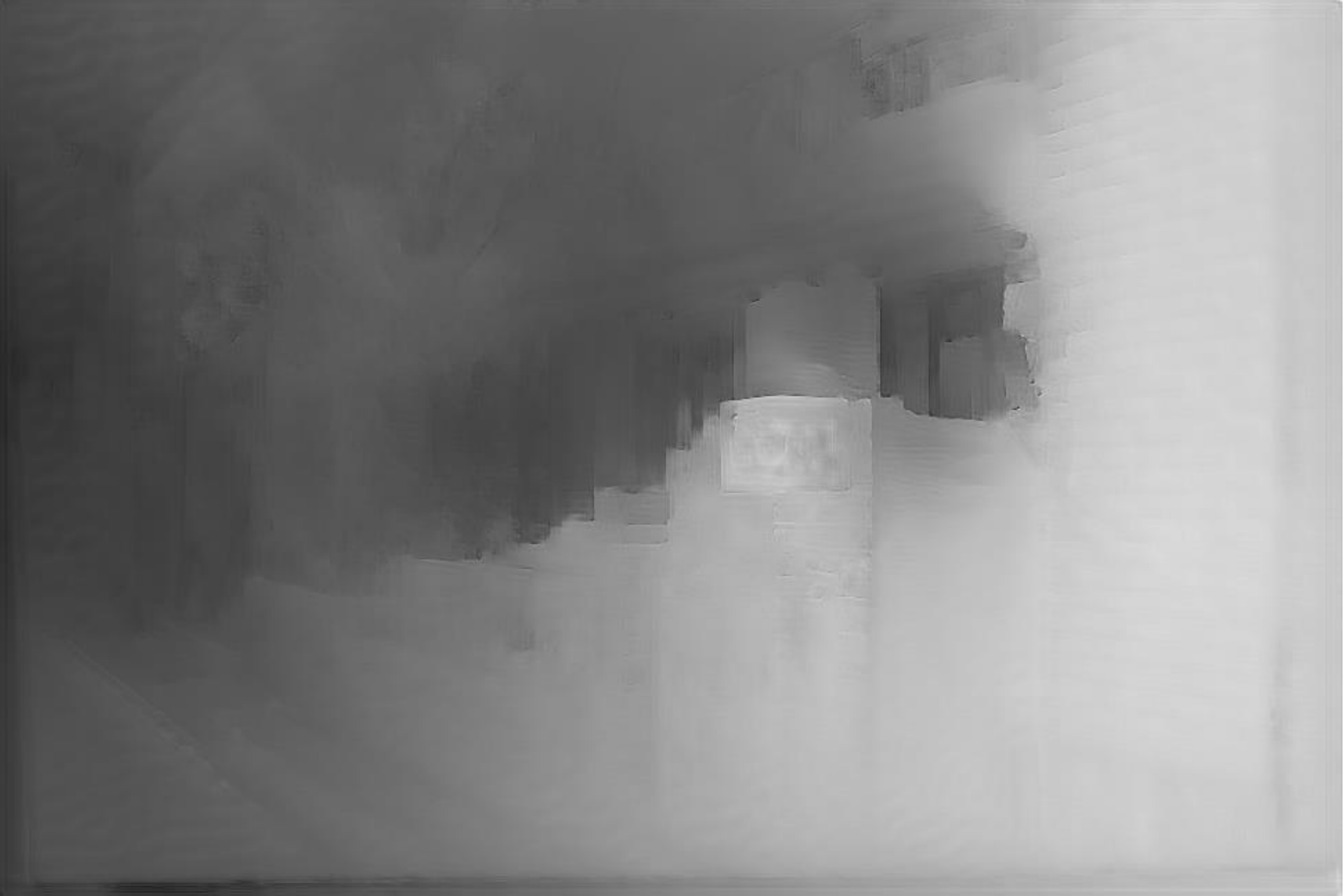}}\\
		\subcaptionbox{\label{efusion} Exposure fusion}{\includegraphics[height=0.31\linewidth]{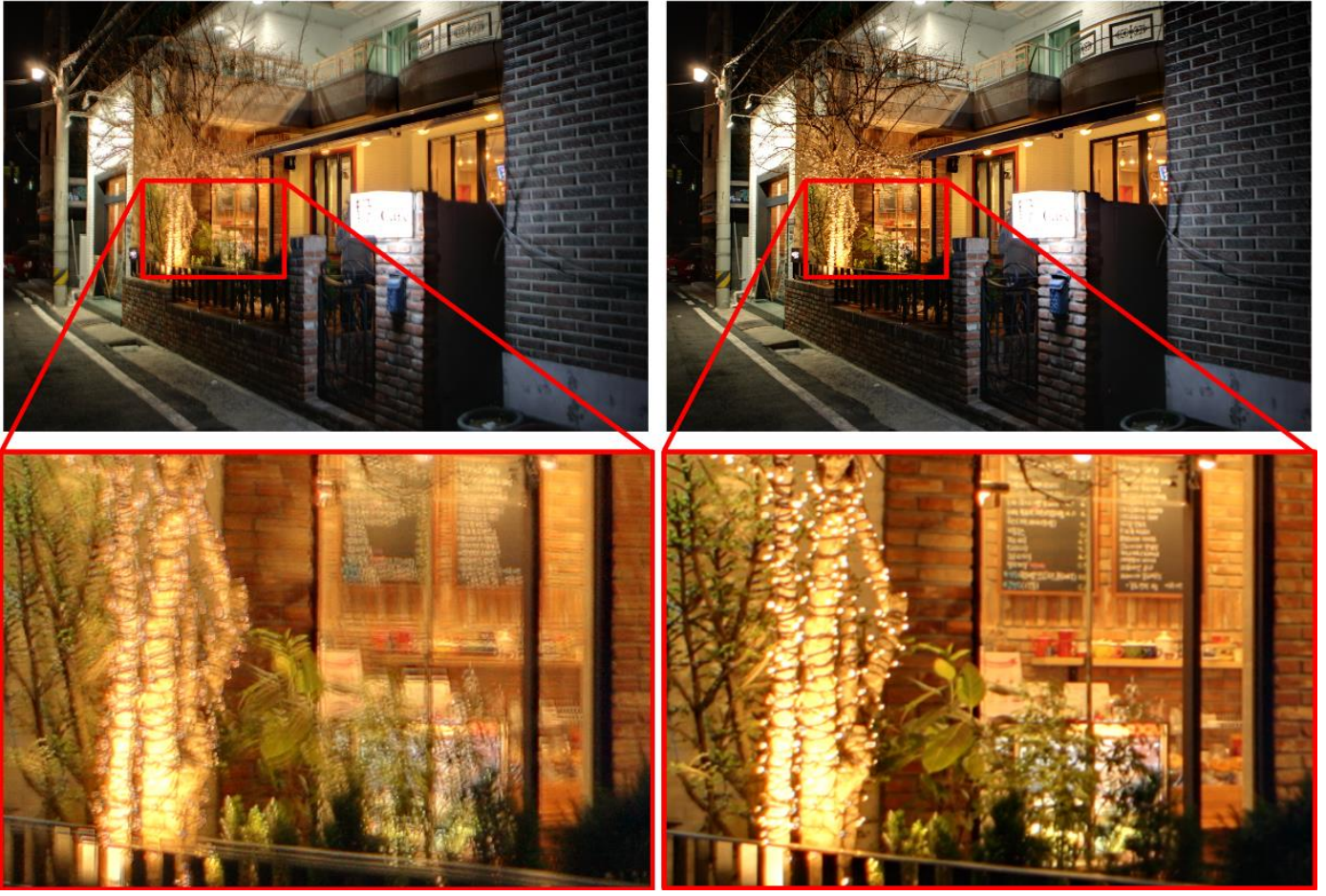}}	
		\subcaptionbox{\label{efusion2} Synthetic refocusing}{\includegraphics[height=0.31\linewidth]{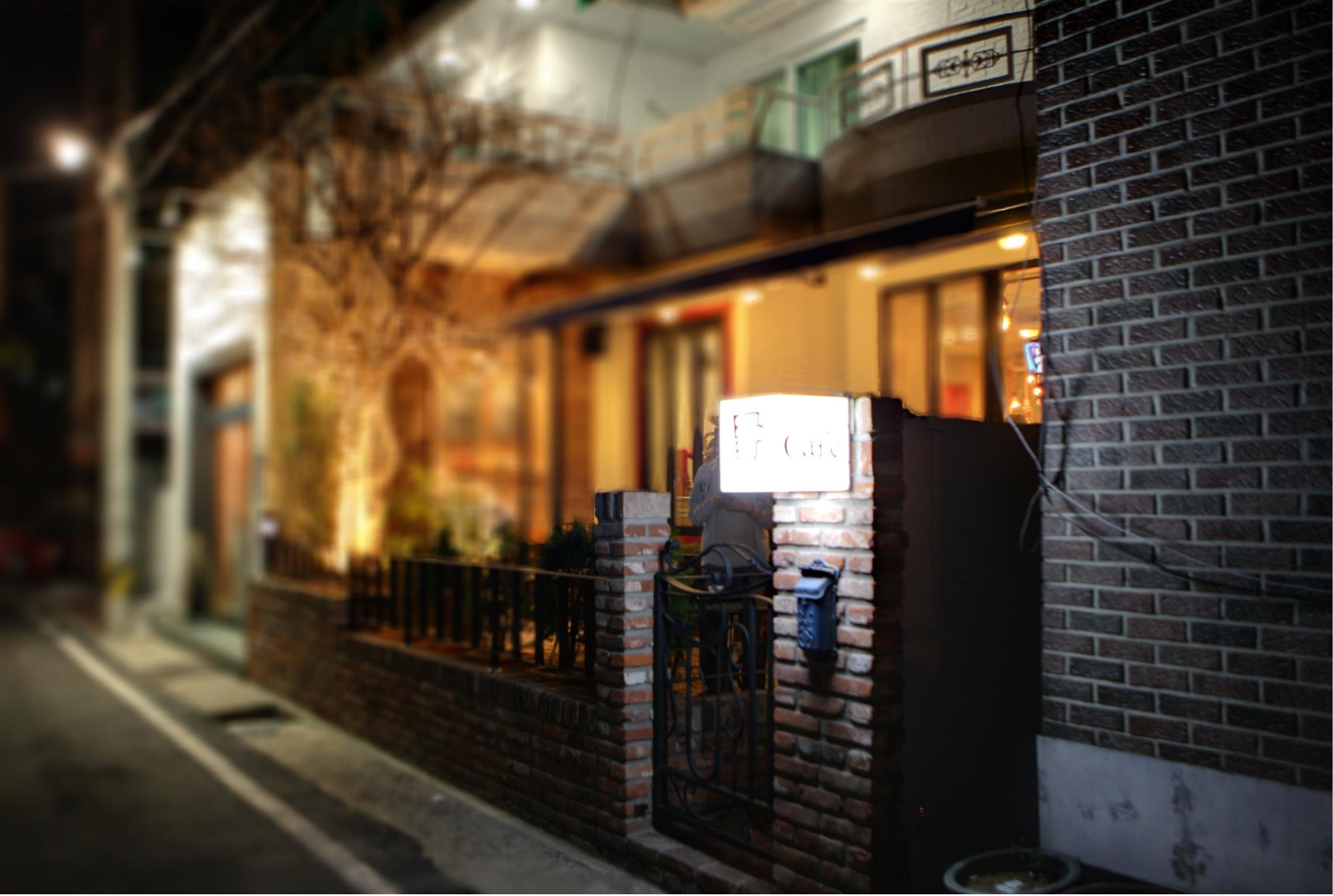}}
	\end{tabular}
	\caption{Given exposure bracketed images (a), we estimate camera pose (b) and depth map (c). Our results are applicable to image quality enhancement (d) and depth-aware application (e). We compare exposure fusion results from input images (\textit{L}) and aligned images using our depth (\textit{R}).}
	\label{fig:overview}
\end{figure}	

Many photographers want to capture high-quality images of indoor or night scenes that are insufficiently exposed to light. To do so, they increase exposure time or ISO, but these adjustments can cause other imaging problems, such as motion blur or noise amplification.
In an effort to mitigate the physical limitation of camera hardware, several image processing methods have been widely employed, such as single image denoising~\cite{dabov2007image,buades2005non} or edge preserving filtering~\cite{tomasi1998bilateral,he2013guided}. However, those approaches often degrade the sharpness of the image or produce cartoonish and surreal results.

The function to take several successive shots with different camera settings called Auto-bracketing (\eg, Exposure, ISO or Flash) or in a very short time called Burst shot has become ubiquitous in most hand-held imaging devices. These photographic techniques for gathering more light have recently attracted interest from the field of computational photography~\cite{Liu14,Hasinoff16}. 
Assuming that the images are all well-aligned, they are commonly utilized for various image restorations (\eg, Denoising or HDR). 
However, multiple image alignment is an important issue, since motion inevitably occurs when users press the camera shutter.

In this work, we determined that the inevitable motion, considered as a nuisance in previous burst photography~\cite{Liu14,Hasinoff16}, can be used as an important clue to estimate the depth.
The estimated depth can be utilized for precise image alignment, which rely highly on discretized homography or optical flow in the conventional methods. Moreover, we show that our depth is useful for various depth-aware applications such as photographic editing or augmented/virtual reality. 

Previous studies~\cite{Yu14,Im15,Ha16} on the so-called \textit{depth from small motion} (DfSM), have introduced a depth estimation approach based on multiple images with narrow baselines.
However, conventional DfSM works have serious limitations, such as (1) noise-sensitive characteristics and (2) high computational complexity, so the estimated depth is not reasonably applicable to hand-held devices as a means of improving image quality.
Instead, we propose a learning-based multi-view stereo method combined with the geometric inference.

\textit{Deep neural network} (DNN) has recently been shown to perform well for various computer vision tasks, such as image classification, detection and optical flow estimation. 
In particular, learning-based optical flow estimation methods~\cite{dosovitskiy2015flownet,bai2016exploiting,ilg2016flownet} outperform conventional optimization-based approaches in accuracy and speed~\cite{menze2015object}. 
However, modern geometric interpretations~\cite{hartley2003multiple}  have great advantages in terms of generality and accuracy over the learning-based approaches, \eg,~pose estimation~\cite{Ummenhofer17}, and re-localization~\cite{kendall2015posenet}. 
To accomplish a robust and fast approach, we complement DNN and modern geometric understandings, and take full advantage of each study.

We first compute a scene geometry including sparse 3D points and camera poses~\secref{sec:sfsm}, from an input image sequence captured in a burst mode or bracketing mode as shown in~\figref{input}. 
An output of the scene geometry is then used to obtain a dense depth map by integrating with DNN in~\secref{sec:DMVS}.
Moreover, we show that the estimated depth map can be utilized for precise image alignment in~\secref{sec:alignment}.
We have carefully evaluated our algorithm using a variety of synthetic and real-world datasets. 
In the presence of moderate or strong noise and varied intensity in input sequence, our results show considerable improvement over state-of-the-art DfSM methods.

Of course, there are simplified versions of exposure fusion which utilize an image sequence with the same exposure times as the input~\cite{Hasinoff16,Hasinoff10}.
Having the same exposures significantly reduces the difficulty in aligning images captured at different times. 
However, we observe that the burst images can be suffered from many under- or over-exposed pixels when the appropriate exposure time is not determined.
The bracketed images are necessary to truly achieve HDR or exposure fusion. We show that our depth can minimize these performance degradations by aligning the images with varying exposures and it is useful for a variety of applications.

\section{Our Approach}
\label{sec:approach}
This section describes an effective pipelines for depth and pose estimation method from short burst shots, especially exposure bracketed sequences. First, we introduce robust pose estimation method for intensity variation, which is slightly modified from the Structure from Small Motion (SfSM) method~\cite{Im15} in~\secref{sec:sfsm}. 
Second, we propose a robust depth estimation method tailored for short burst shots even with varied intensity or noise in~\secref{sec:DMVS}. Lastly, we briefly describe the image alignment method based on our depth and pose information in~\secref{sec:alignment}.

\subsection{Structure-from-Small-Motion (SfSM)}
\label{sec:sfsm}

We first extract features from the reference image using Harris corner detection~\cite{harris1988combined}, and track the features in a pair of histogram-equalized images using the Kanade-Lucas-Tomasi (KLT) tracker~\cite{tomasi1991detection}. 
Before the feature extraction, we perform histogram equalization on all images. 
Although most commercial cameras have non-linear response functions, this process alleviates the color inconsistency problem in the feature matching step.
The equalized images are only used in the feature extraction. 

Given the pre-calibrated intrinsic parameters $\mathbf{K}$, we estimated the relative camera poses and sparse 3D points by solving the following equation: 
\begin{gather}
\begin{split}
\label{eq:bundlecost}
\argmin_{\mathbf{r}, \mathbf{t}, \mathbf{X}} \sum^{n}_{i=1}\sum^{m}_{j=1}\parallel\mathbf{u}_{ij}-\big \langle\mathbf{K}\big[\mathbf{R}_{i}|\mathbf{t}_i\big]\begin{bmatrix} \mathbf{X}_j\\ 1 \end{bmatrix}\big \rangle\parallel_{2},\\
\mathbf{K}=\begin{bmatrix}
f_x& 0 & c_x \\
0& f_y & c_y \\
0& 0 & 1  
\end{bmatrix}, 
\big[\mathbf{R}_i|\mathbf{t}_i\big] = \begin{bmatrix}
1& $-$r_i^z & r_i^y &\hspace{-1.1mm}\vline\hspace{-1.1mm}& t_i^x \\
r_i^z& 1 & $-$r_i^x &\hspace{-1.1mm}\vline\hspace{-1.1mm}& t_i^y\\
$-$r_i^y&r_i^x & 1 &\hspace{-1.1mm}\vline\hspace{-1.1mm}& t_i^z  
\end{bmatrix},
\end{split}
\end{gather}  
where $\mathbf{r}=[r^x, r^y, r^z]^\intercal$, $\mathbf{t}=[t^x, t^y, t^z]^\intercal$ and $\mathbf{X}_j$ are the rotation, translation components and 3D world coordinates of features. 
$\mathbf{u}=[u,v,1]^\intercal=\mathbf{K}\mathbf{x}$ and $\mathbf{x}=[x,y,1]^\intercal$ are the image coordinates and normalized image coordinates, respectively. 
$n$ and $m$ are the number of images and features. $\parallel\cdot\parallel_{2}$ is the L2 norm and $\langle\cdot\rangle$ is the projection function, that is $\big \langle [a,b,c]^\intercal \big \rangle=[a/c, b/c]^\intercal$.

We initialize all camera components $\mathbf{r}, \mathbf{t}$ to zero and the 3D points $\mathbf{X}$ by multiplying the normalized image coordinates $\mathbf{x}$ by a random depth value.
We use the Levenberg-Marquardt (LM) optimization~\cite{more1978levenberg} to solve~\Eqnref{eq:bundlecost}. 

\begin{figure*}[!htb]
	\centering
	\begin{tabular}{c@{\hspace{1mm}}c@{\hspace{1mm}}c@{\hspace{1mm}}c@{\hspace{1mm}}c@{\hspace{1mm}}}
		{\includegraphics[height=0.128\linewidth]{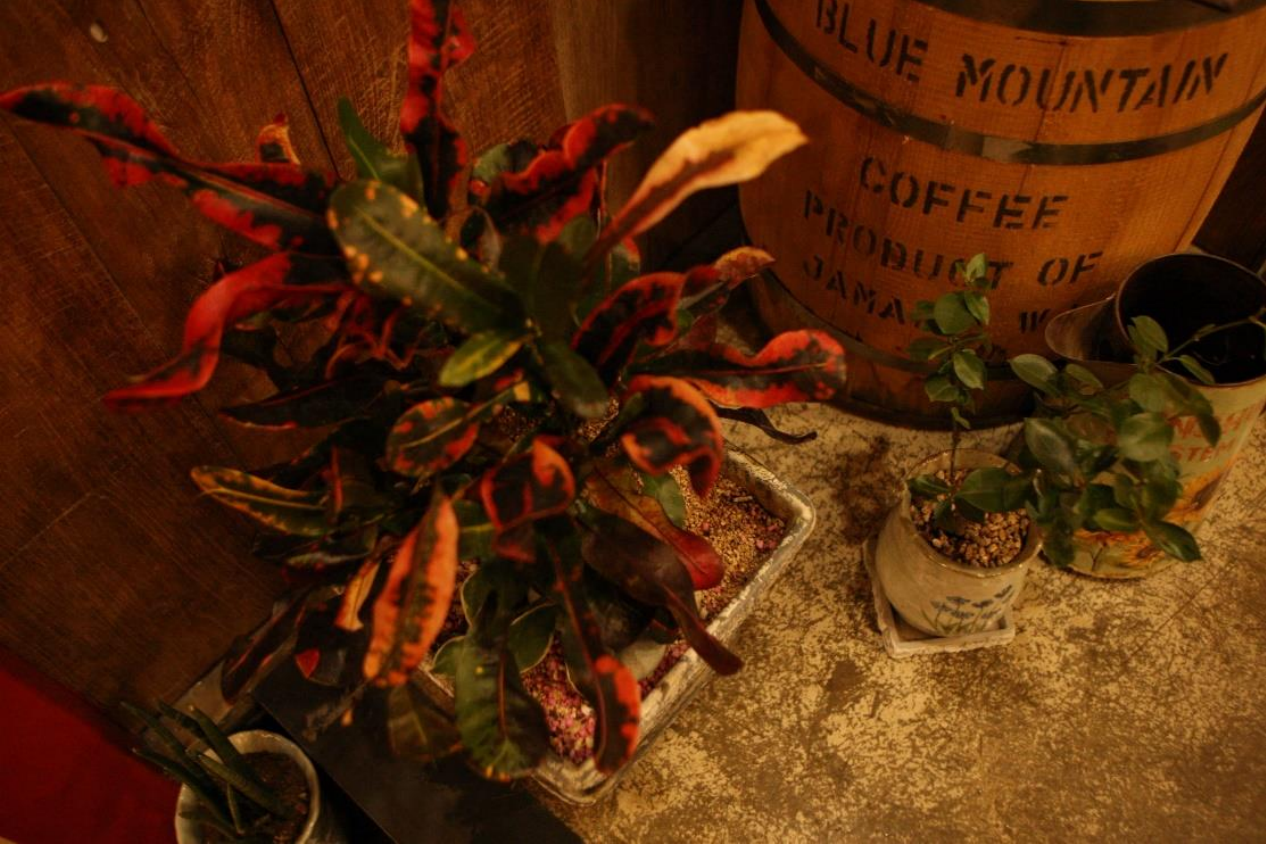}}
		{\includegraphics[height=0.128\linewidth]{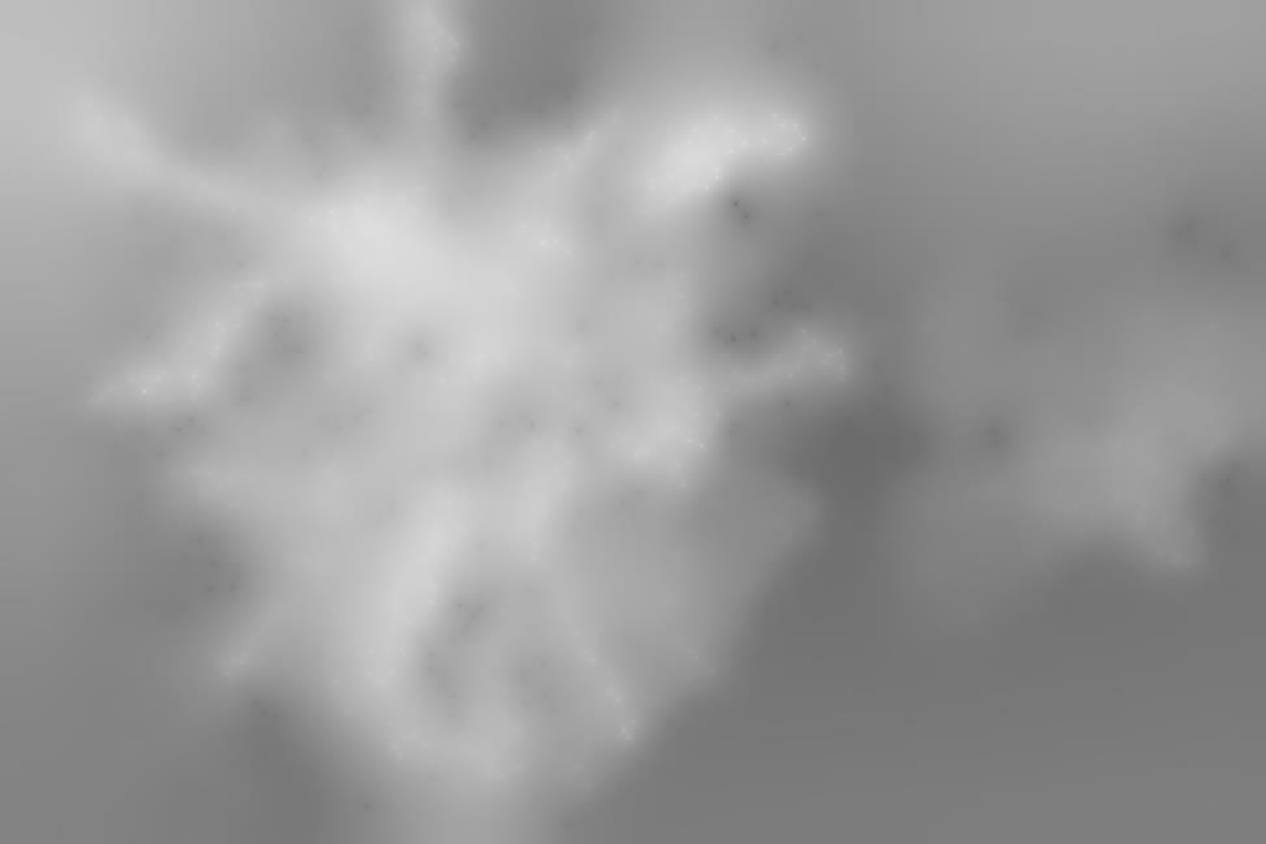}}
		{\includegraphics[height=0.128\linewidth]{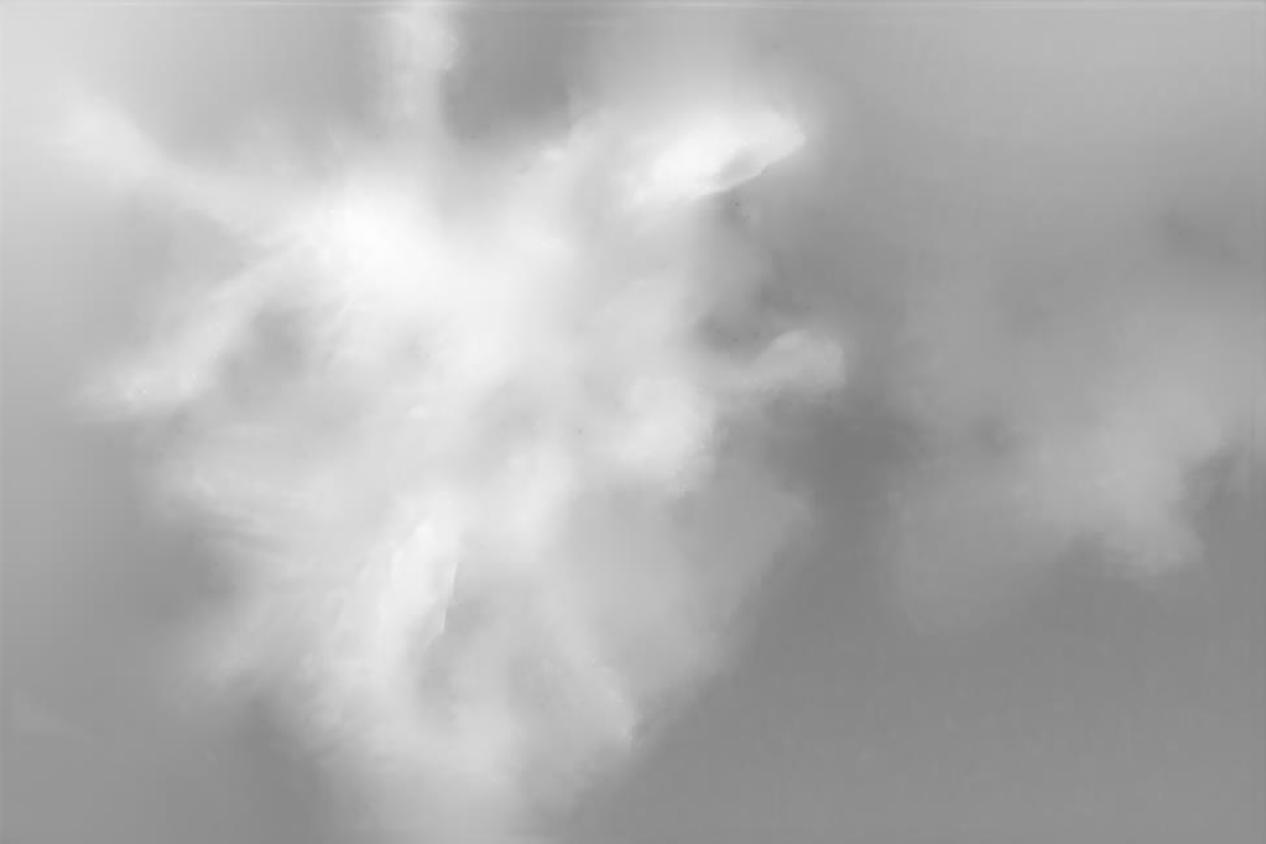}}
		{\includegraphics[height=0.128\linewidth]{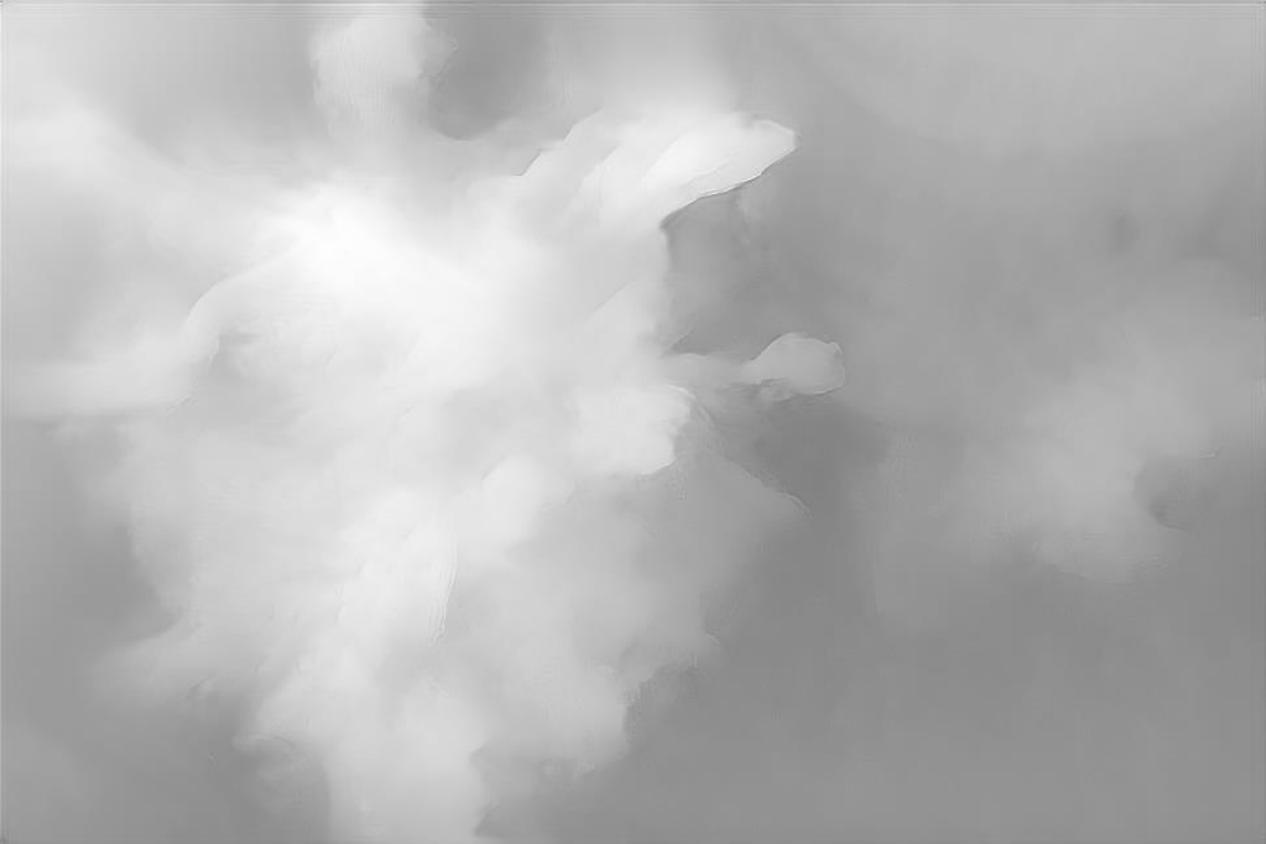}}
		{\includegraphics[height=0.128\linewidth]{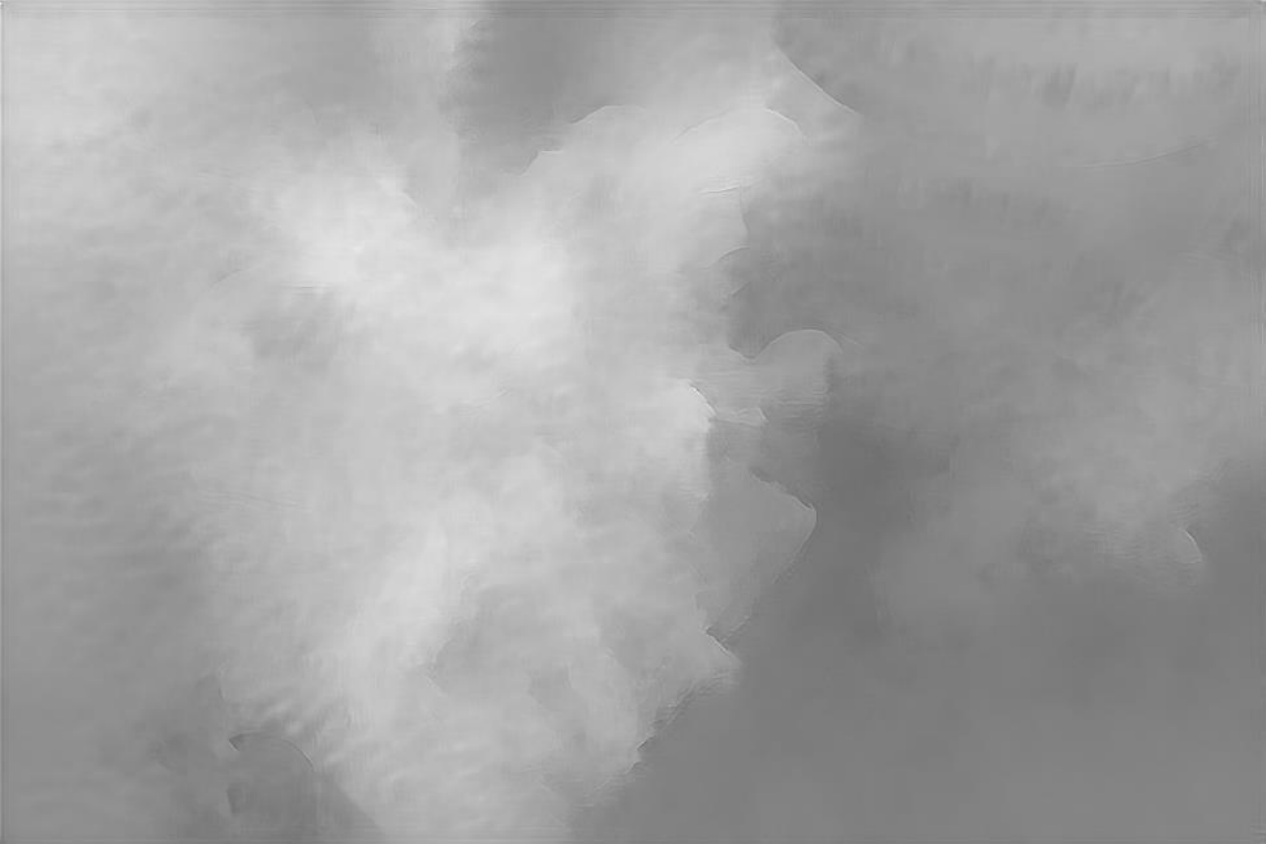}} \\
		\subcaptionbox{\label{iter_input} Reference images}{\includegraphics[height=0.128\linewidth]{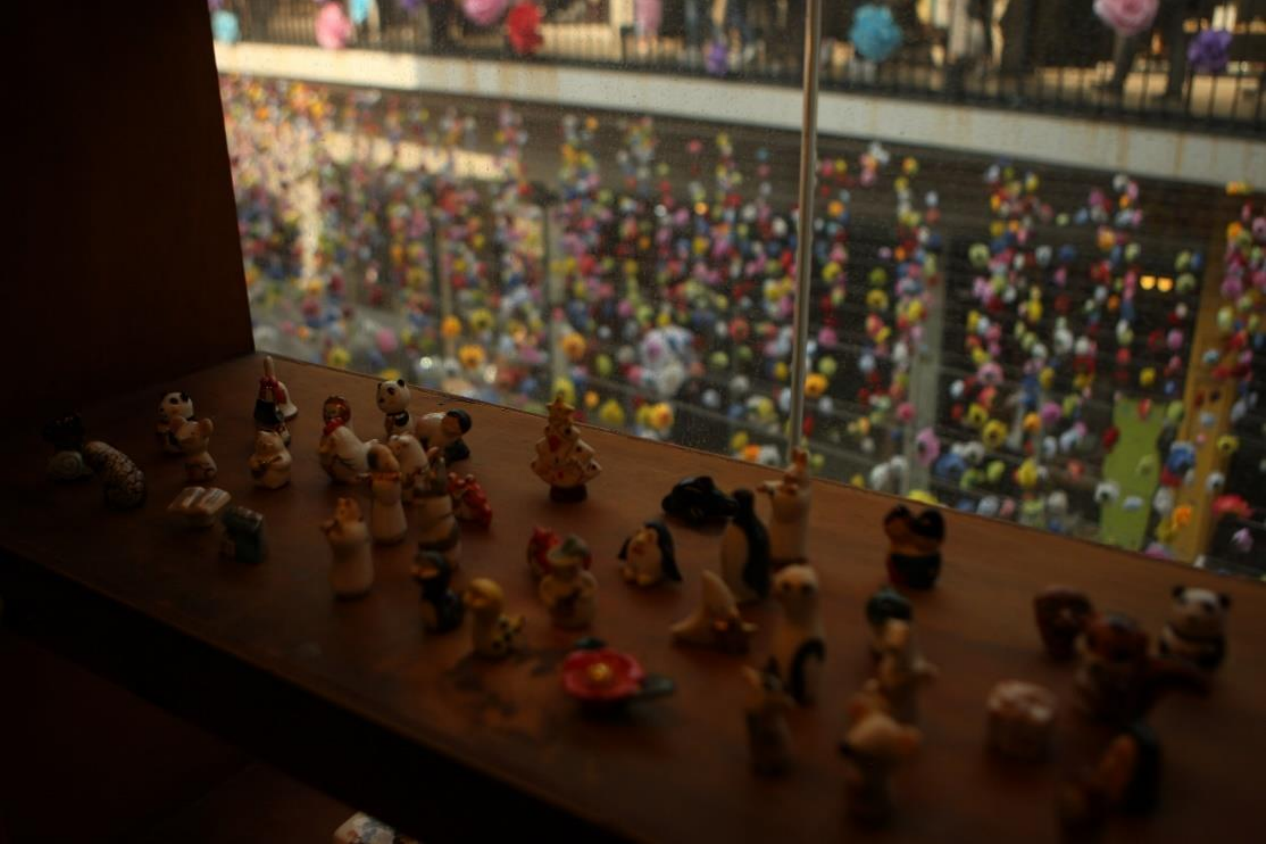}}	
		\subcaptionbox{\label{iter_init} Initial depths}{\includegraphics[height=0.128\linewidth]{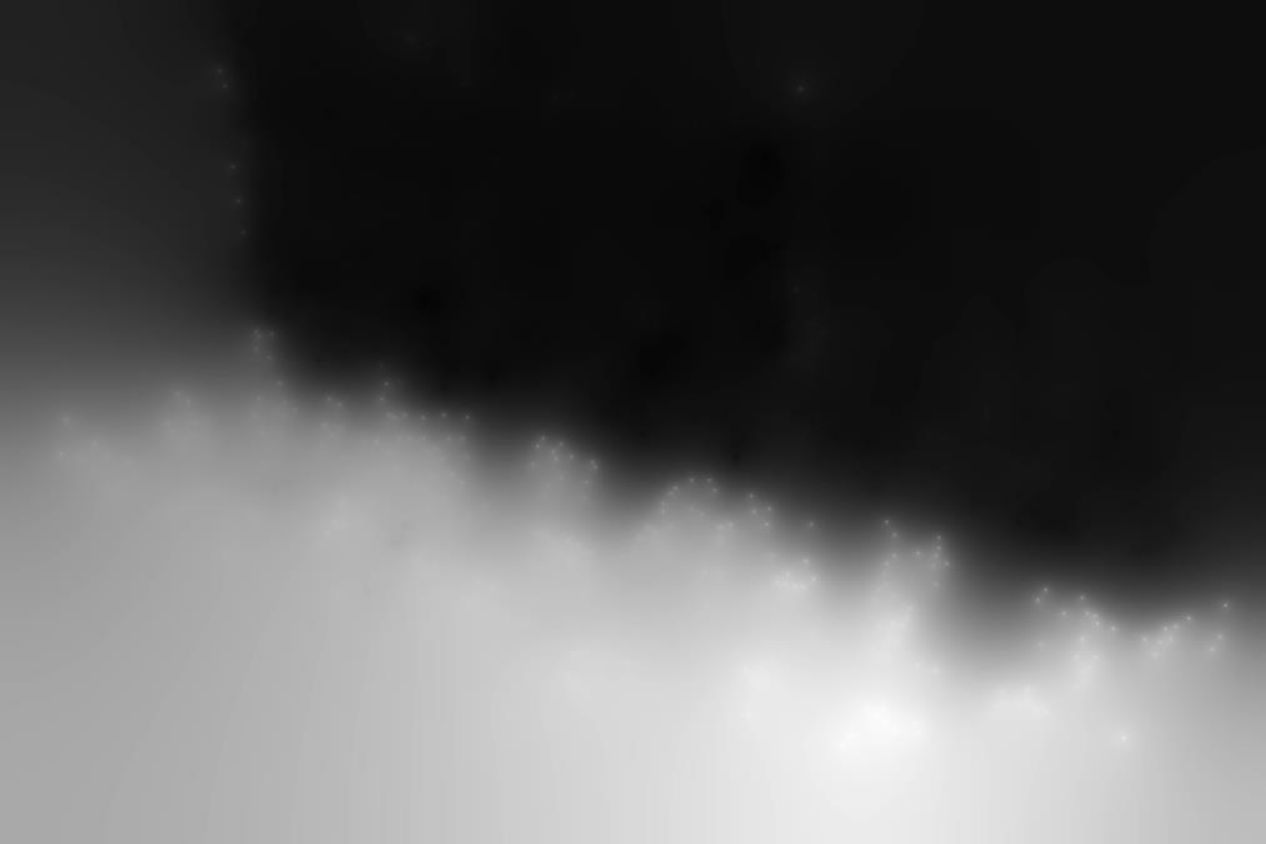}}	
		\subcaptionbox{\label{iter_mid} Intermediate depths}{\includegraphics[height=0.128\linewidth]{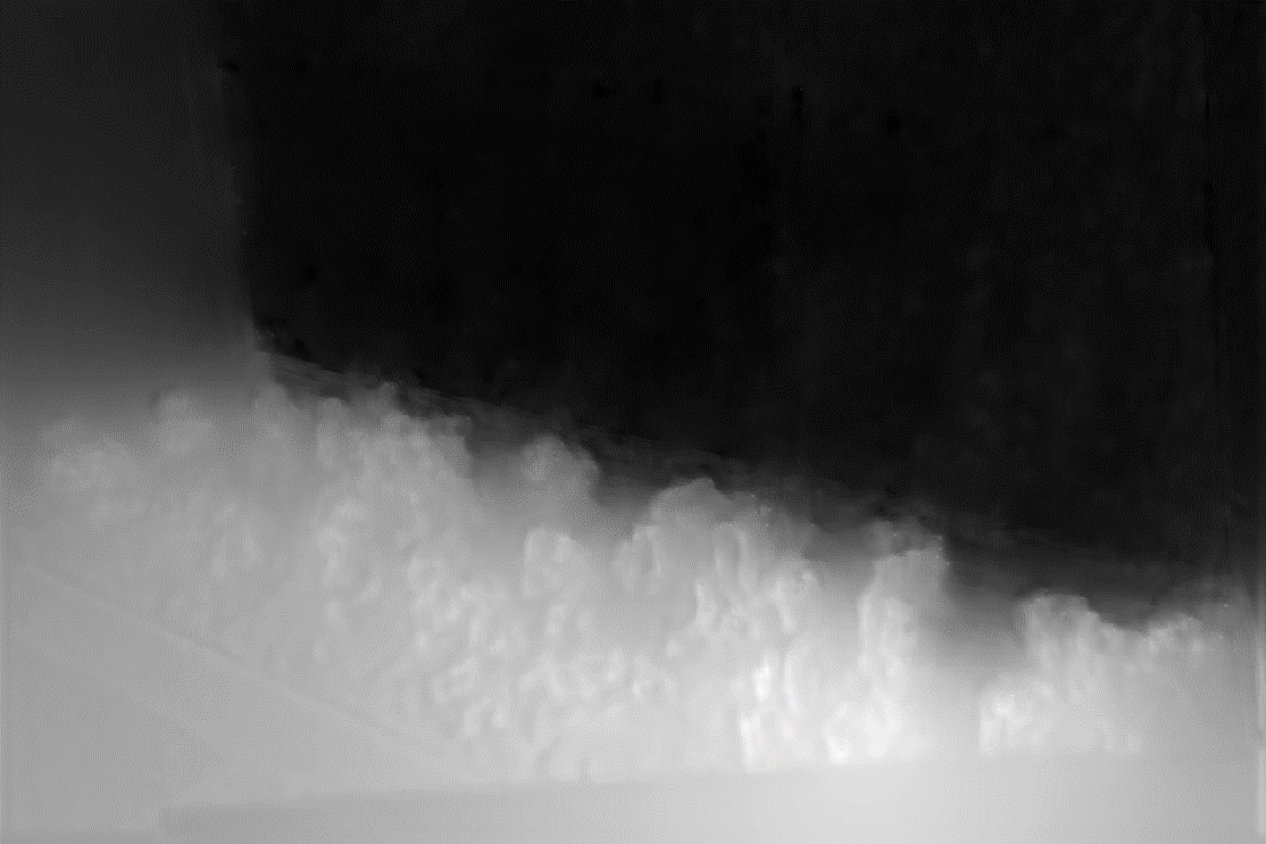}}	
		\subcaptionbox{\label{iter_final} Final depths}{\includegraphics[height=0.128\linewidth]{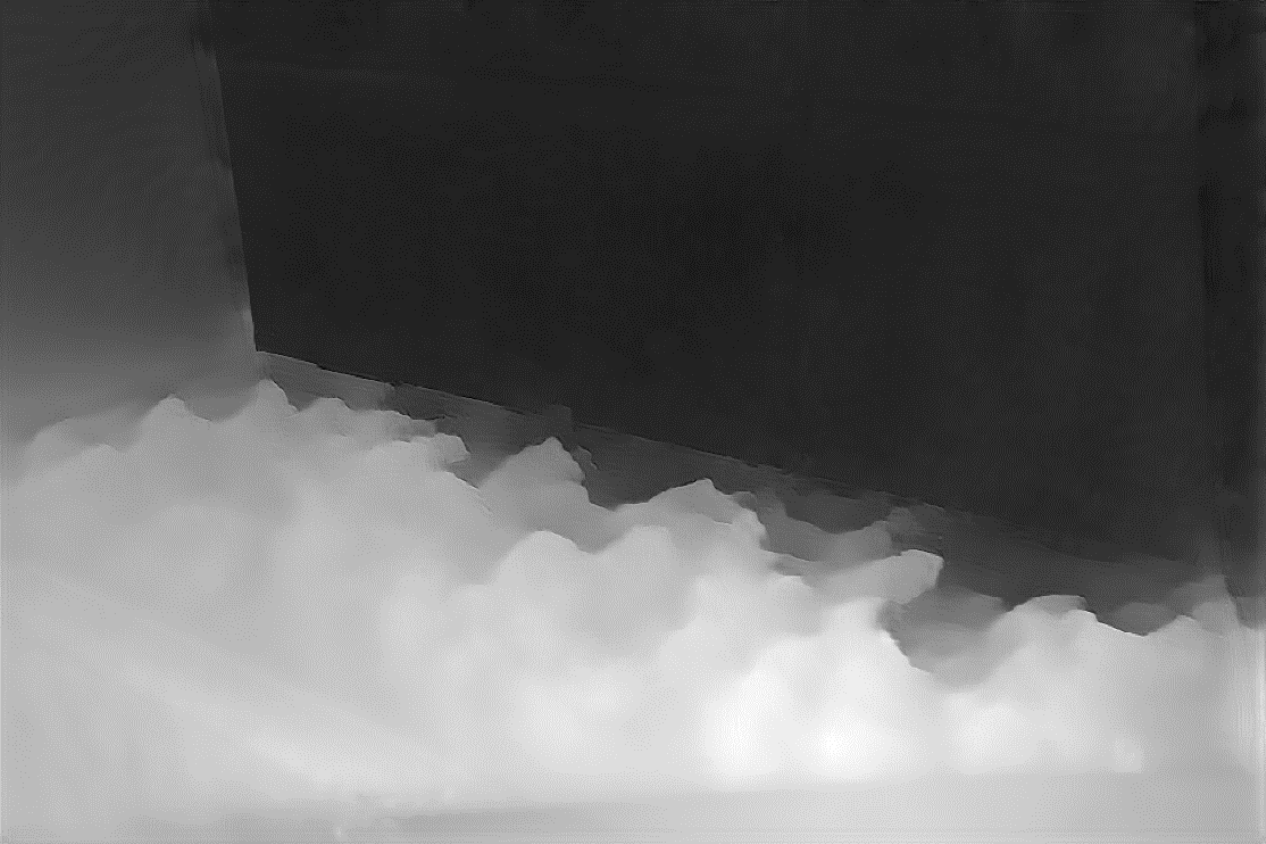}}	
		\subcaptionbox{\label{iter_noft} w/o fine-tuning}{\includegraphics[height=0.128\linewidth]{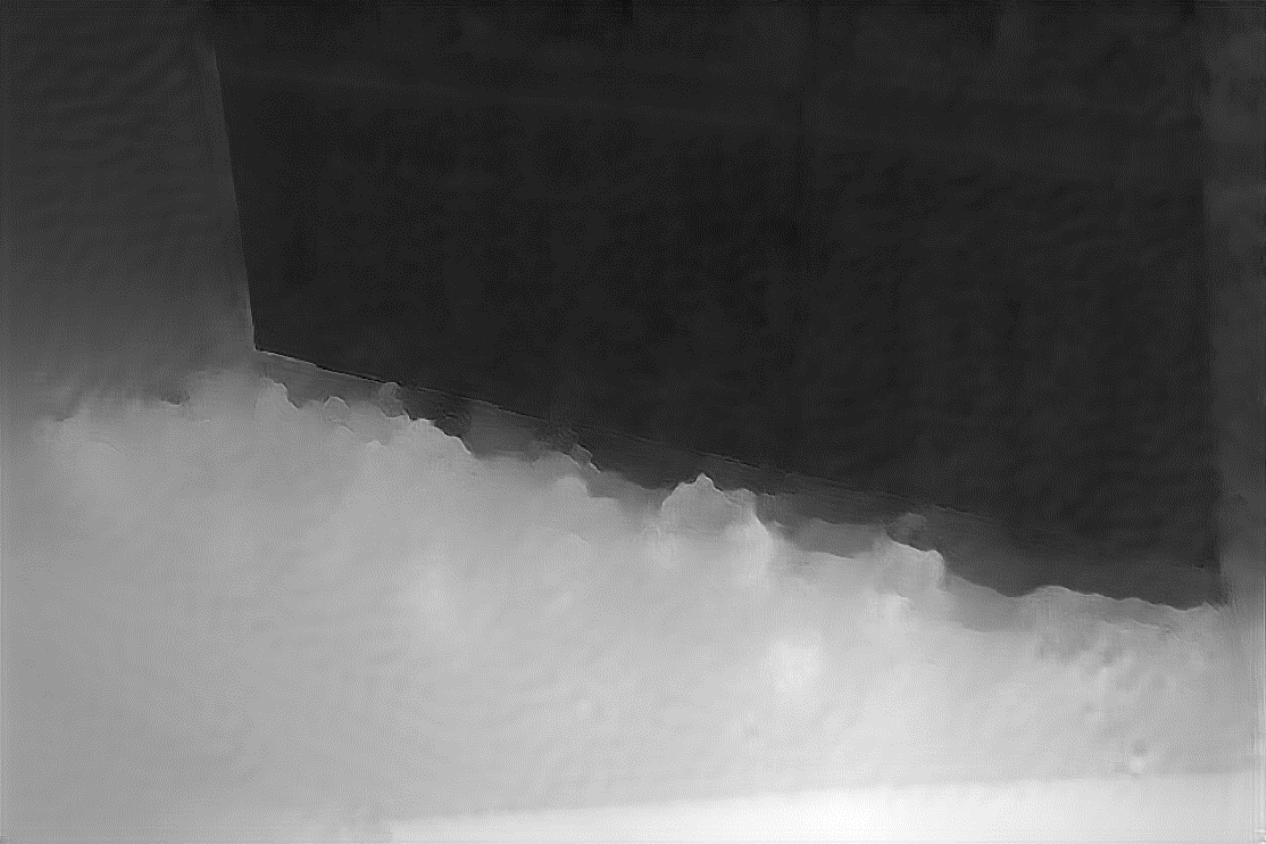}}	
	\end{tabular}
	\caption{Depth maps according to the number of iterations and fine-tuning. (a) Reference images. (b) The very first initial depths. (c) Intermediate depths from DNN. (d) Our final depths from DNN. (e) Depths from DNN without fine-tuning.}
	\label{fig:iter}		
\end{figure*}	

\subsection{Deep Multi-view Stereo Matching (DMVS)}
\label{sec:DMVS}
In this subsection, we describe the detail of our residual-flow network and the derivation of our geometrical transformation that enables to effectively match multiple images. Then, we present our DNN-based multi-view stereo method that incorporates the network and transformation. 

\noindent{\bf{Transformation of optical flow to depth}}\quad Rotation alignment reduces the complexity of the transformation between optical flow and depth, which makes our problem more tractable. To disregard the rotational motion, we start by rotating the optical axis of all images to be parallel to that of the reference image.
Given the camera intrinsic $\mathbf{K}$ and rotation $\mathbf{R}$ for all images, the synthesized images $I_i$ can be generated by warping the original images $\hat{I}_i$:




\begin{gather}
\begin{split}
\label{eq:rotating}
I_i(\mathbf{u})=\hat{I}_i\big(\big \langle\mathbf{K}\mathbf{R}_i\inv{\mathbf{K}}\mathbf{u}\big\rangle\big), i\in \{1,...,n\}.\\
\end{split}
\end{gather}
We use a bicubic interpolation for this warping process. 
Occlusion regions are ignored because the baseline of the input images is extremely narrow.
All of the images are warped except for the reference image, and the rotationally aligned images are used as the input of DNN.
Using the images with pure translation, we can derive the 2D projection of 3D points $\mathbf{X}_{j}$ (the multiplication of the normalized image coordinates of the reference frame $\mathbf{x}_{1j}$ and its depth $z_j$) into the image plane as:

\begin{gather}
\begin{split}
s \tilde{\mathbf{u}}_{ij}=\mathbf{K}\big[\mathbf{I}|\mathbf{t}_i\big]\begin{bmatrix} \mathbf{x}_{1j}z_j\\ 1 \end{bmatrix} = \begin{bmatrix} u_{1j}z_j + f_xt_i^x + c_xt_i^z \\ v_{1j}z_j + f_yt_i^y + c_yt_i^z \\ z_j + t^z_i \end{bmatrix}\\= \begin{bmatrix}
z_j & 0 & f_xt_i^x + c_xt_i^z\\
0 & z_j & f_yt_i^y + c_yt_i^z\\
0 & 0 & z_j+t_z\\
\end{bmatrix} \begin{bmatrix}
u_{1j} \\ v_{1j} \\ 1
\end{bmatrix},
\end{split}
\label{eq:projection}
\end{gather}
where $\tilde{\mathbf{u}}$ is the projected image coordinates and $s$ is the scale factor. 
Since the $z$-axis translation of the image is much smaller than the minimum scene depth ($t^z \ll z_{min}$)~\cite{Ha16}, we can assume that $(z_j + t^z)$ is approximately equivalent to $z_j$ $(\approx z_j + t^z)$. 
The projection matrix in~\eqnref{eq:projection} can be simplified as:
\begin{gather}
\begin{split}
\begin{bmatrix}
\tilde{u}_{ij} \\ \tilde{v}_{ij}
\end{bmatrix} = \begin{bmatrix}
1 & 0 & T_i^x/z_j\\
0 & 1 & T_i^y/z_j\\
\end{bmatrix}\begin{bmatrix}
u_{1j} \\ v_{1j} \\ 1
\end{bmatrix} = \begin{bmatrix}
u_{1j} \\ v_{1j}
\end{bmatrix} + w_j\begin{bmatrix}
T_i^x \\ T_i^y
\end{bmatrix},\\
T_i^x = f_xt_i^x + c_xt_i^z,~~T_i^y = f_yt_i^y + c_yt_i^z,
\end{split}
\label{eq:transform}
\end{gather}
where $w_j$ is the inverse depth $1/z_j$. Based on~\eqnref{eq:transform}, the transformation vector $\mathbf{T}_i$ can convert the inverse depth vector $\mathbf{w}$ into the flow field $\mathbf{v}_i$ from the reference image to $i^{th}$ target image as follows:
\begin{gather}
\begin{split}
\mathbf{v}_i = \mathbf{T}_i\mathbf{w},~~\text{where}~\mathbf{T}_i = \begin{bmatrix}
T_i^x \\ T_i^y
\end{bmatrix}.\\
\end{split}
\end{gather}

\begin{figure}[t]
	\centering
	\begin{tabular}{c@{\hspace{1mm}}}
		\includegraphics[height=0.56\linewidth]{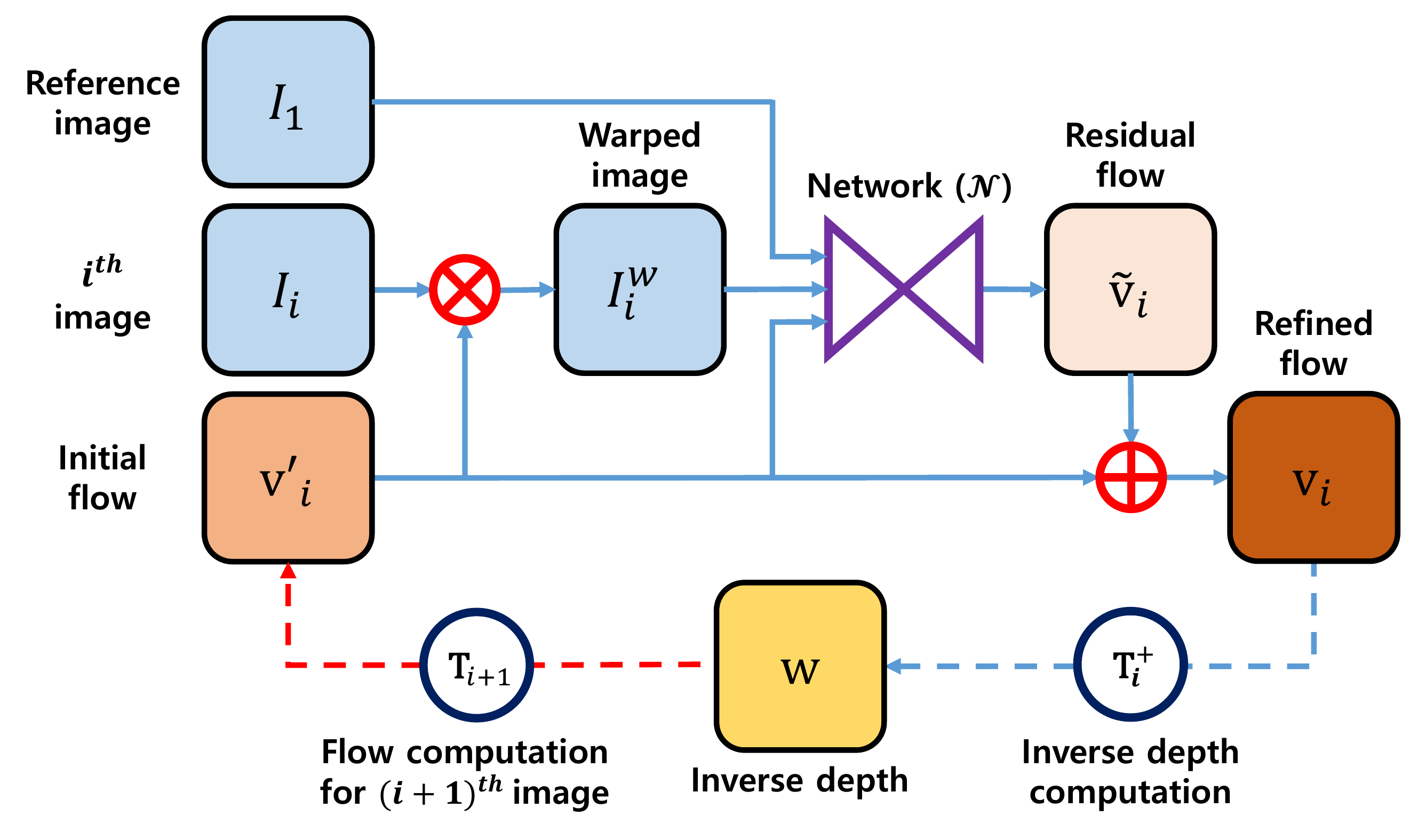}
	\end{tabular}
	\caption{Overview of DMVS. The solid line shows the optical flow refinement process. The blue dashed line shows the conversion
		of optical flow into inverse depth. The red dashed line shows the inverse depth, which is converted into an optical flow and used for the initial flow of the next frame.}
	\label{fig:network}
\end{figure}

\begin{figure*}[t]
	\centering
	\begin{tabular}{c@{\hspace{1mm}}c@{\hspace{1mm}}c@{\hspace{1mm}}c@{\hspace{1mm}}}
		\includegraphics[height=0.128\linewidth]{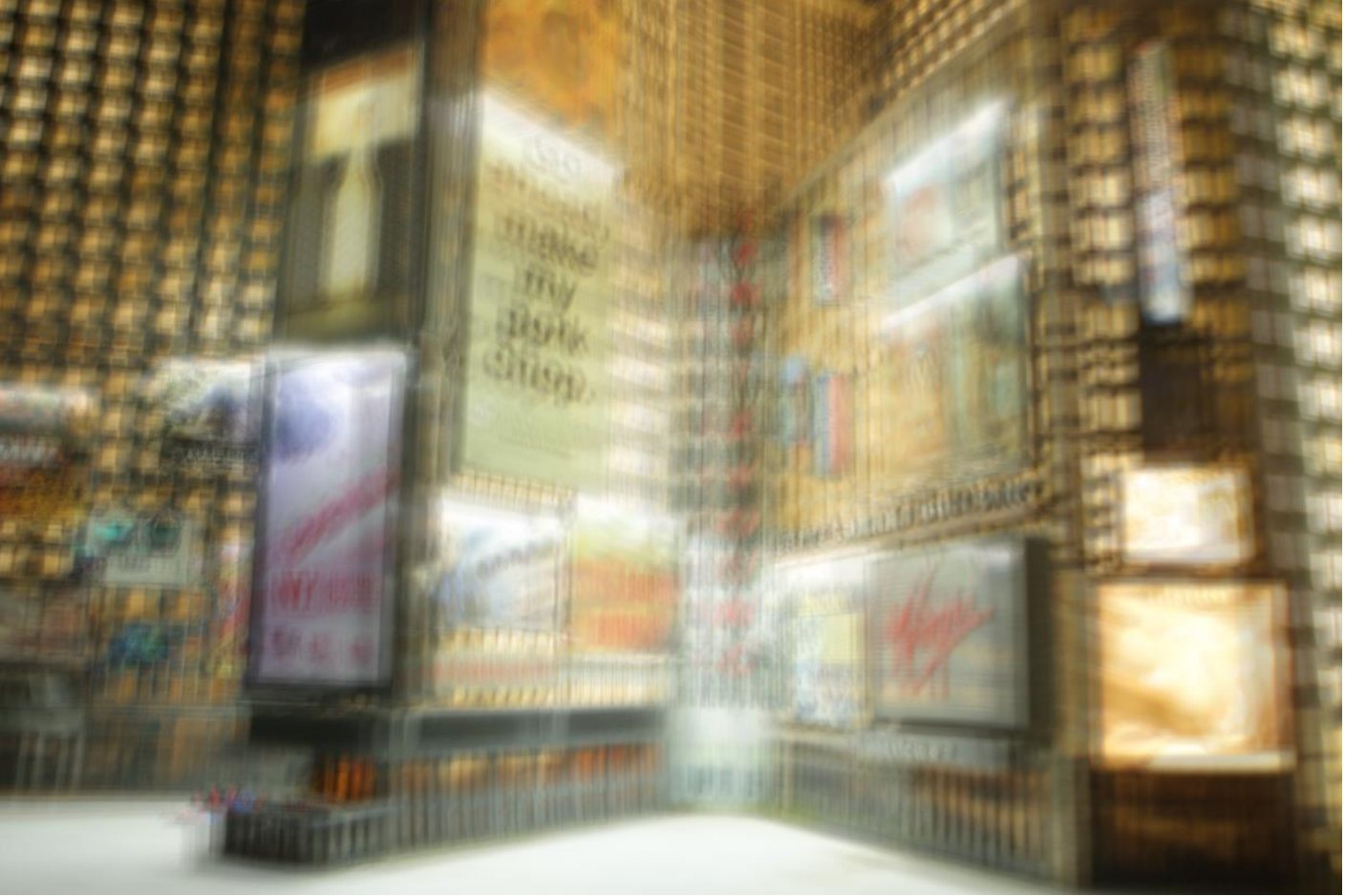}	
		\includegraphics[height=0.128\linewidth]{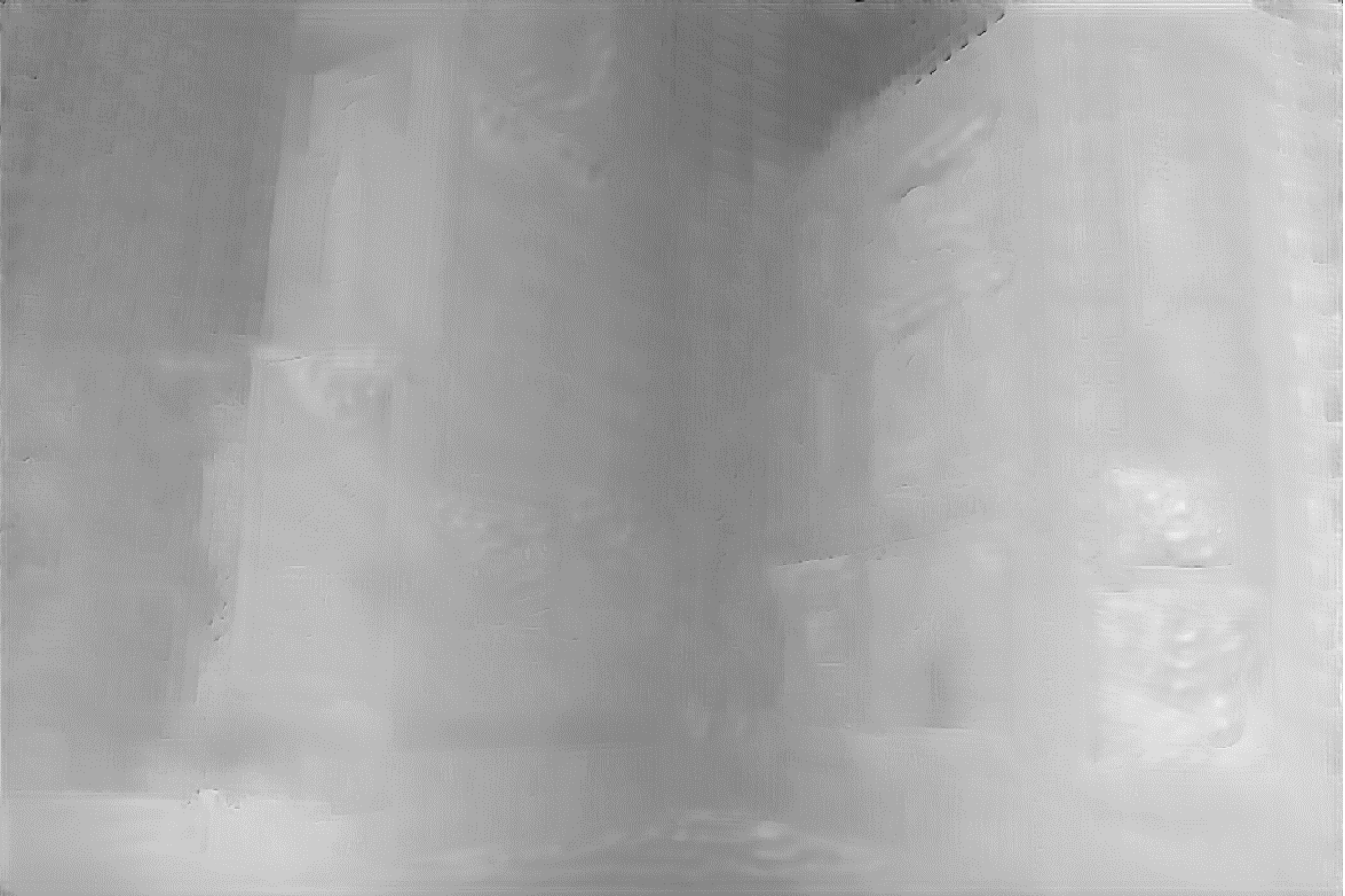}		
		\includegraphics[height=0.128\linewidth]{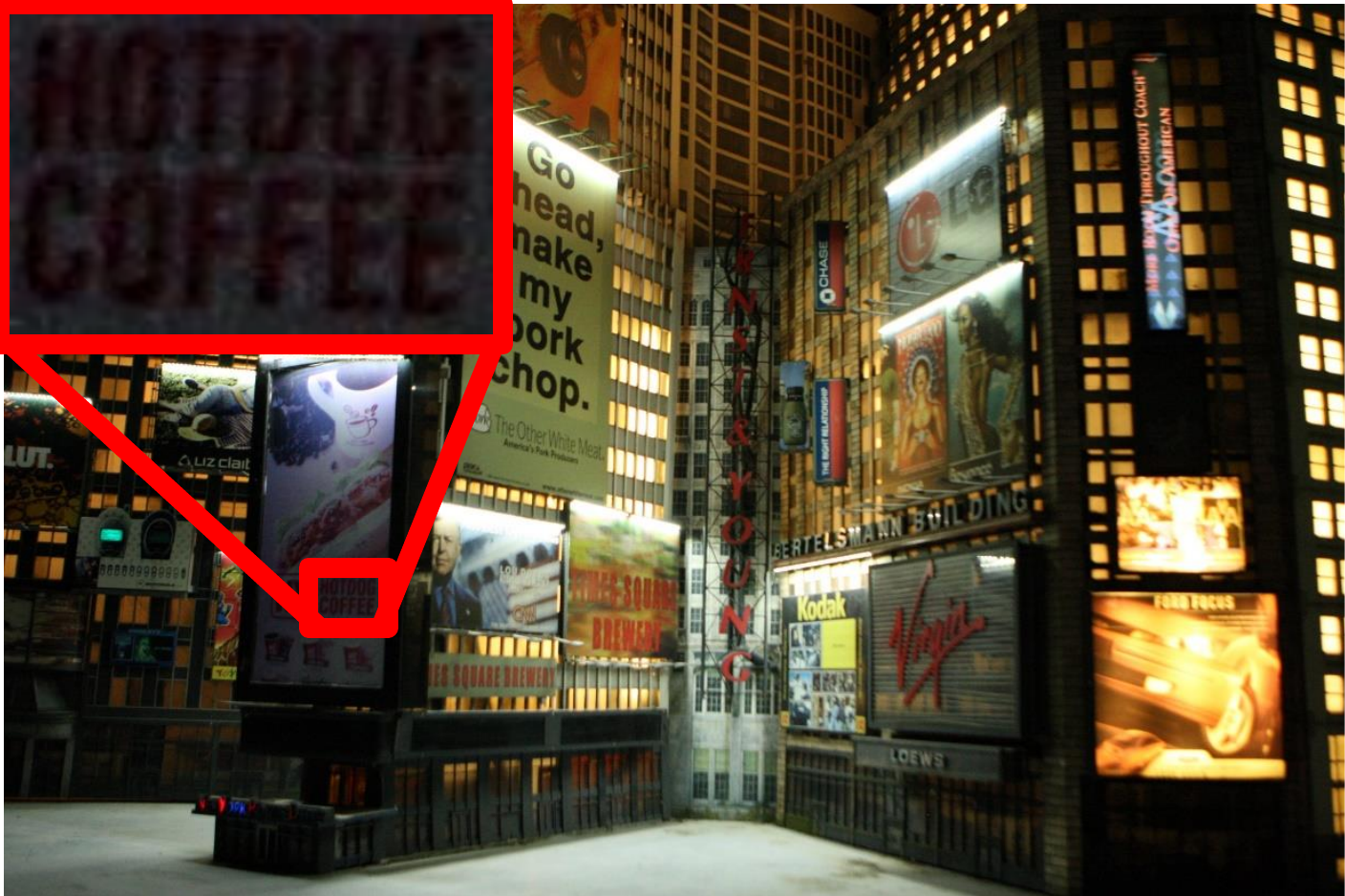}	
		\includegraphics[height=0.128\linewidth]{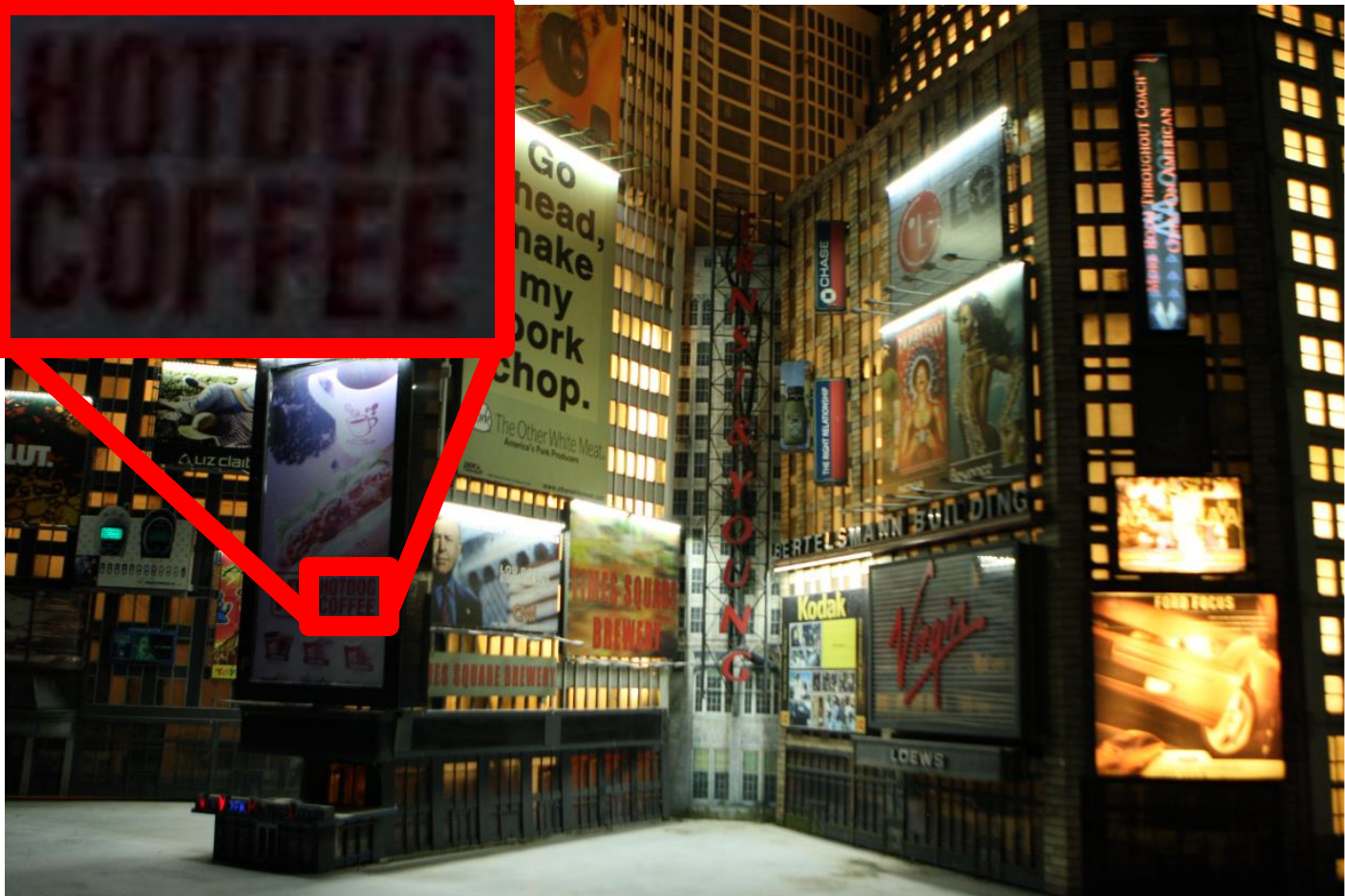}	
		\includegraphics[height=0.128\linewidth]{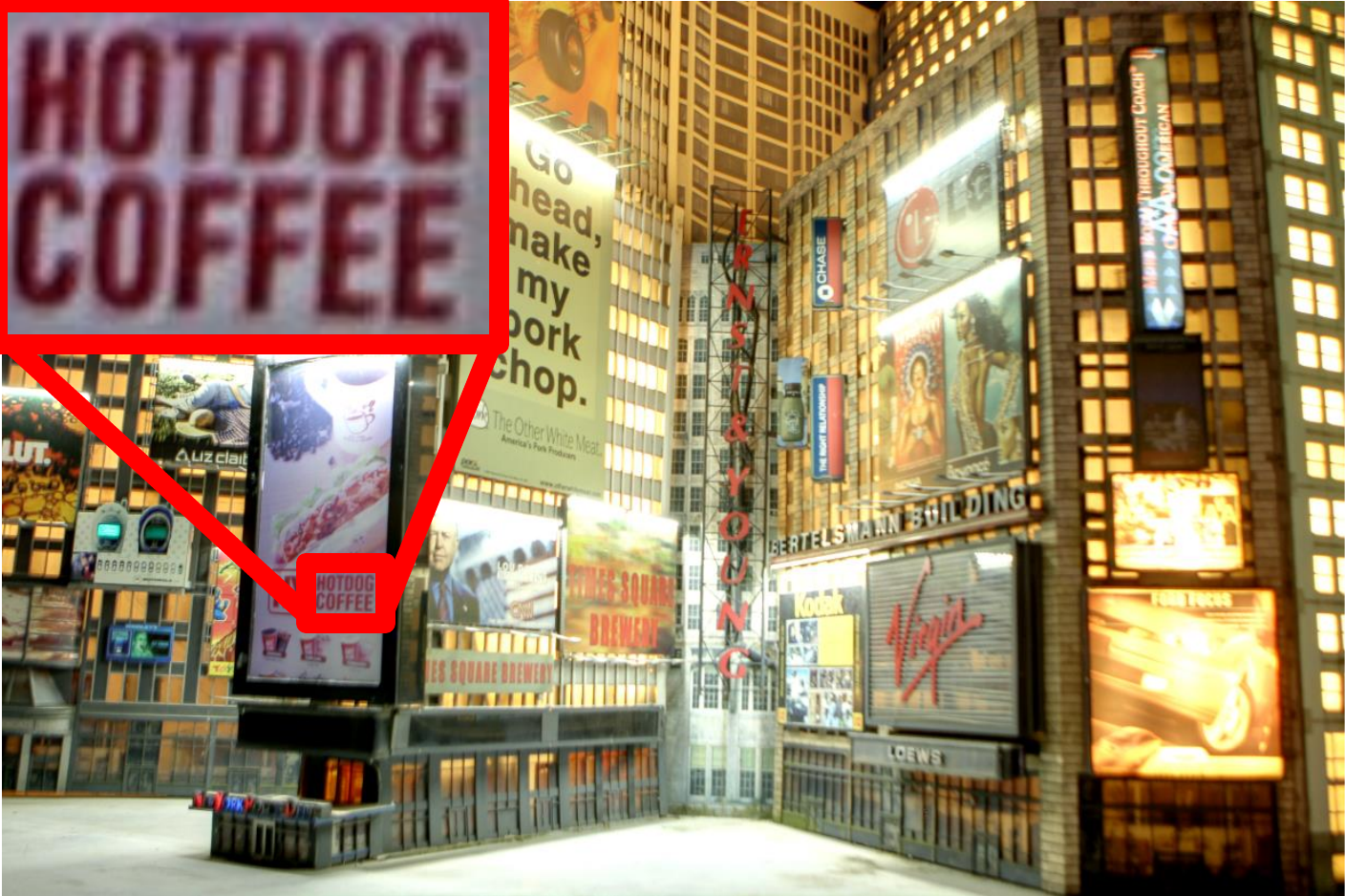}\\
		\subcaptionbox{\label{final_ai} Averaged images}{\includegraphics[height=0.128\linewidth]{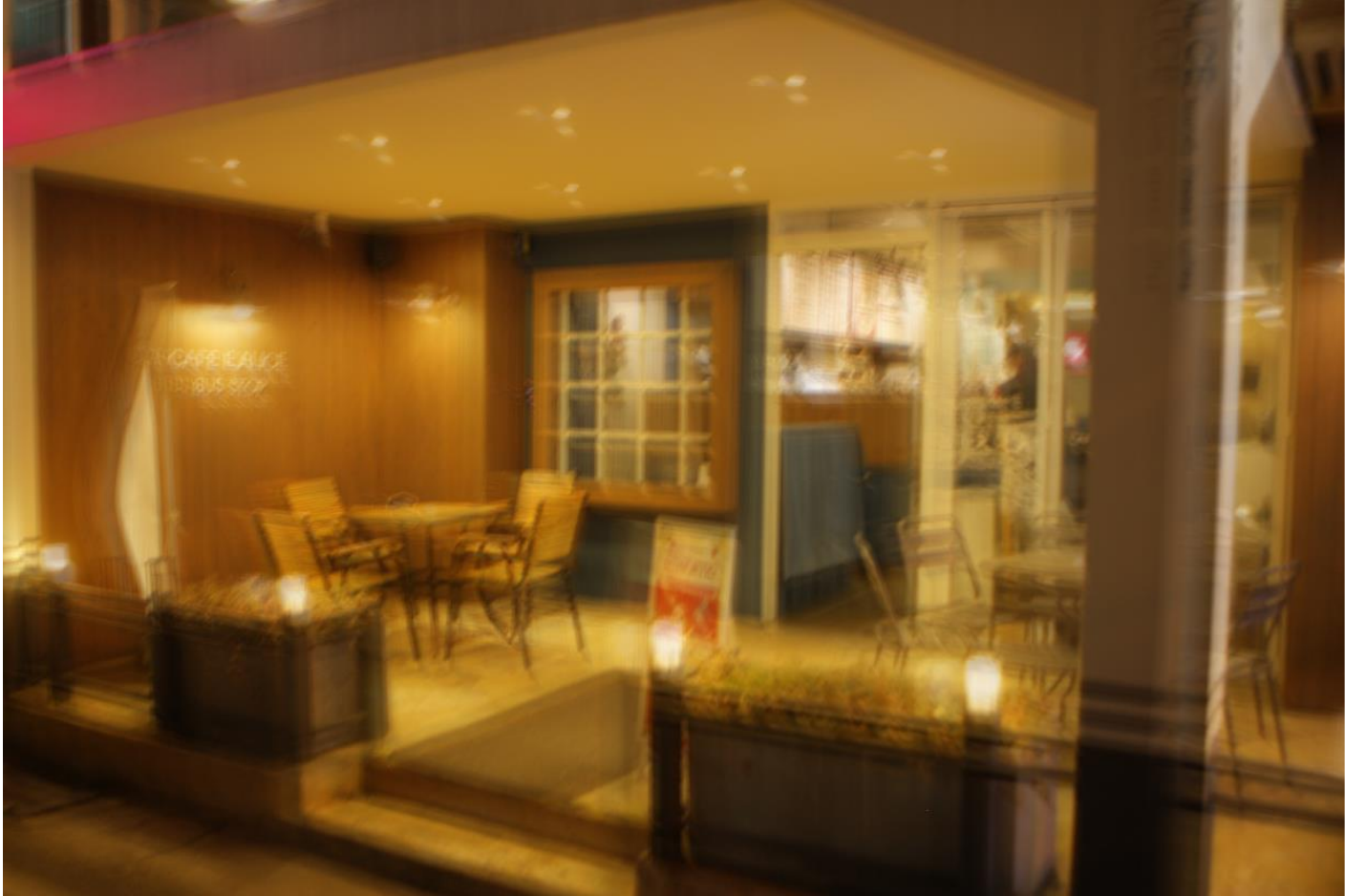}}	
		\subcaptionbox{\label{final_d} Our depths}{\includegraphics[height=0.128\linewidth]{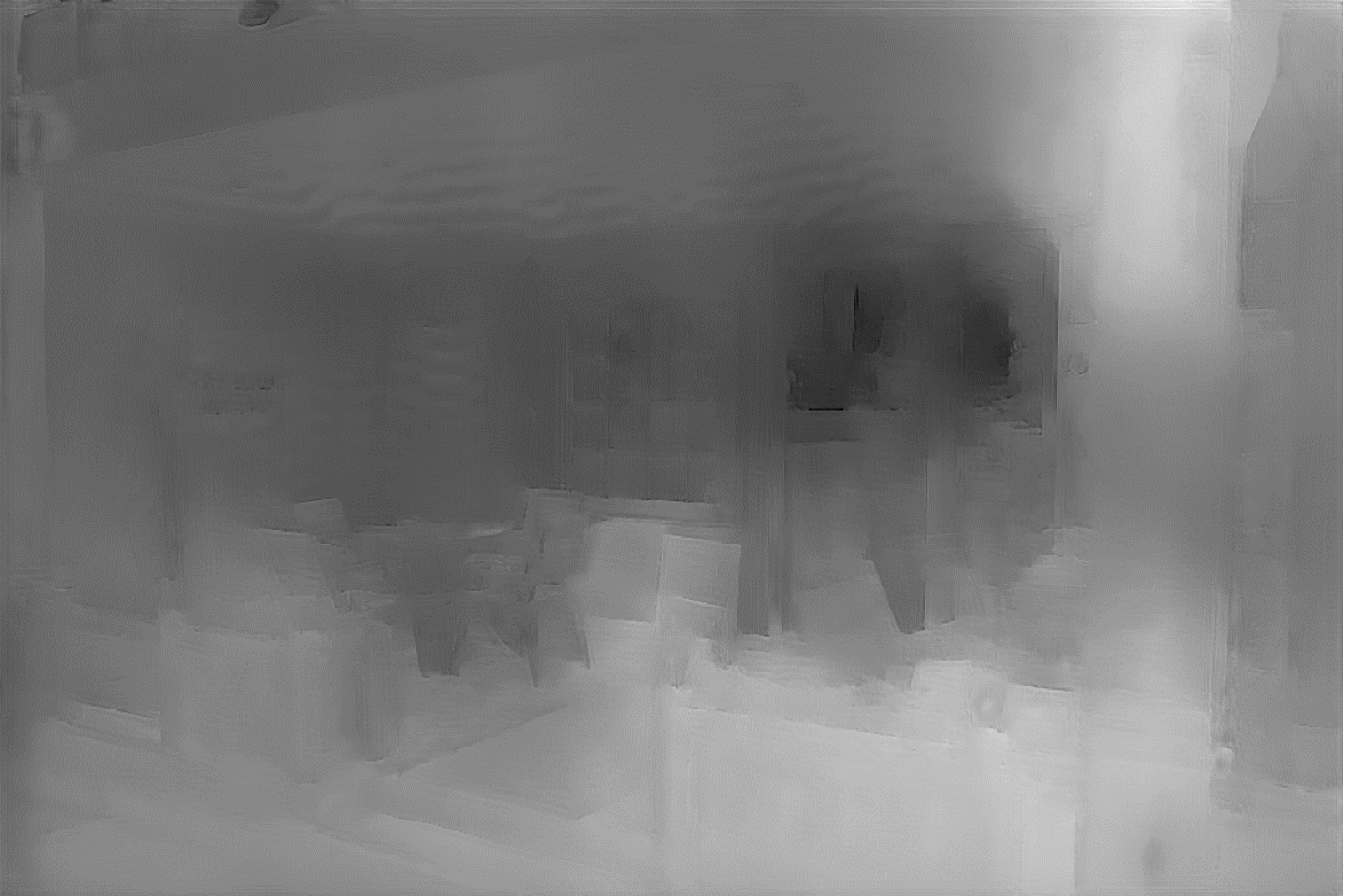}}		
		\subcaptionbox{\label{final_input} Reference images}{\includegraphics[height=0.128\linewidth]{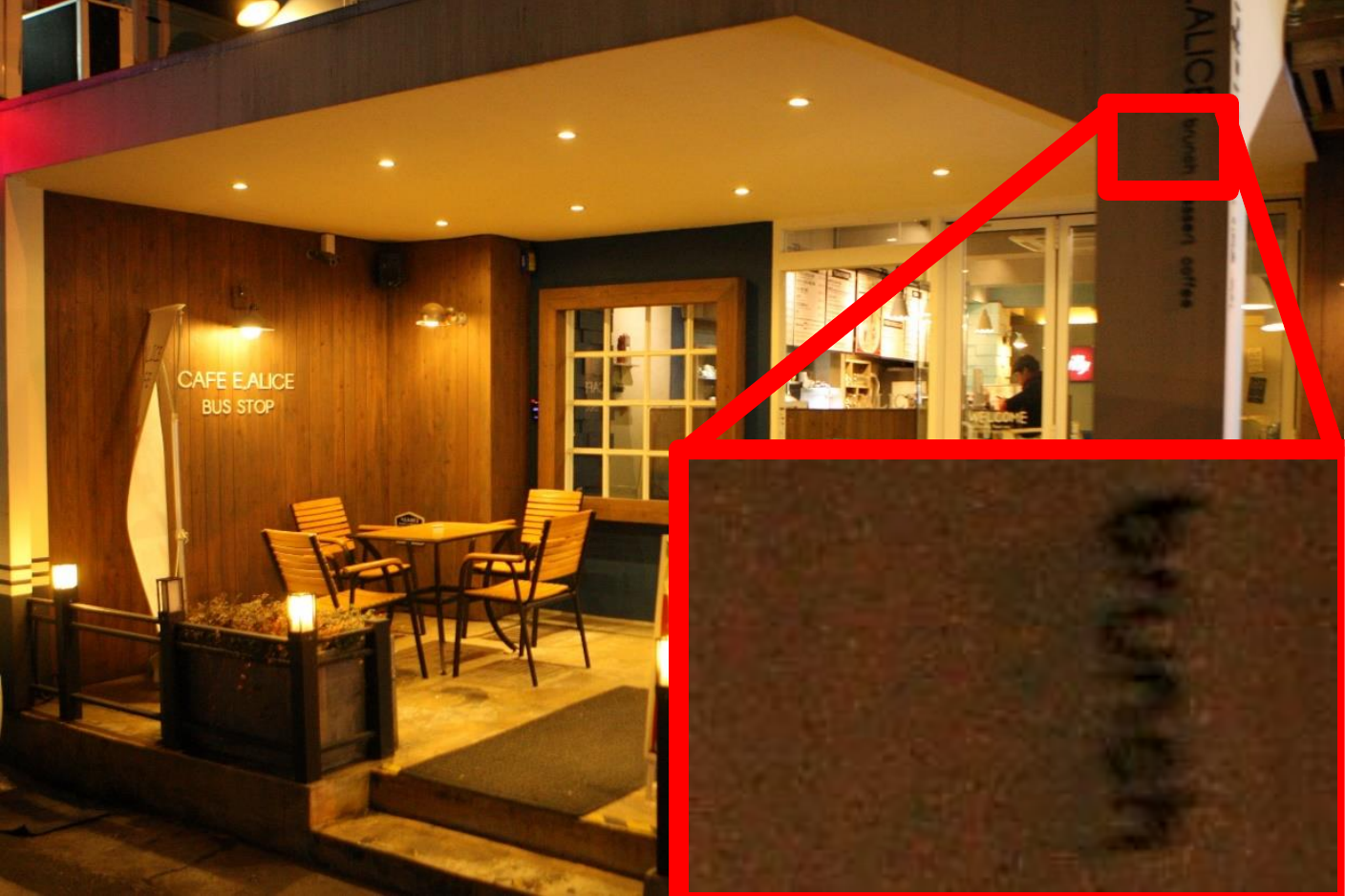}}	
		\subcaptionbox{\label{final_dn} Denoising}{\includegraphics[height=0.128\linewidth]{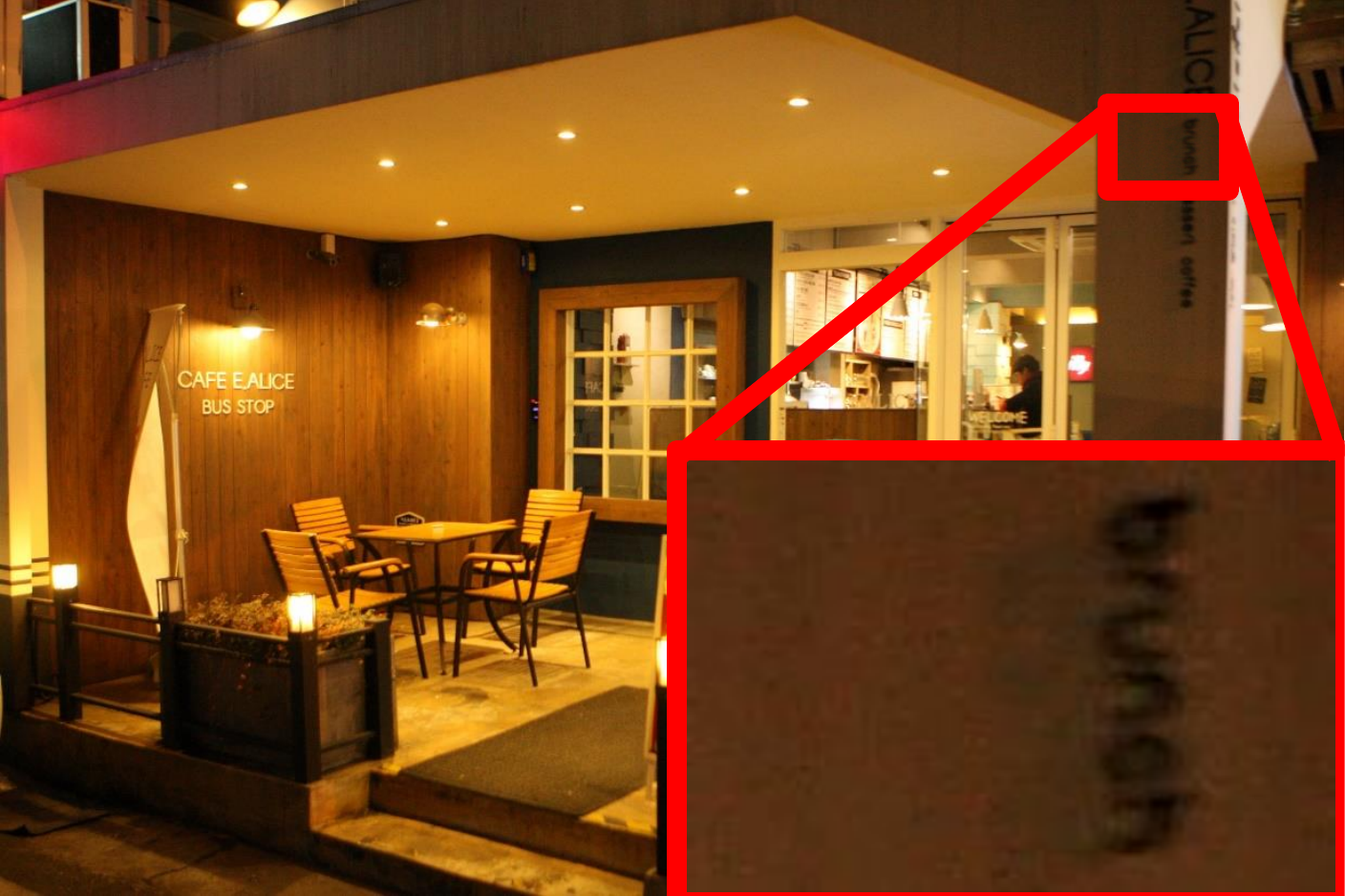}}	
		\subcaptionbox{\label{final_ef} Exposure fusion }{\includegraphics[height=0.128\linewidth]{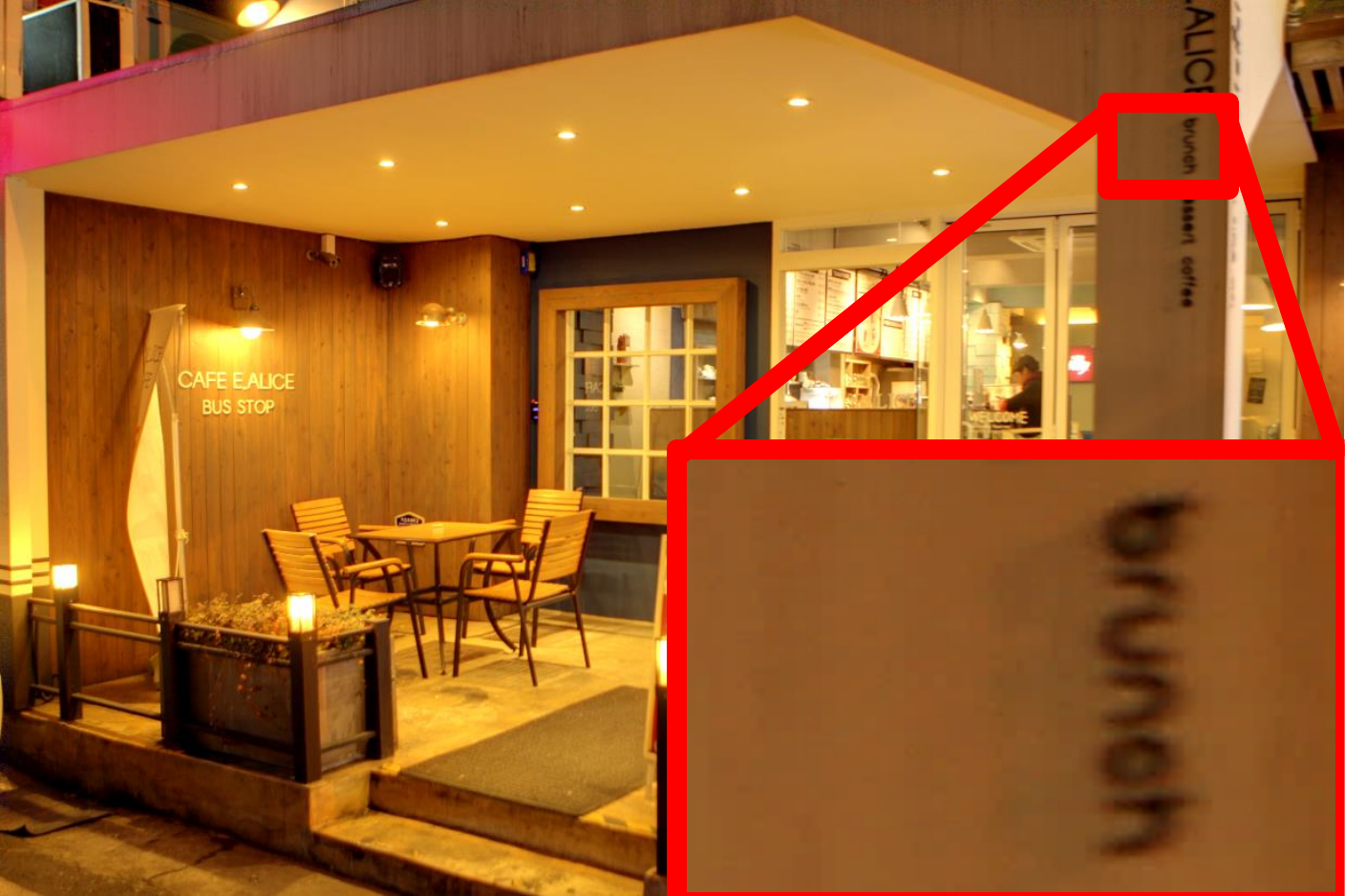}}
	\end{tabular}
	\caption{Averaged image of input exposure bracketed images, our depths and example of photographic applications (denoising, exposure fusion) using aligned images. }
	\label{fig:final}
	
\end{figure*}

\noindent{\bf{Depth estimation using residual flow network}} \quad 
The basic idea of our depth estimation scheme is to iteratively refine the inverse depth $\mathbf{w}$ using the optical flow estimated by the DNN as shown in~\figref{fig:network}. 
The network $\mathcal{N}$ computes the residual flow $\tilde{\mathbf{v}}_i$ with the 8-channel input: the reference image $I_1$, the warped image $I^w_i$ and the initial optical flow $\mathbf{v}'_i$.  
The initial flow is obtained by propagating the sparse 3D points in~\secref{sec:sfsm} using the closed form solution~\cite{Im15}, which then is transformed into a flow field.
We obtain the warped image using the bilinear sampler $S(I_i,\mathbf{v}'_i)$. 
After the residual flow is estimated, the initial flow and the residual flow are added to obtain the refined flow $\mathbf{v}_i$. 
We convert the refined flow to the flow of the next frame using the transformation vectors by utilizing them as an initial flow:
\begin{gather}
\mathbf{v}'_i = \mathbf{T}_{i}\mathbf{T}^{+}_{i-1}\mathbf{v_{i-1}},
\end{gather}
where $\mathbf{T}^+$ is the pseudo inverse of the vector $\mathbf{T}$. 
We estimate the final depth $\mathbf{z}$ by transforming the optical flow of the last image $\mathbf{T}^+_{n}\mathbf{v}_{n}$ into the inverse depth $\mathbf{w}$ and dividing it by one. 

\figref{fig:iter} shows the effectiveness of the refinement process. 
The initial depth maps in~\figref{iter_init} show inaccurate depth discontinuities, which is not suitable for the precise image alignment and other depth-aware photographic applications.
On the other hand, the intermediate and final depth in~\figref{iter_mid} and~\figref{iter_final} shows that our DNN produces more detailed and artifact-free depth results.

\noindent{\bf{Training and network architecture}} \quad
Our network consists of two convolution and three deconvolution layers with the fixed kernel size ($7\times7$) and stride (1) as described in~\tabref{tab:spec}. 
All layers with the exception of the last layer are followed by a Rectified Linear Unit (ReLU). 	
Taking a coarse-to-fine strategy similar to the optical flow estimation, we train the network to learn residual flow $\tilde{\mathbf{v}}$, instead of directly estimating the depth or optical flow. 
We stack the reference image, the warped pair image and the initial optical flow to form an 8-channel input for our network. 
We set the target residual flows $\tilde{\mathbf{v}}^{gt}_i$ as the difference between the target flow $\mathbf{v}^{gt}_i$ and the optical flow $\mathbf{v}'_i$ obtained from the trained network at the $5^{th}$ pyramid level~\cite{RanjanB16}:
\begin{gather}
\tilde{\mathbf{v}}^{gt}_i = \mathbf{v}^{gt}_i - \mathbf{v}'_i.
\end{gather}
In the training step, we minimize the average endpoint error (EPE), which is the standard error measure for optical flow estimation. 
This is the Euclidean distance between the residual flow $\tilde{\mathbf{v}}_i$ and the target residual flows $\tilde{\mathbf{v}}^{gt}_i$. 

The optimization is carried out using ADAM~\cite{kingma2014adam} with the recommended parameters $\beta_1 = 0.9$ and $\beta_2 = 0.999$. 
The initial learning rate is $\lambda = 1e{-}4$, then decreased to $1e{-}5$ after 60 epochs. 
We use the Flying Chairs dataset~\cite{dosovitskiy2015flownet} with a resolution of $384\times512$ at training time. 
The training is performed with a customized version of Torch7~\cite{collobert2011torch7} on a Nvidia 1080 GPU, which usually takes 24 hours.

We chose to perform various types of data augmentation during training. 
We perform spatial (rotation, scaling) and chromatic transformations (color, brightness, contrast, Gaussian noise).
We augment input patches with random rotations within $[-17^\circ,17^\circ]$ and scaling within $[1,2]$.
The noise level is uniformly sampled from $\mathcal{N}(0, 0.1)$. 
We also apply color jitter with additive brightness, contrast and saturation sampled from a Gaussian, $\mathcal{N}(0,0.4)$. 
At the end, we normalize the intensity of the images using a mean and standard deviation computed from a large corpus of ImageNet~\cite{russakovsky2015imagenet}.

The trained network produces accurate residual flow on images captured with constant camera settings, but it causes some artifacts with a different setting (\eg, exposure, ISO) as shown in~\figref{iter_noft}. 
To alleviate this problem, we fine-tune the network using the different color jitter value $\mathcal{N}(0, 0.4)$ in the reference $I_1$ and the target images $I_i$. The fine-tuning step generates a synthetic image pair with the different camera settings (\eg, exposure, ISO). 
We also apply the other data augmentation and intensity normalization in this fine-tuning step using the learning rate $\lambda = 1e{-}5$. \figref{iter_final} shows the performance improvement in the network fine-tuning.


\begin{table}[t]
	\centering
	\small
	\caption{Specification of our architecture}
	\begin{tabular}{|c|c|c|c|c|c|}
		\hline
		Name & Kernel & Str. & Ch I/O & Input \\ \hline
		conv1 & 7$\times$7 & 1 & 8/32 & Images/Flow \\ \hline
		conv2 & 7$\times$7 & 1 & 32/64 & conv1 \\ \hline
		deconv2 & 7$\times$7 & 1 & 64/32 & conv2 \\ \hline
		deconv1 & 7$\times$7 & 1 & 32/16 & deconv2 \\ \hline
		deconv0 & 7$\times$7 & 1 & 16/2 & deconv1 \\ \hline
	\end{tabular}
	\label{tab:spec}
\end{table}		

\begin{figure*}[!t]
	\centering		
	\begin{tabular}{c@{\hspace{1mm}}c@{\hspace{1mm}}c@{\hspace{1mm}}c@{\hspace{1mm}}c@{\hspace{1mm}}}
		{\includegraphics[height=0.14\linewidth]{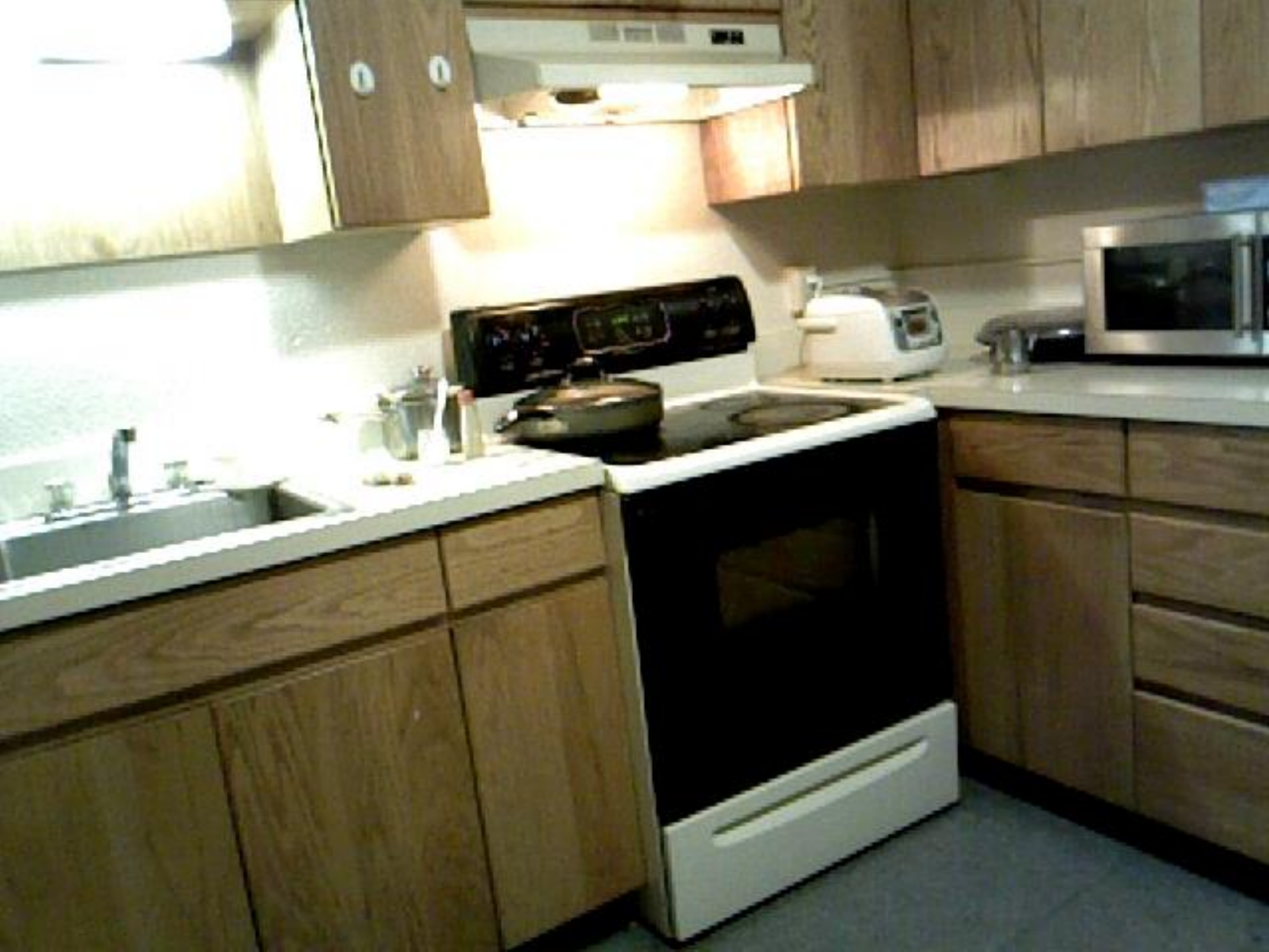}}
		{\includegraphics[height=0.14\linewidth]{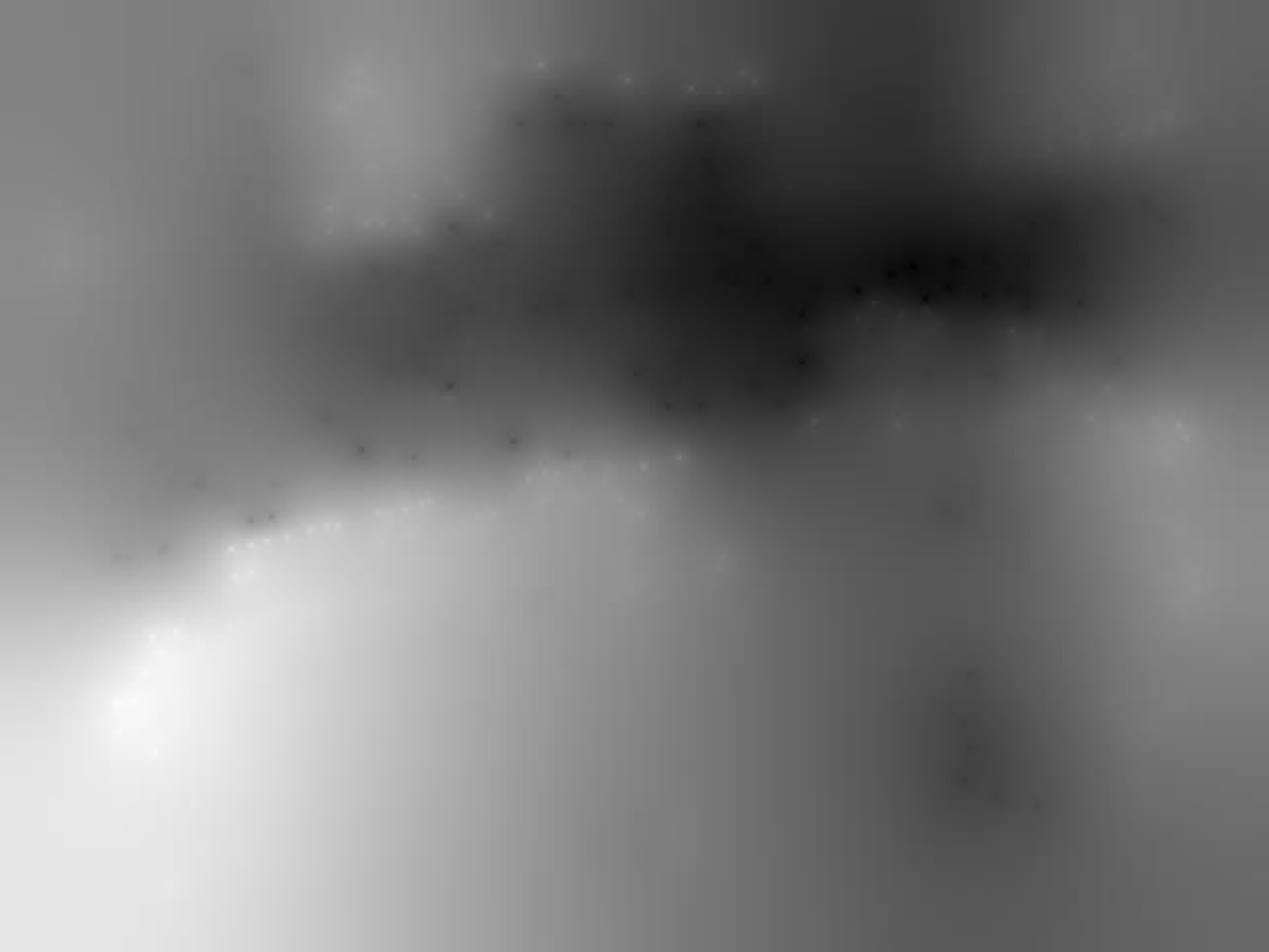}}
		{\includegraphics[height=0.14\linewidth]{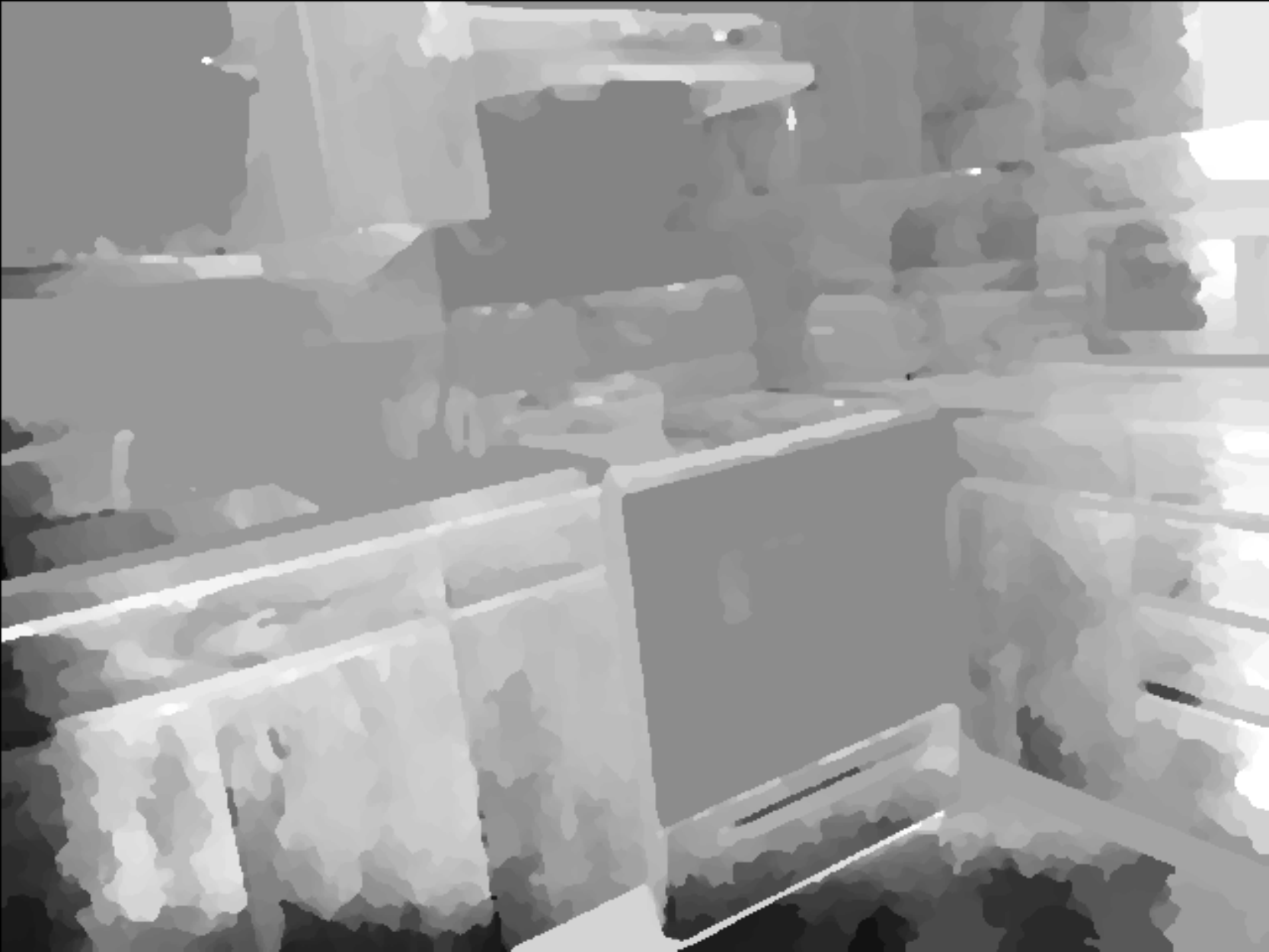}}
		{\includegraphics[height=0.14\linewidth]{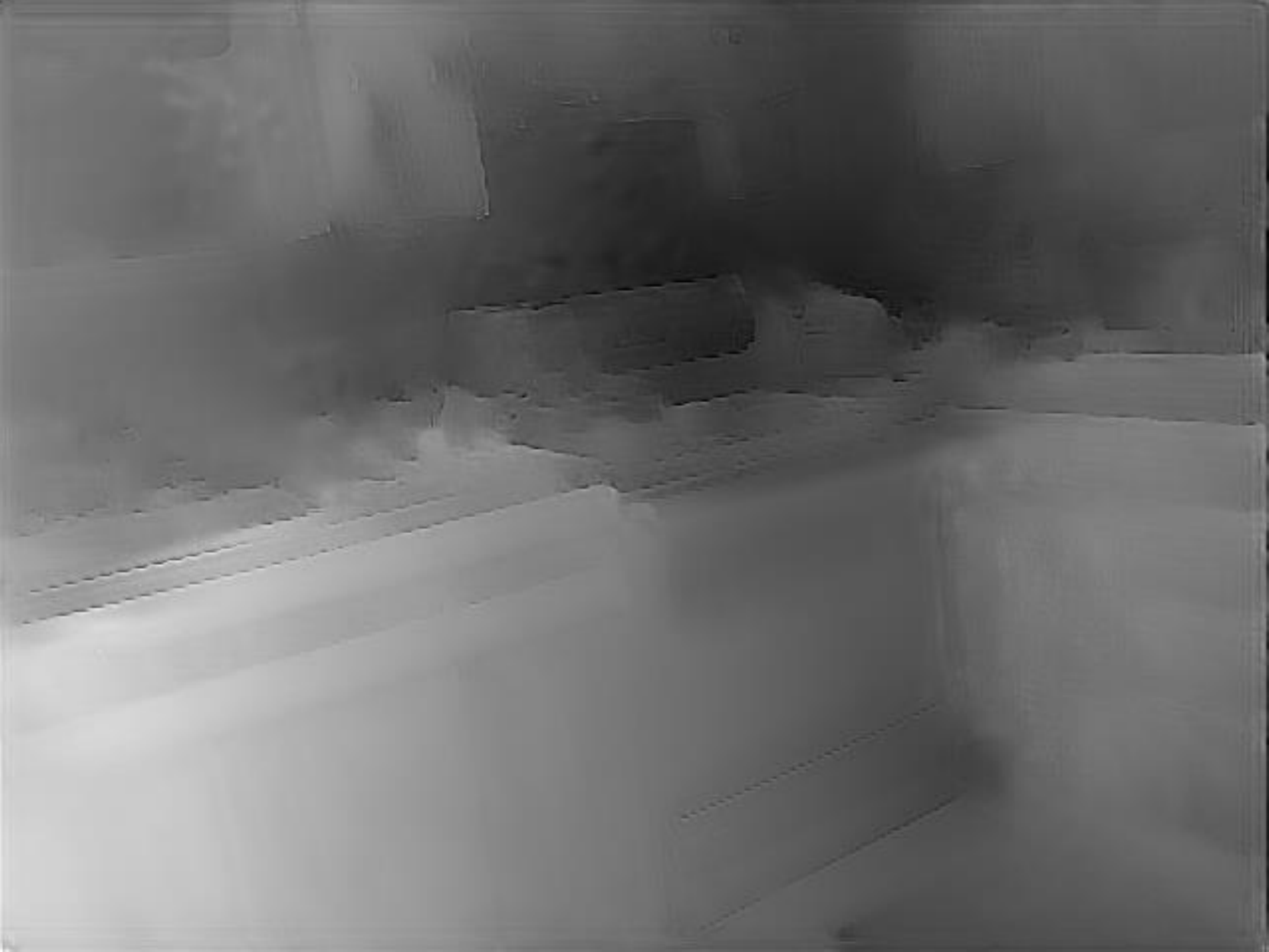}}
		{\includegraphics[height=0.14\linewidth]{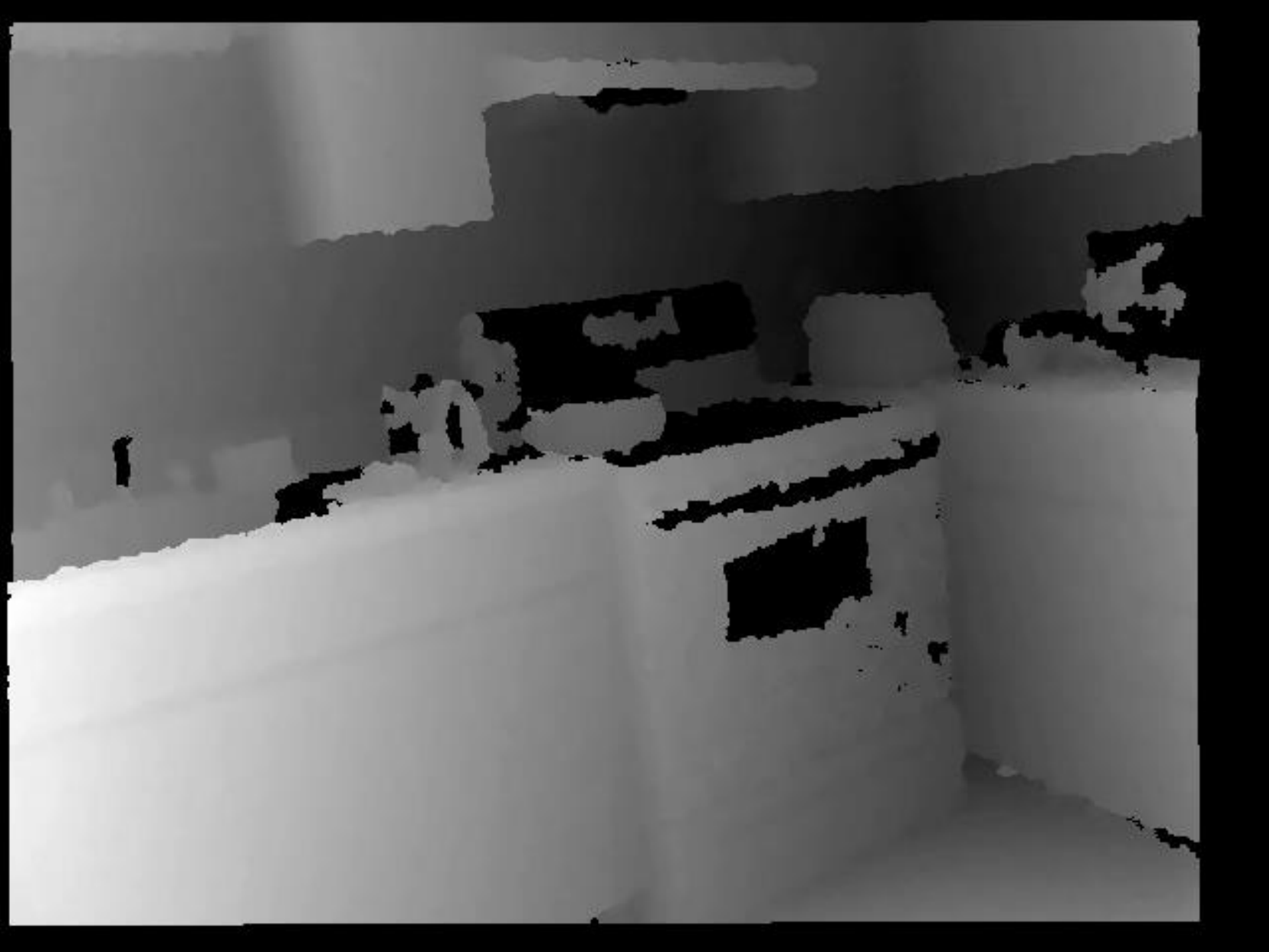}}  \\
		{\includegraphics[height=0.14\linewidth]{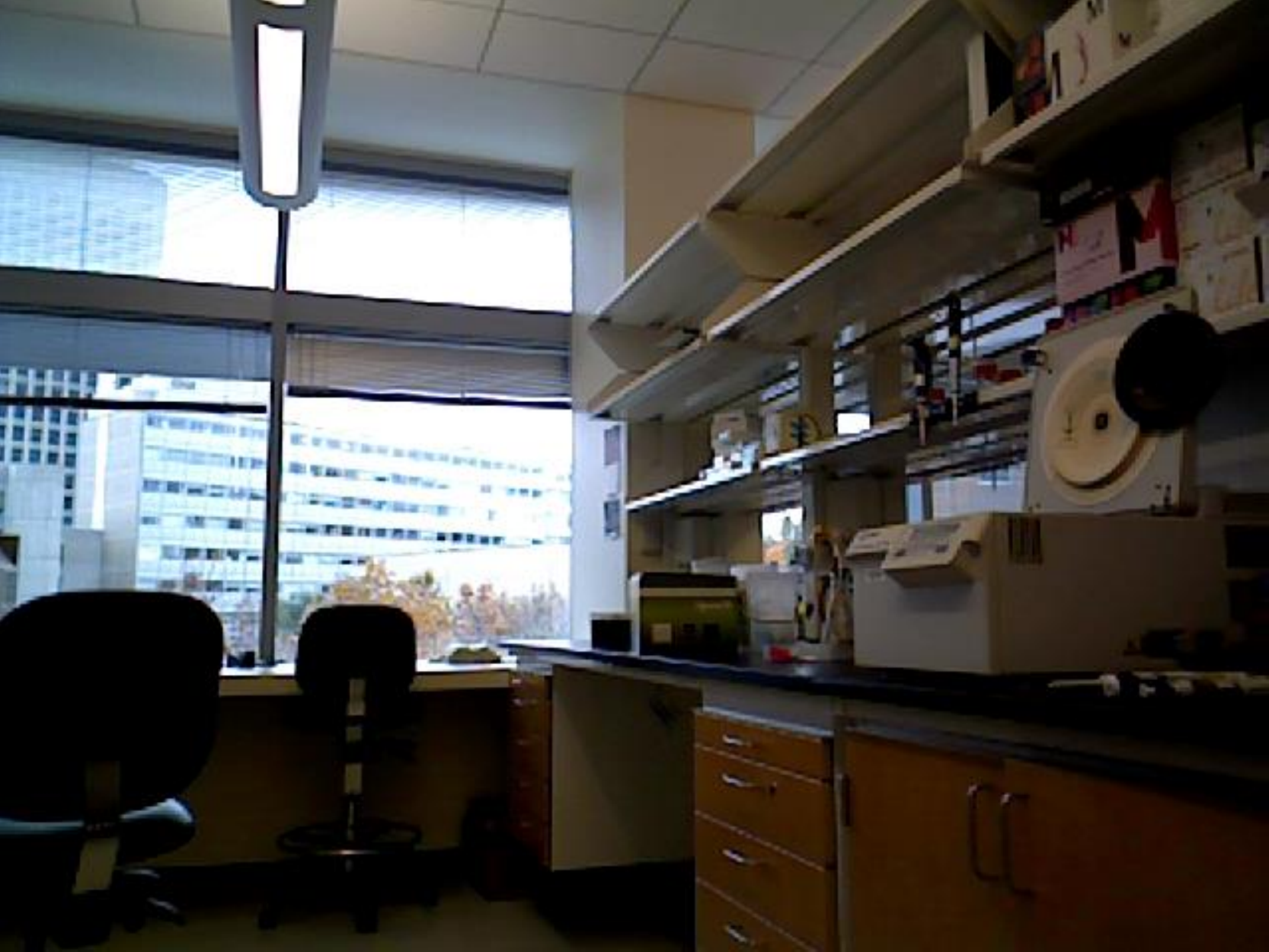}}
		{\includegraphics[height=0.14\linewidth]{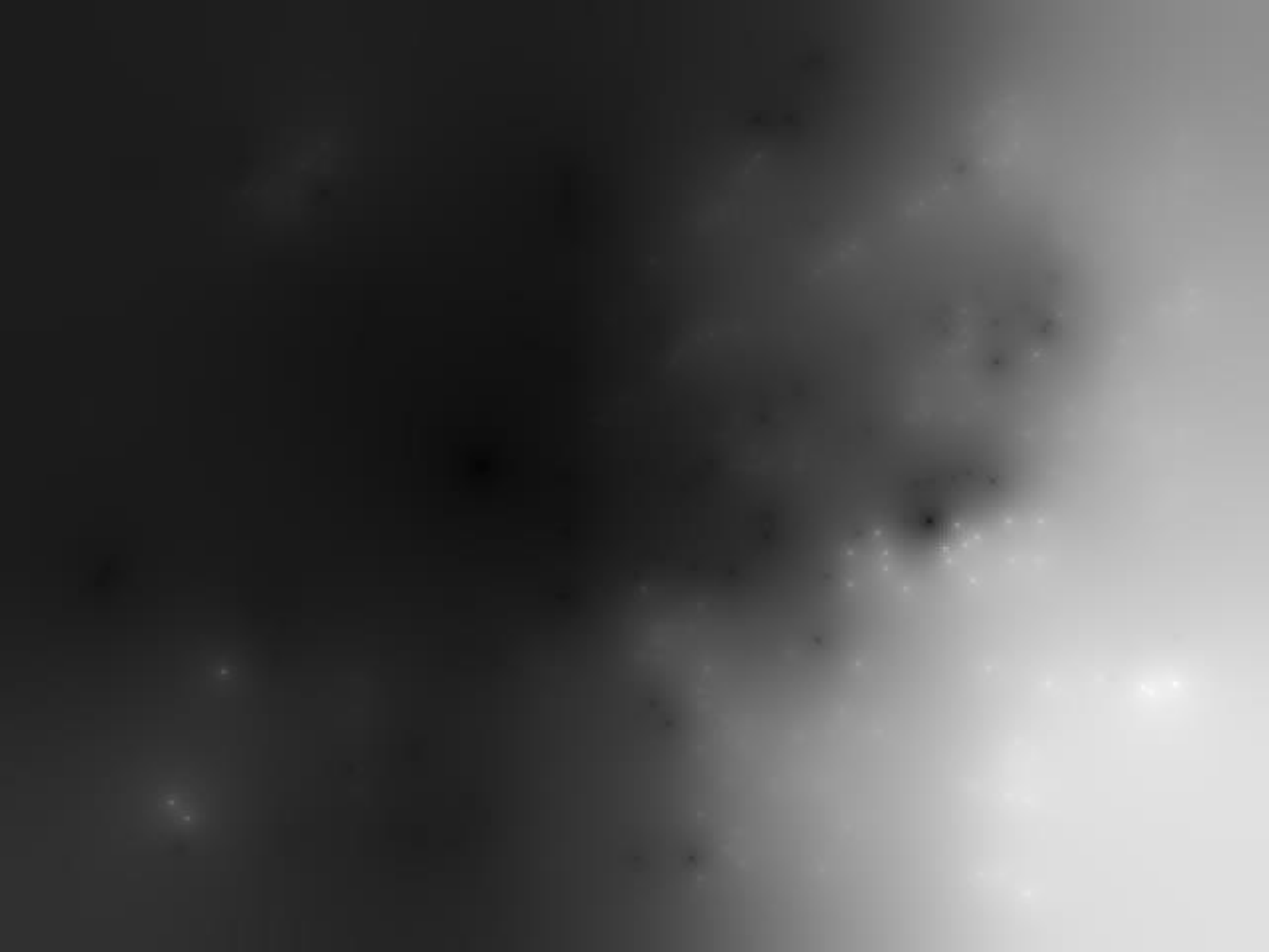}}
		{\includegraphics[height=0.14\linewidth]{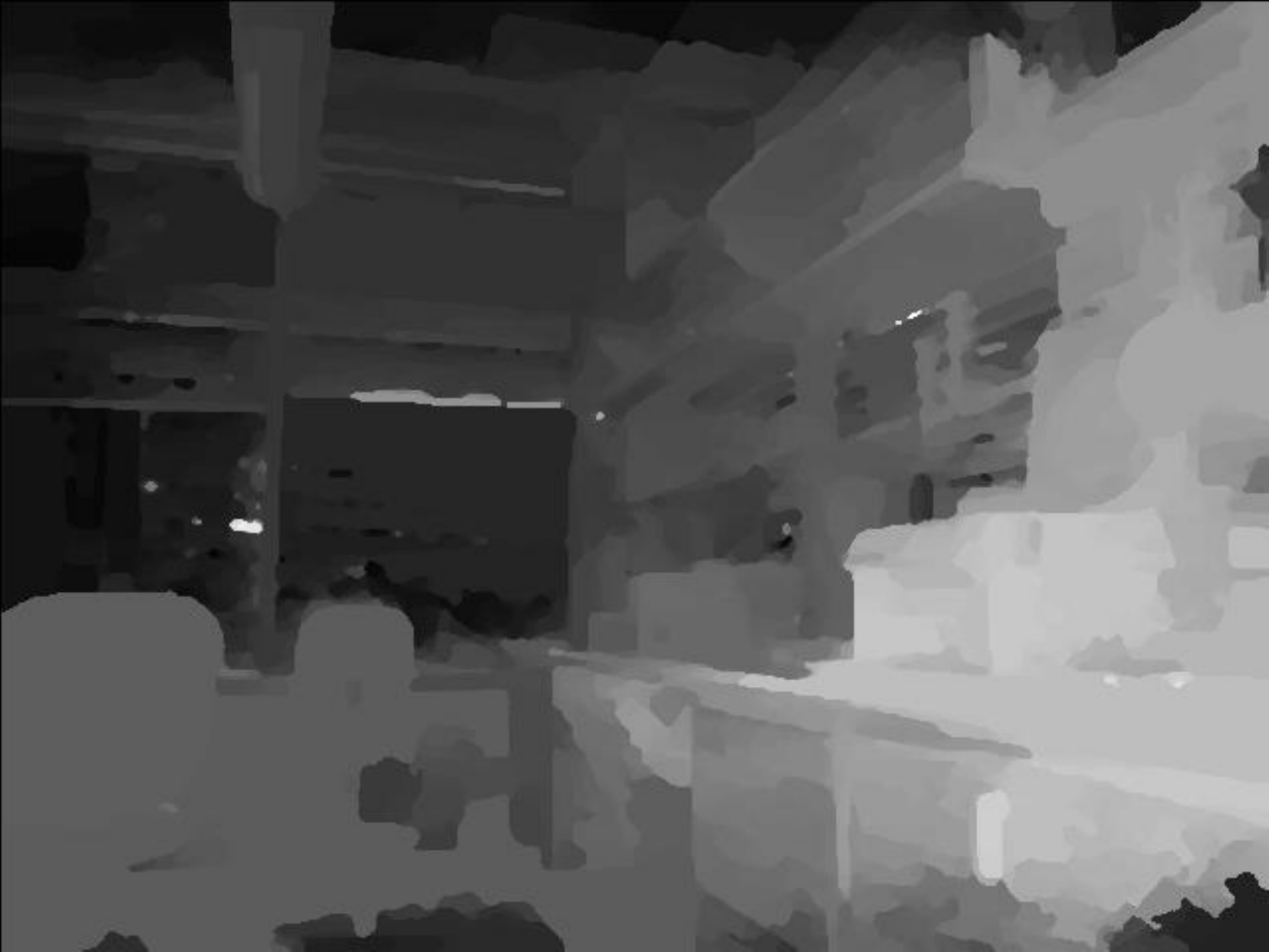}}
		{\includegraphics[height=0.14\linewidth]{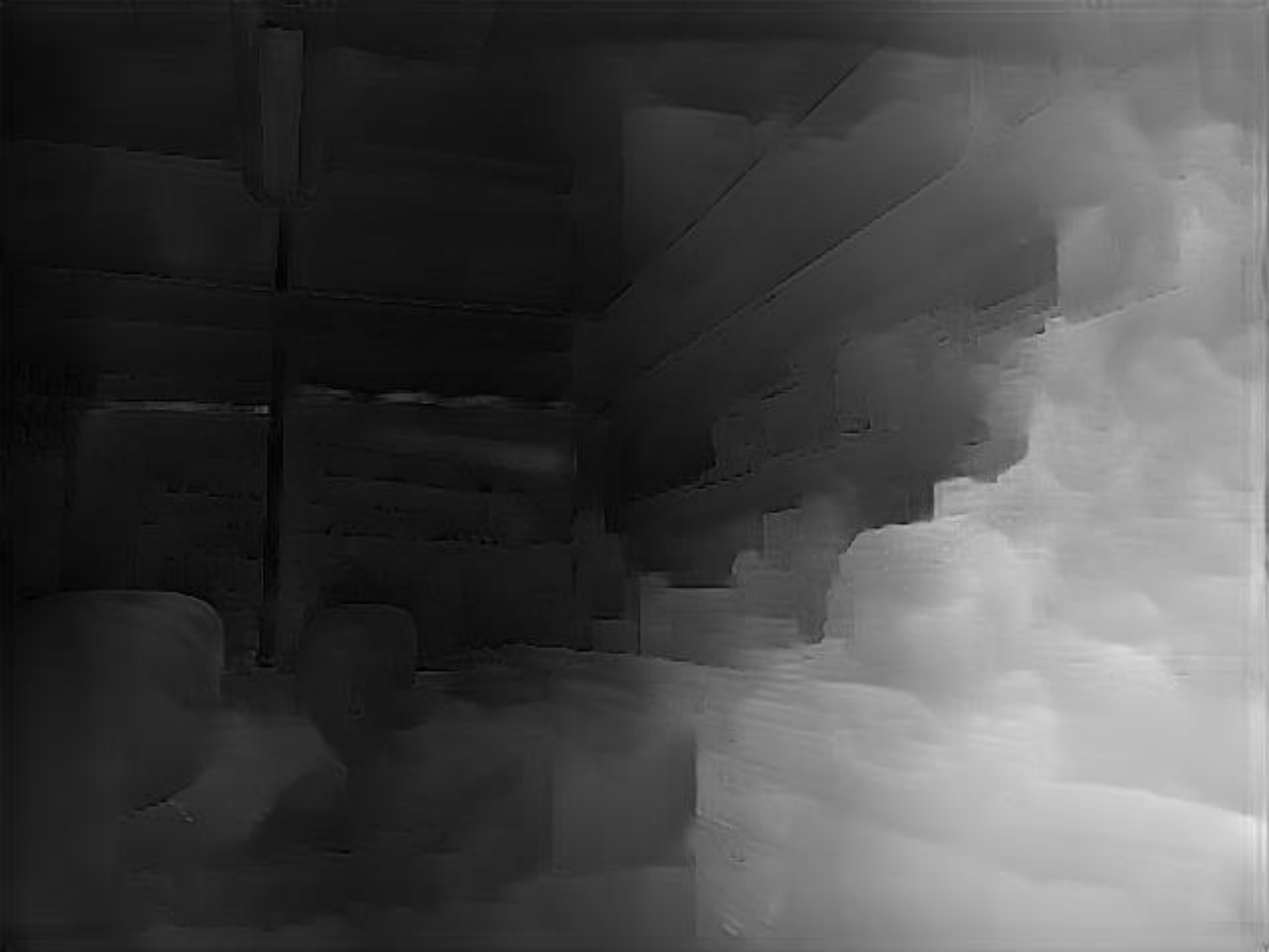}}
		{\includegraphics[height=0.14\linewidth]{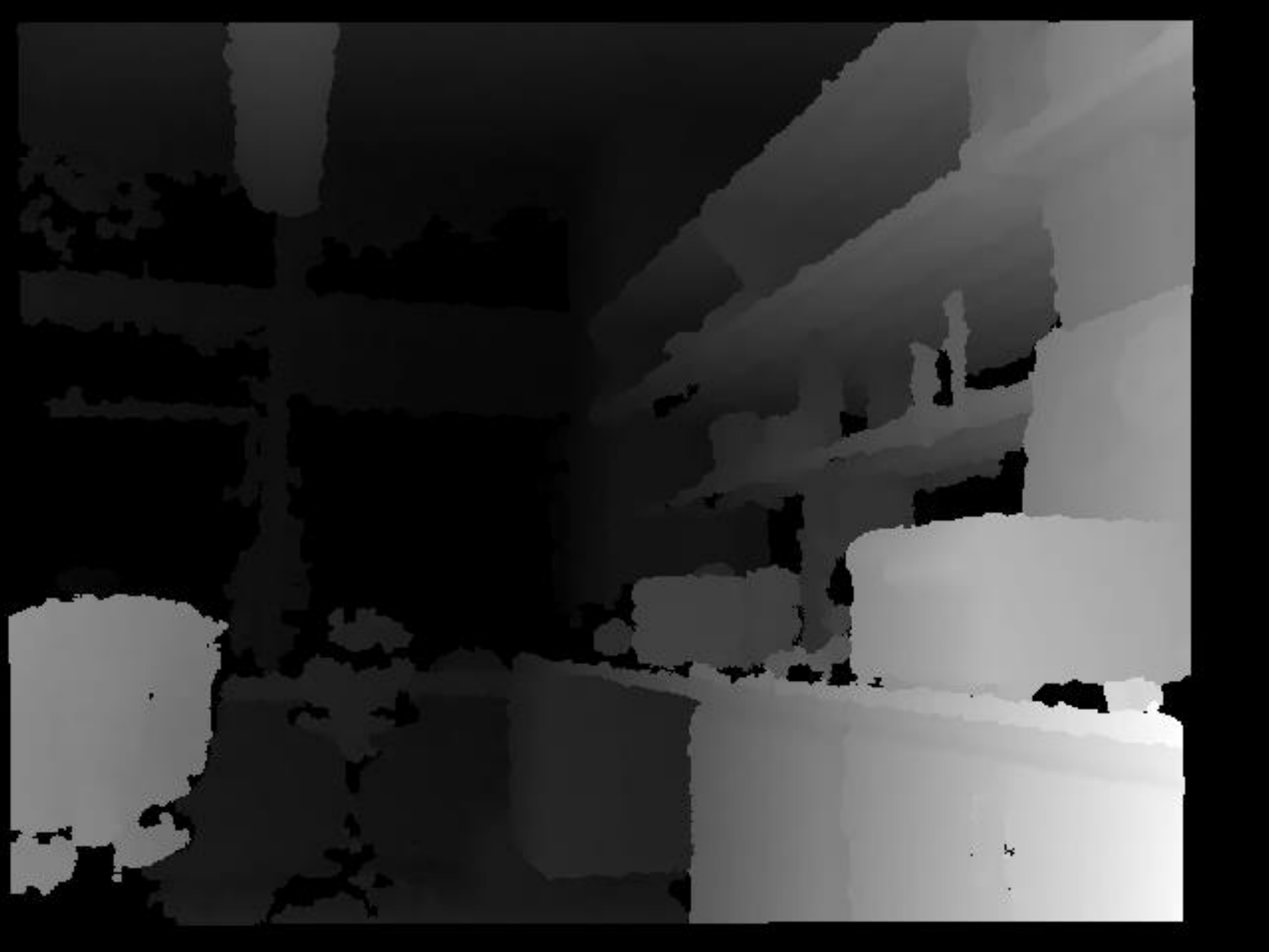}}  \\
		{\includegraphics[height=0.14\linewidth]{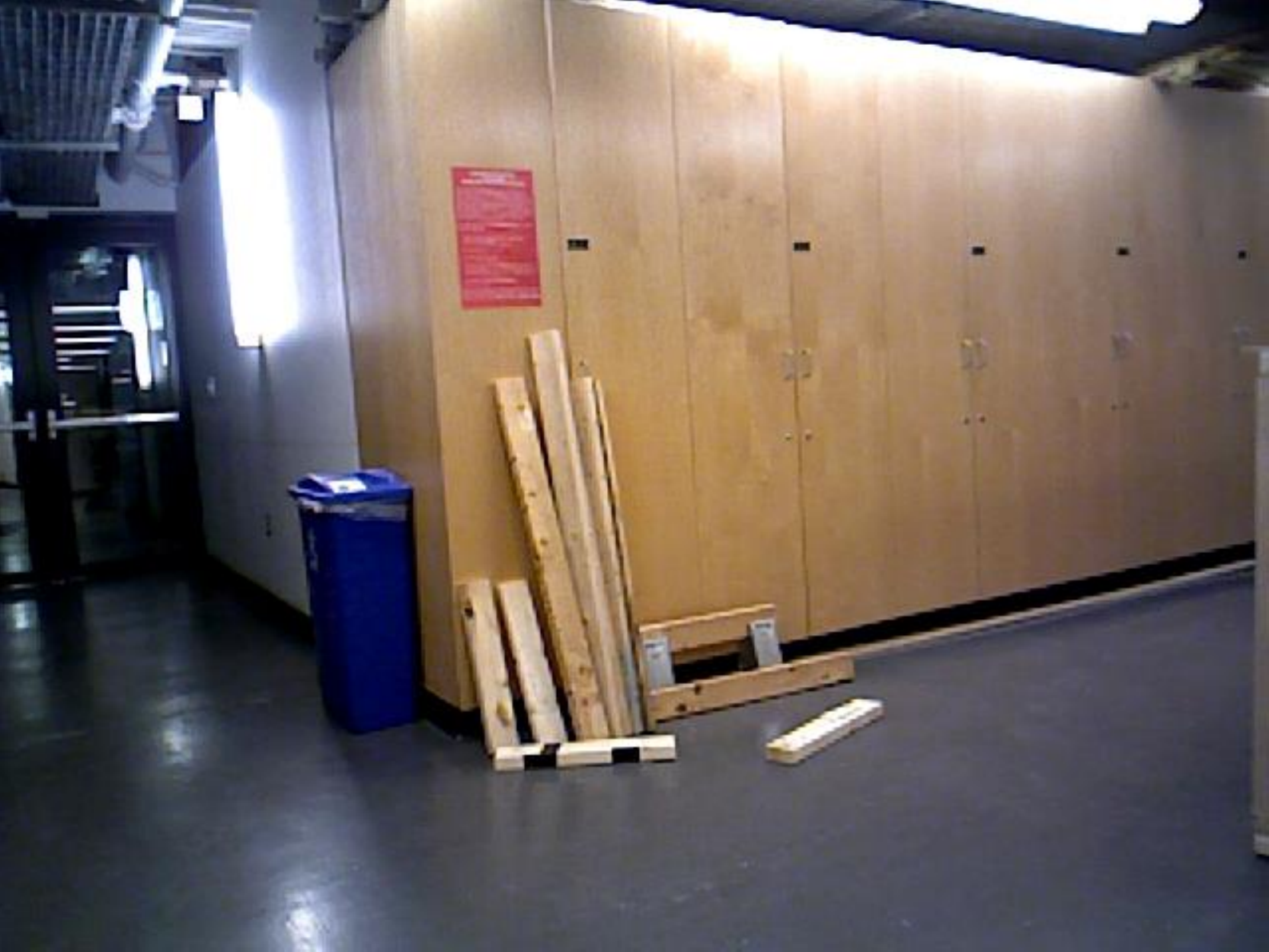}}
		{\includegraphics[height=0.14\linewidth]{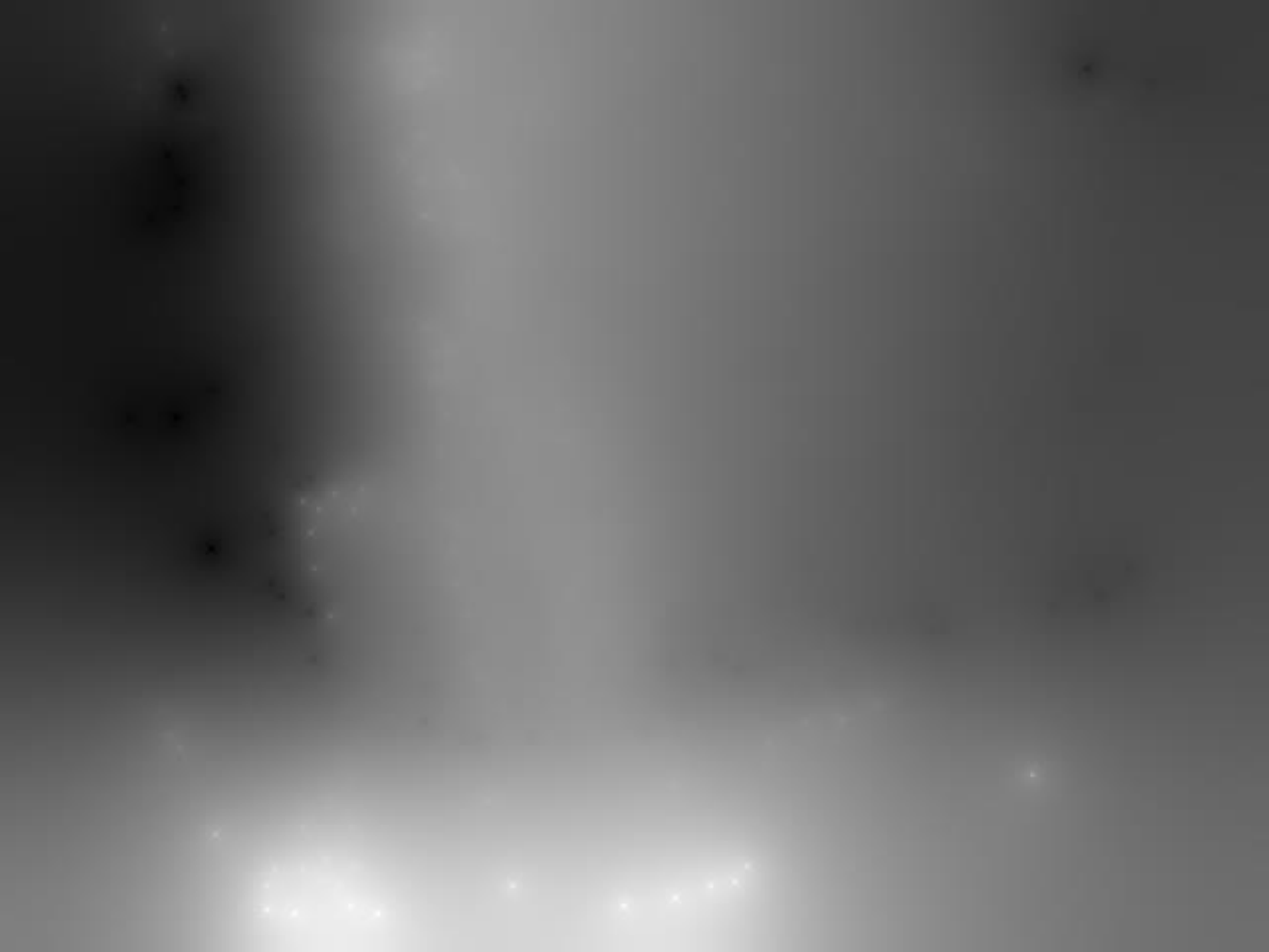}}
		{\includegraphics[height=0.14\linewidth]{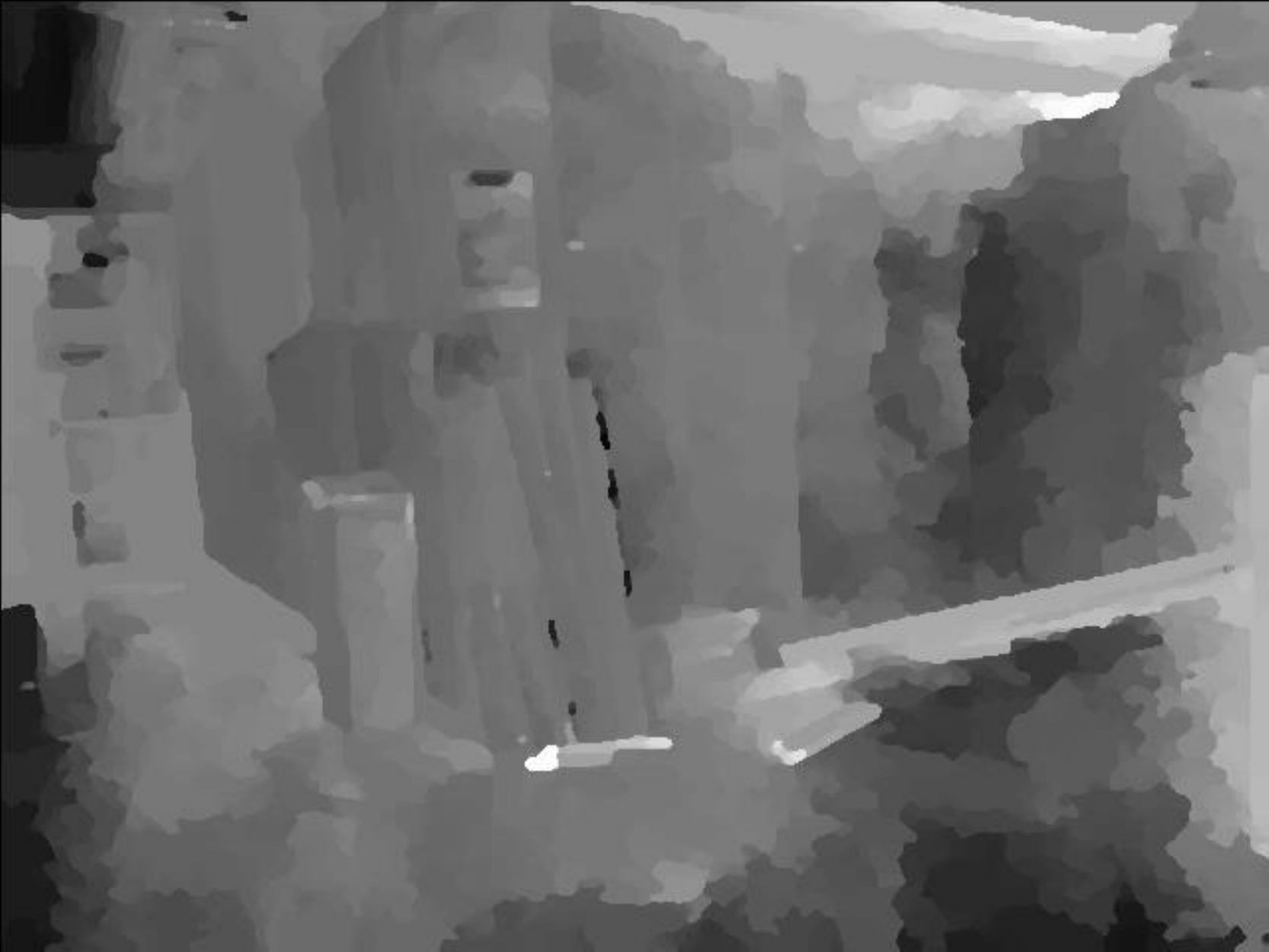}}
		{\includegraphics[height=0.14\linewidth]{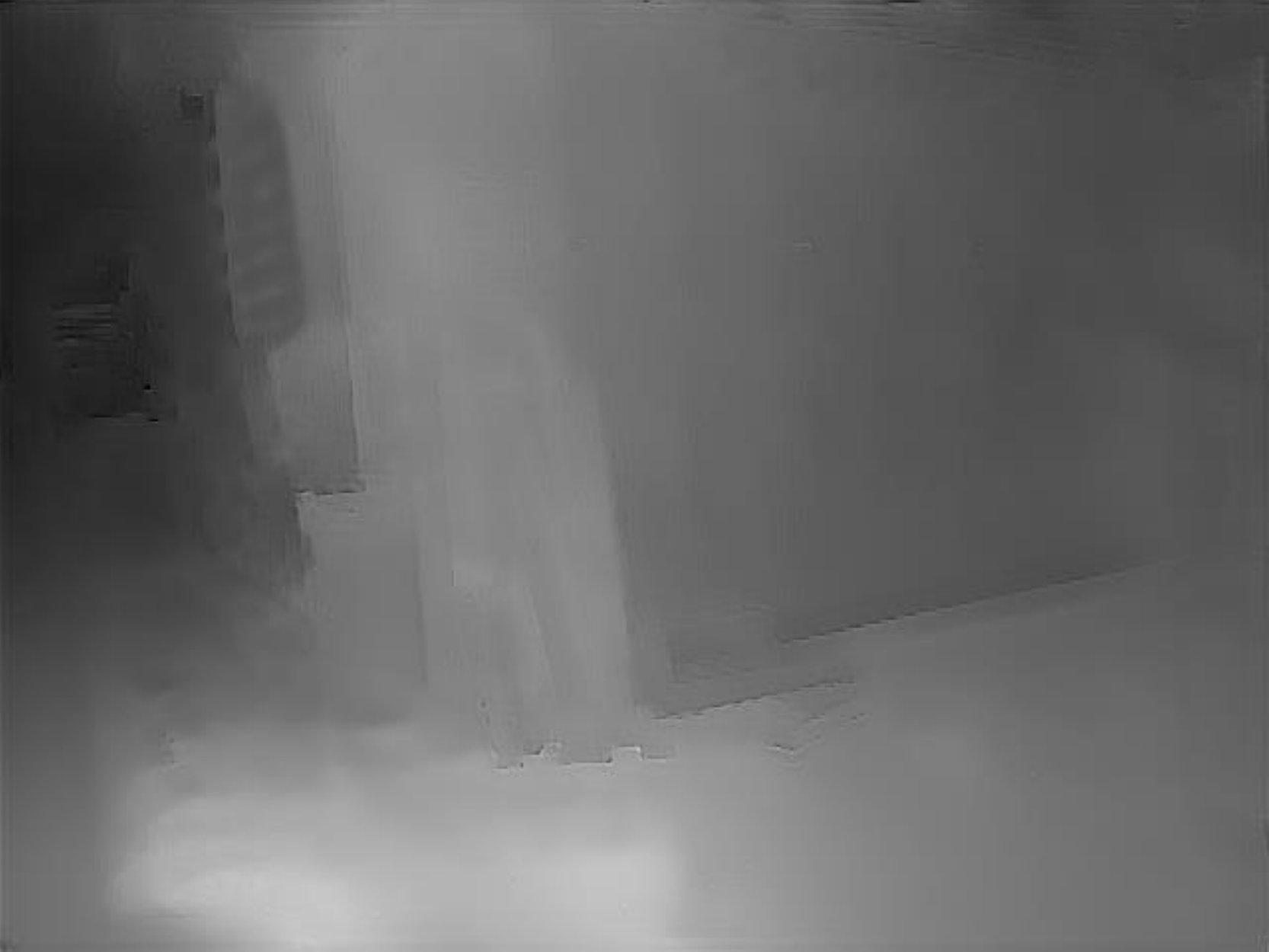}}
		{\includegraphics[height=0.14\linewidth]{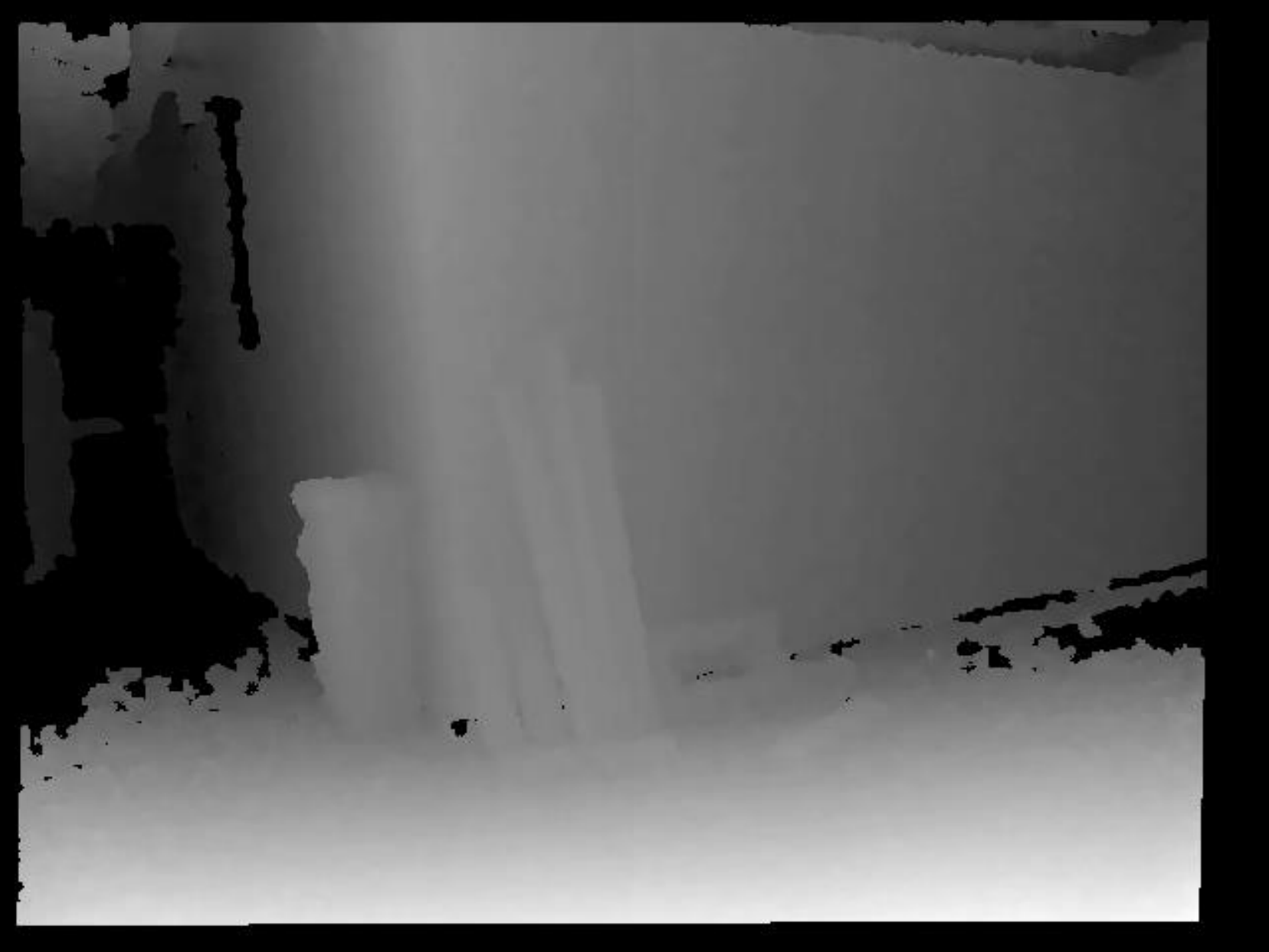}} \\
		\subcaptionbox{\label{sun_input} Reference images}{\includegraphics[height=0.14\linewidth]{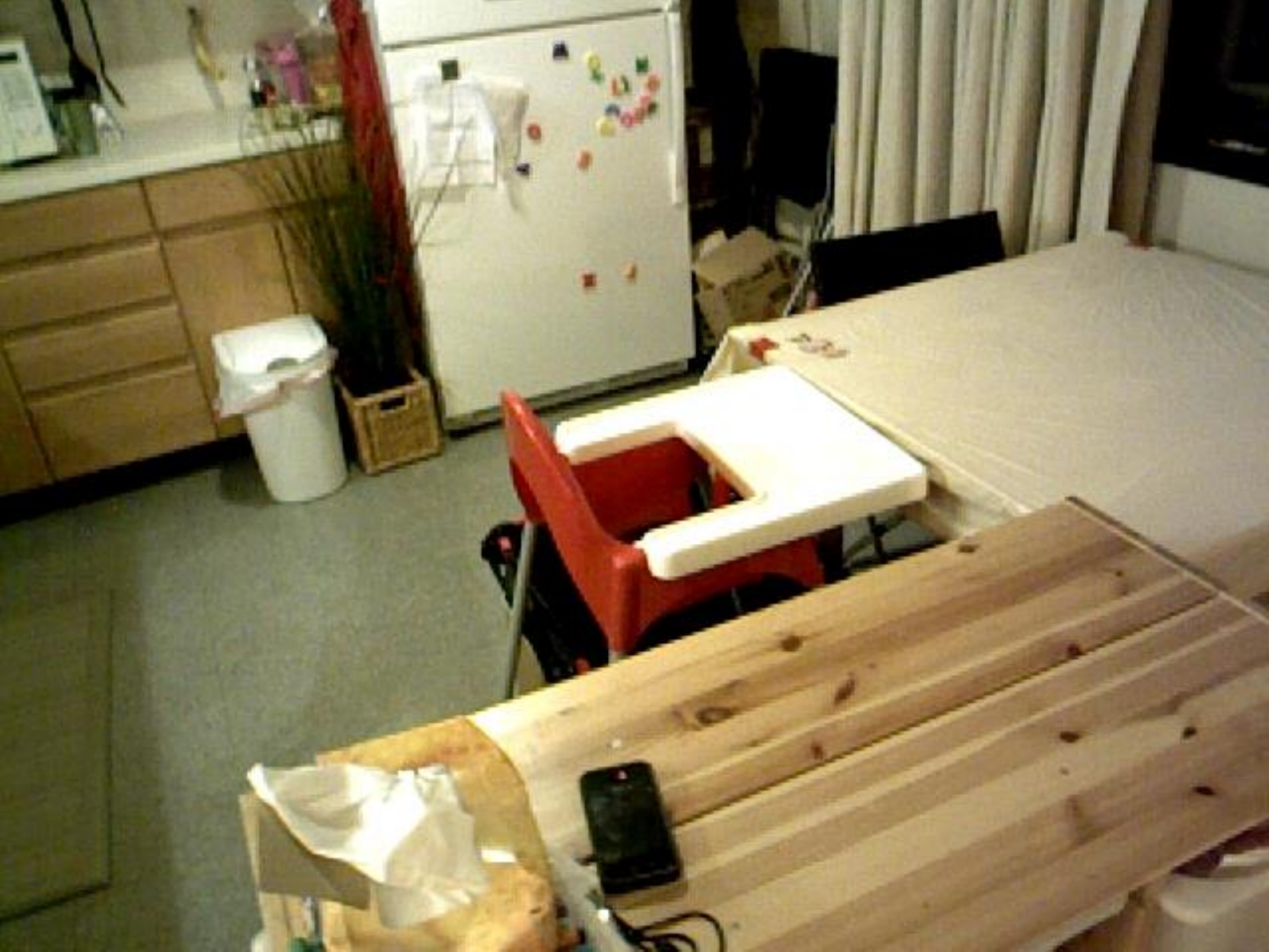}}	
		\subcaptionbox{\label{sun_iccv} Im \etal~\cite{Im15}}{\includegraphics[height=0.14\linewidth]{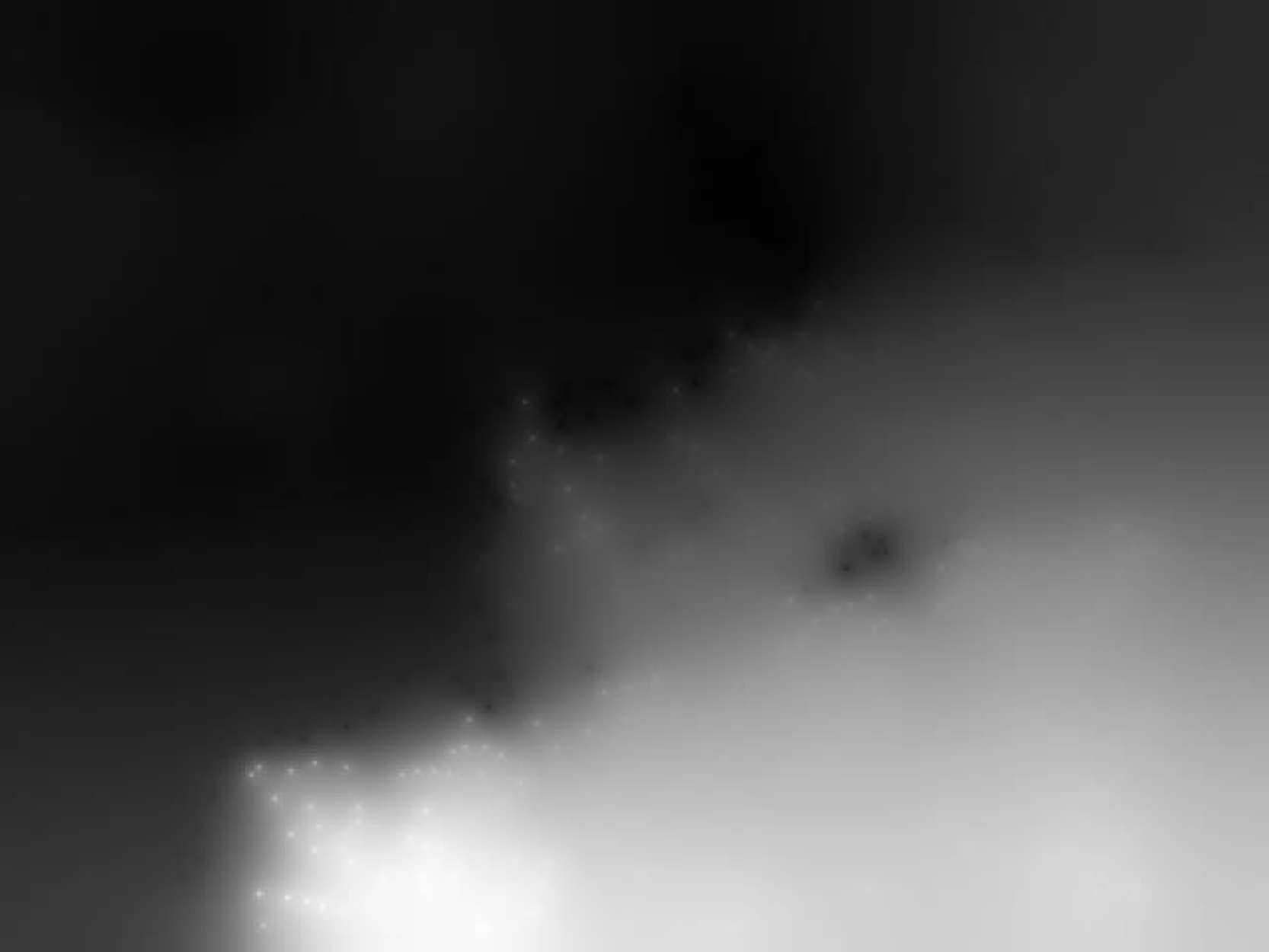}}	
		\subcaptionbox{\label{sun_cvpr} Ha \etal~\cite{Ha16} }{\includegraphics[height=0.14\linewidth]{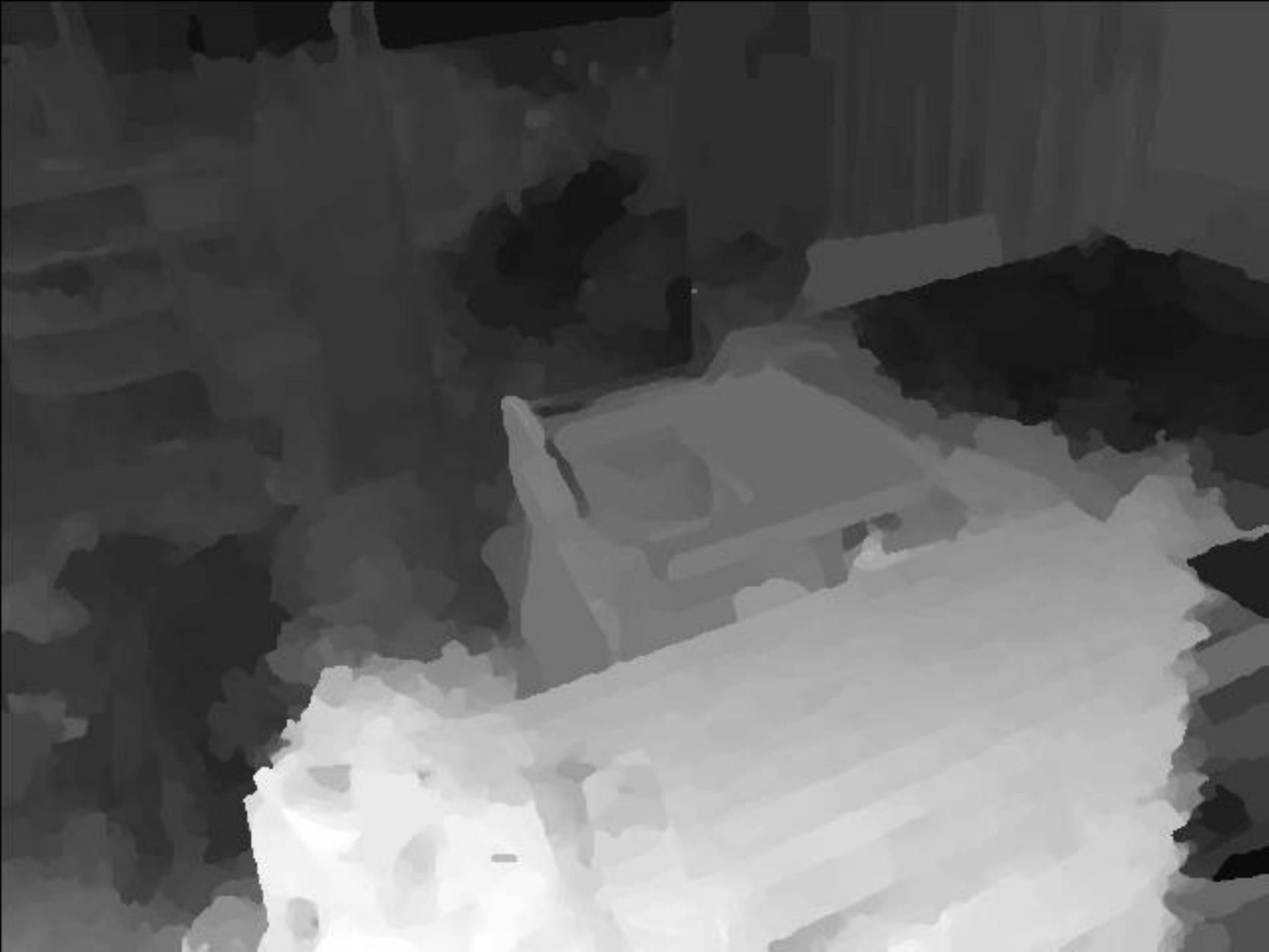}}	
		\subcaptionbox{\label{sun_oursf} Our depths}{\includegraphics[height=0.14\linewidth]{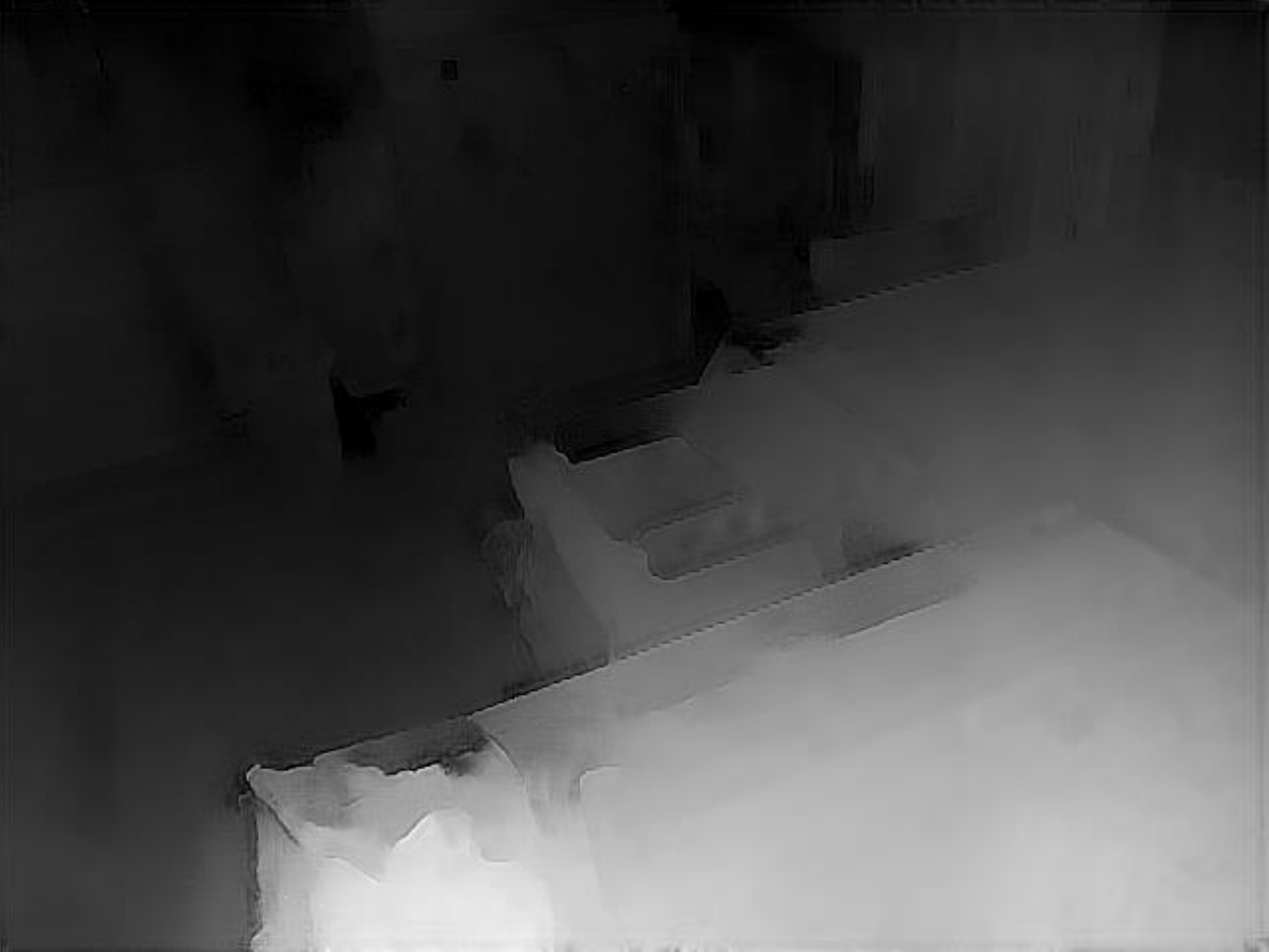}}
		\subcaptionbox{\label{sun_gt} Ground truth}{\includegraphics[height=0.14\linewidth]{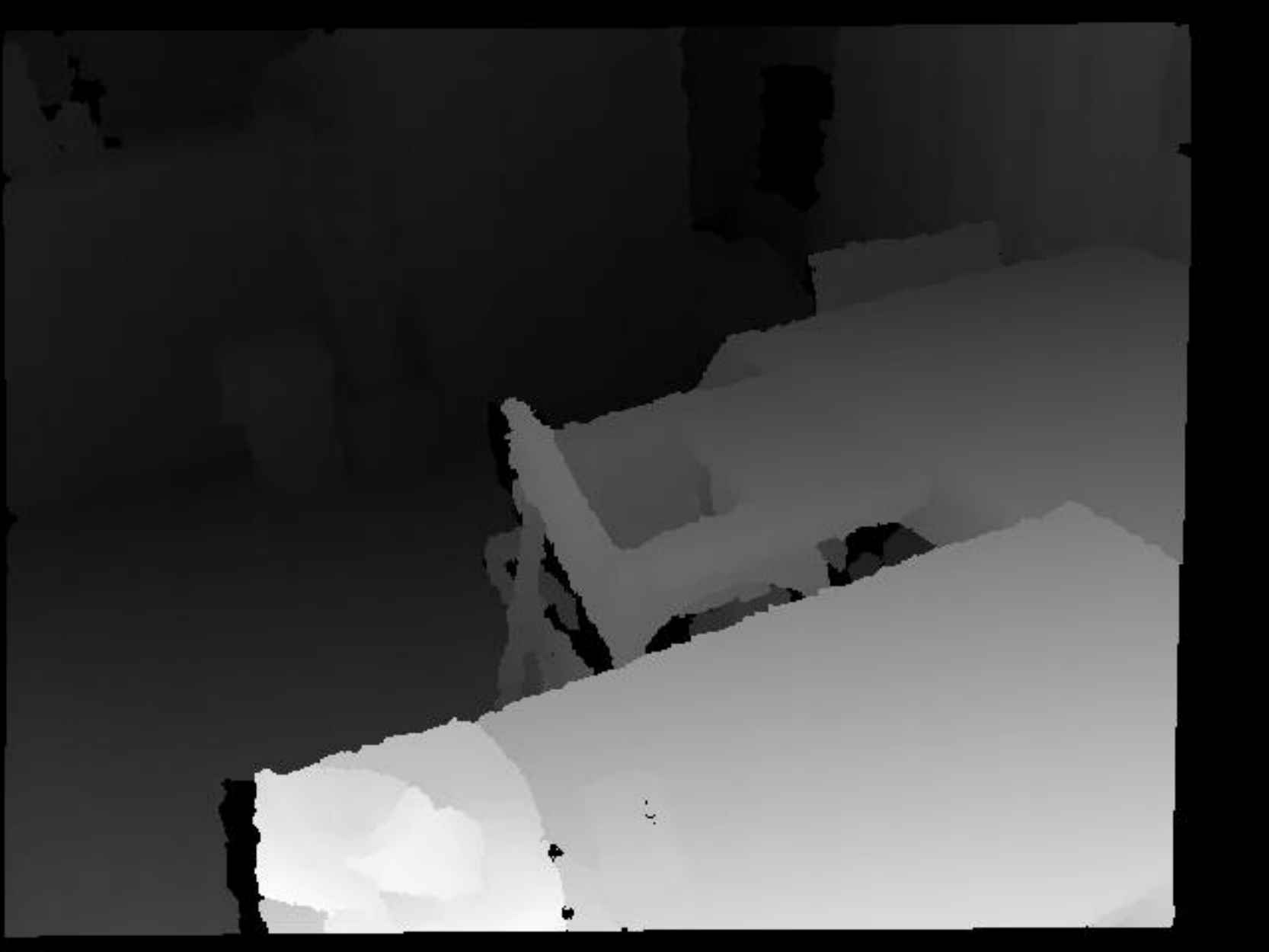}}	
	\end{tabular}
	\caption{Depth map results using SUN3D datasets~\protect\cite{xiao2013sun3d}. (a) Reference images. (b) Depth maps from propagation~\protect\cite{Im15}. (c) Depth maps from plane sweeping~\protect\cite{Ha16}. (d) Our depth maps. (e) Kinect depth maps. \textit{mit\_w85k1}, \textit{mit\_lab\_koch}, \textit{mit\_lab\_16} and \textit{mit\_w85h} (top-to-bottom).}
	\label{fig:sun}
\end{figure*}			

\subsection{Image alignment}
\label{sec:alignment}
Using the camera geometry $\mathbf{K}$, $[\mathbf{R}|\mathbf{t}]$ and scene geometry $\mathbf{z}$ estimated in~\secref{sec:DMVS}, we can simply align all images.
The aligned images $\tilde{I}_i$ where the original image $\hat{I}_i$ appears to have been taken at the reference view point are formulated as:
\begin{gather}
\begin{split}
\label{eq:alignment}
\tilde{I}_i(\mathbf{u})=\hat{I}_i\big(\big \langle\mathbf{K}[\mathbf{R}_i|\mathbf{t}_i]\begin{bmatrix} \mathbf{x}_{1}\mathbf{z}\\ 1 \end{bmatrix} \big\rangle\big), i\in \{1,...,n\}.\\
\end{split}
\end{gather}
We use a bicubic interpolation in this warping process. 
The aligned images can be used for image quality enhancement applications such as noise reduction and exposure fusion as shown in~\figref{fig:final}. Using the estimated depth in~\figref{final_d}, we warp all non-reference images in~\figref{final_ai} into the reference view point.
After aligning the images, we use simple weighted averaging method~\cite{Liu14} for denoising in~\figref{final_dn} and exposure fusion algorithm~\cite{mertens2009exposure} in~\figref{final_ef}.
The results show that our estimated depth and pose can precisely align the input images, which is applicable for image quality enhancement.

%

\section{Experimental Results}
In this section, we demonstrate the effectiveness and robustness of the proposed method using various experiments.
First of all, we compared our depth map results to those obtained from the state-of-the-art DfSM methods~\cite{Im15,Ha16}. In quantitative evaluation, we generated synthetic noisy images from the public RGB-D datasets~\cite{xiao2013sun3d} and utilized them as the input.
We then demonstrate that our method produces accurate depth with varying exposure image sequences captured by the bracketing mode. Finally, we investigated the applicability of the depth results for depth-aware photographic application, as well as image quality enhancement.


All steps were implemented in MATLAB\texttrademark, except for the DNN part, which was implemented by Lua.
We set the random depth value to 100, with the constants $c_1$, $c_2$ and $\sigma_c$ as 1, 10 and 0.2, respectively.
On average, for an image sequence of 28 frames with 640$\times$480 resolution, our method took 4s in total for pose and depth estimation on an Intel i7 3.40GHz CPU and 16GB RAM.
The SfSM (including feature extraction and bundle adjustment) and the DMVS (including depth propagation and geometric transformation) required 2.5s and 1.5s, respectively.

\subsection{Synthetic datasets}
\label{sec:synthetic}

\noindent{\bf{Quantitative evaluation of our DMVS}} \quad 
\label{sec:quant}
We quantitatively compared our developed approach with the state-of-the-art DfSM methods~\cite{Im15,Ha16}, using public RGB-D datasets~\cite{xiao2013sun3d}.
For the datasets, Microsoft Kinect was used to capture the sequential images and the corresponding depth maps.
We used 28 consecutive frames for the comparison (previous works requires about 30 frames as input). Since the datasets are taken moving slowly at 30fps, the baseline of the input sequence is narrow enough for quantitative evaluation of DfSM. 
To simulate realistic camera noise, we applied a signal-dependent Gaussian noise~\cite{schechner2007multiplexing} with a standard deviation $\sigma$ of 0.02. The noise level was determined by averaging the computed noise levels~\cite{liu2013single} in low-light conditions using a Nexus6.

\figref{fig:sun} shows the depth maps from~\cite{Im15},~\cite{Ha16}, our method and Microsoft Kinect using the synthetic noisy sequence.
As shown in~\figref{sun_iccv}, \cite{Im15} fails to show promising results due to inaccurate initial matching cost and the dense depth reconstruction.
Work in~\cite{Ha16} shows relatively accurate depth discontinuity, but also yields inaccurate depths as shown in~\figref{sun_cvpr}.
This is because the plane sweeping algorithm using color similarity as a matching cost is not suitable for images with varying exposures, which produces an unreliable depth map.
On the other hand, our DNN-based approach in~\figref{sun_oursf} has the ability to handle the intensity changes, and to infer an accurate dense depth map, unlike~\cite{Im15,Ha16}.

\begin{figure}[t]
	\centering
	\begin{tabular}{c@{\hspace{1mm}}}
		\includegraphics[height=0.7\linewidth]{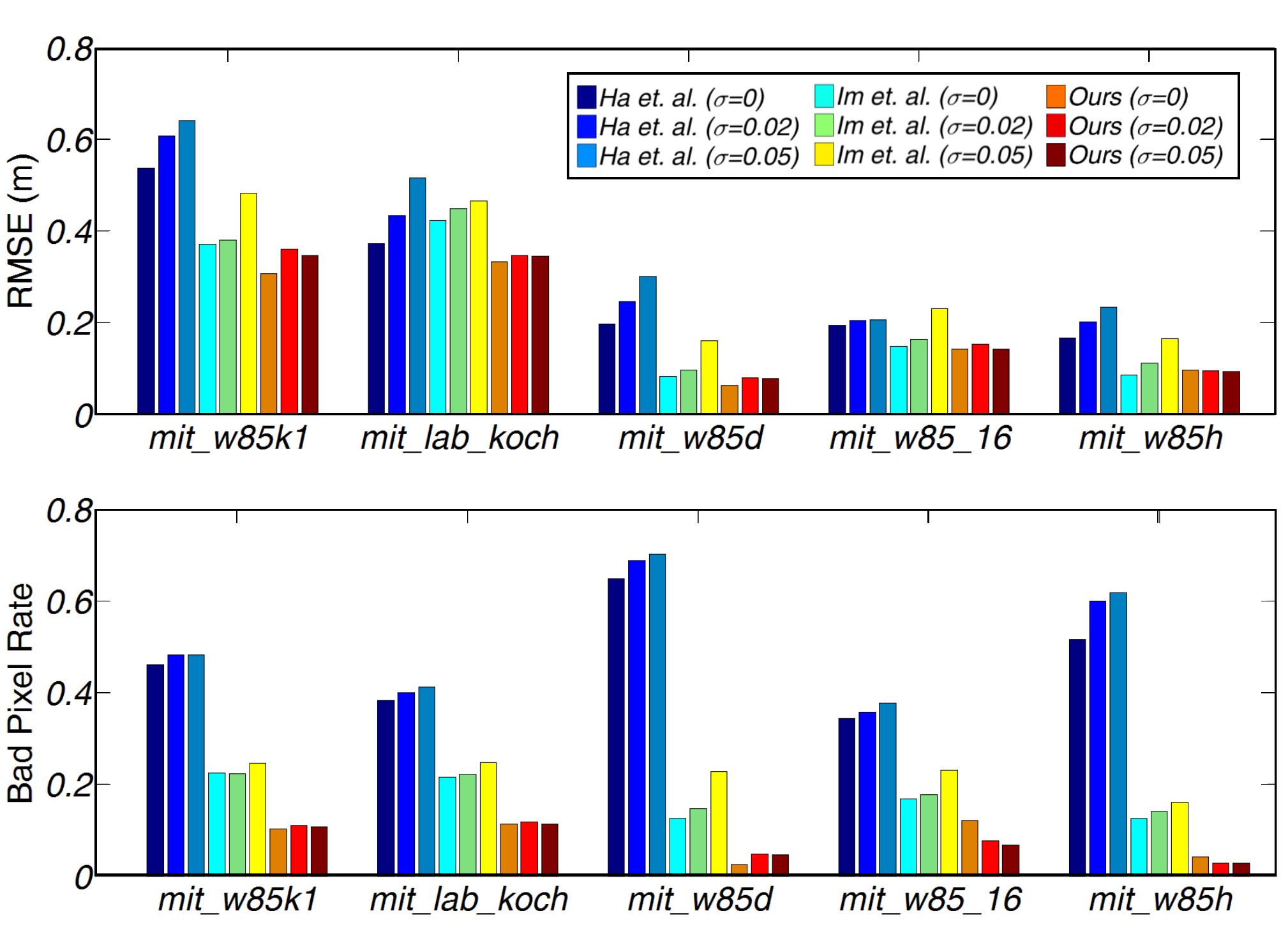}
	\end{tabular}
	\caption{Quantitative evaluation results with state-of-the-art DfSM methods.}
	\label{fig:RMSE}
\end{figure}

For a more detailed analysis, we measure a bad pixel rate and \textit{Root-mean-square-error} (RMSE) with varying noise levels ($\sigma=0,~ 0.02,~0.05$). \textit{Bad pixel rate} denotes the percentage of pixels that have a distance error of less than 10\% of the maximum depth value in the scene. We excluded the unmeasured depth regions due to the hardware limitations of Microsoft Kinect (dark areas in the Kinect depth maps in~\figref{sun_gt}) in the error measurement. The results of test across datasets in~\figref{fig:RMSE} shows our method has less RMSE and bad pixel rate than both the state-of-the-art methods for all noise levels.
We can see that the conventional methods give acceptable results when noise is not issue, but as noise increases, these measures degrade rapidly. Compared to the competing methods, our method achieved the best results regardless of noise levels, with the least degradation of performance.


\subsection{Real-world datasets}
\label{sec:realworld}

\noindent{\bf{Qualitative evaluation of our DMVS}} \quad
We designed a real-world experiment to verify that the proposed method could be applied to actual exposure bracketed images. 
First of all, we performed a qualitative comparison of DfSMs~\cite{Im15,Ha16} using exposure bracketing sequences. We took 28 frames with 7 exposure levels for one second in a commercial DSLR camera (Canon 1D Mark~\RN{3}). Since the state-of-the-art methods have not considered intensity changes, we equalized the histogram of all images to adjust image intensity and used them as an input of the methods. Raw images were used for our input. 

\begin{figure*}[t]
	\centering
	\begin{tabular}{c@{\hspace{1mm}}c@{\hspace{1mm}}c@{\hspace{1mm}}c@{\hspace{1mm}}}
		{\includegraphics[height=0.128\linewidth]{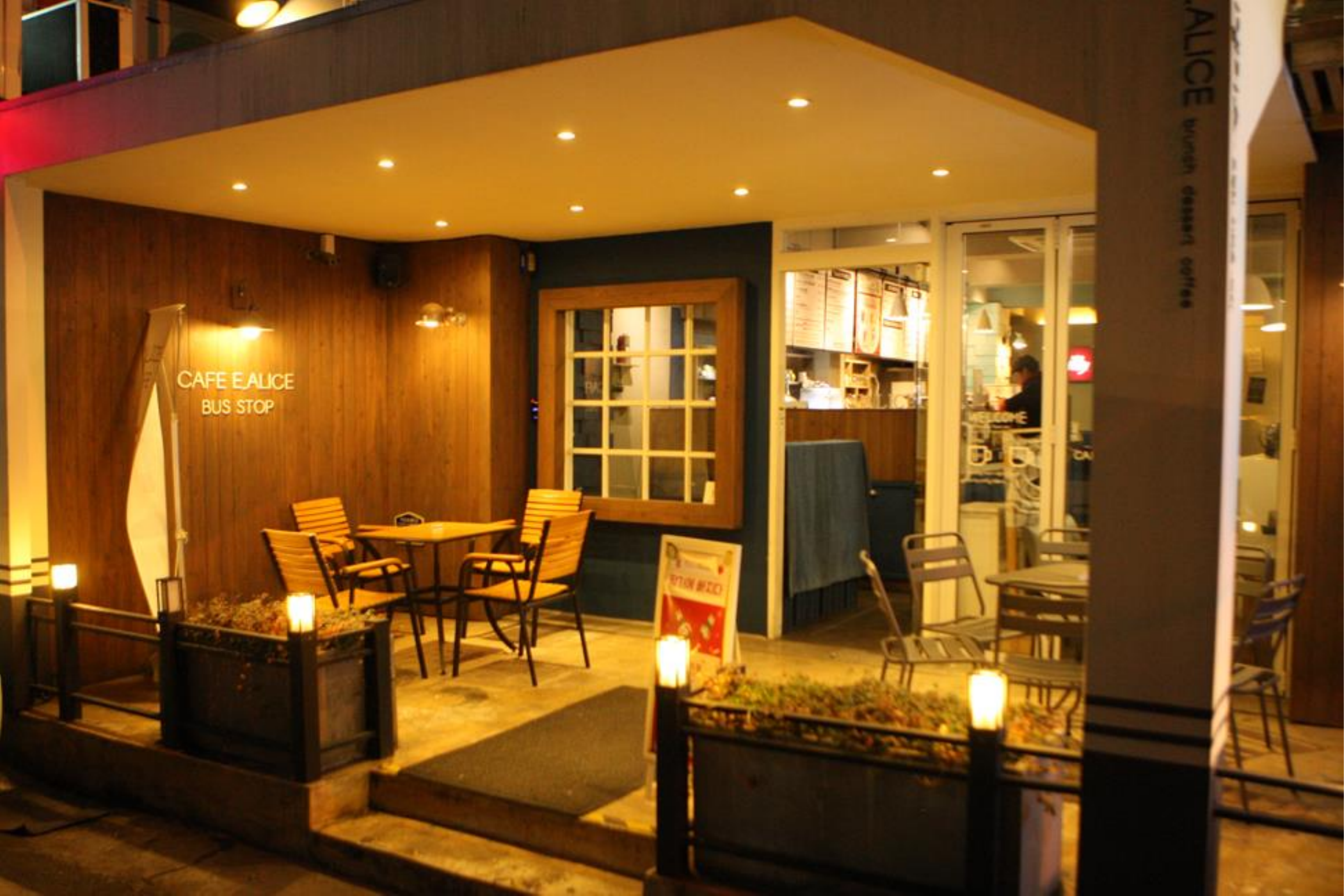}}
		{\includegraphics[height=0.128\linewidth]{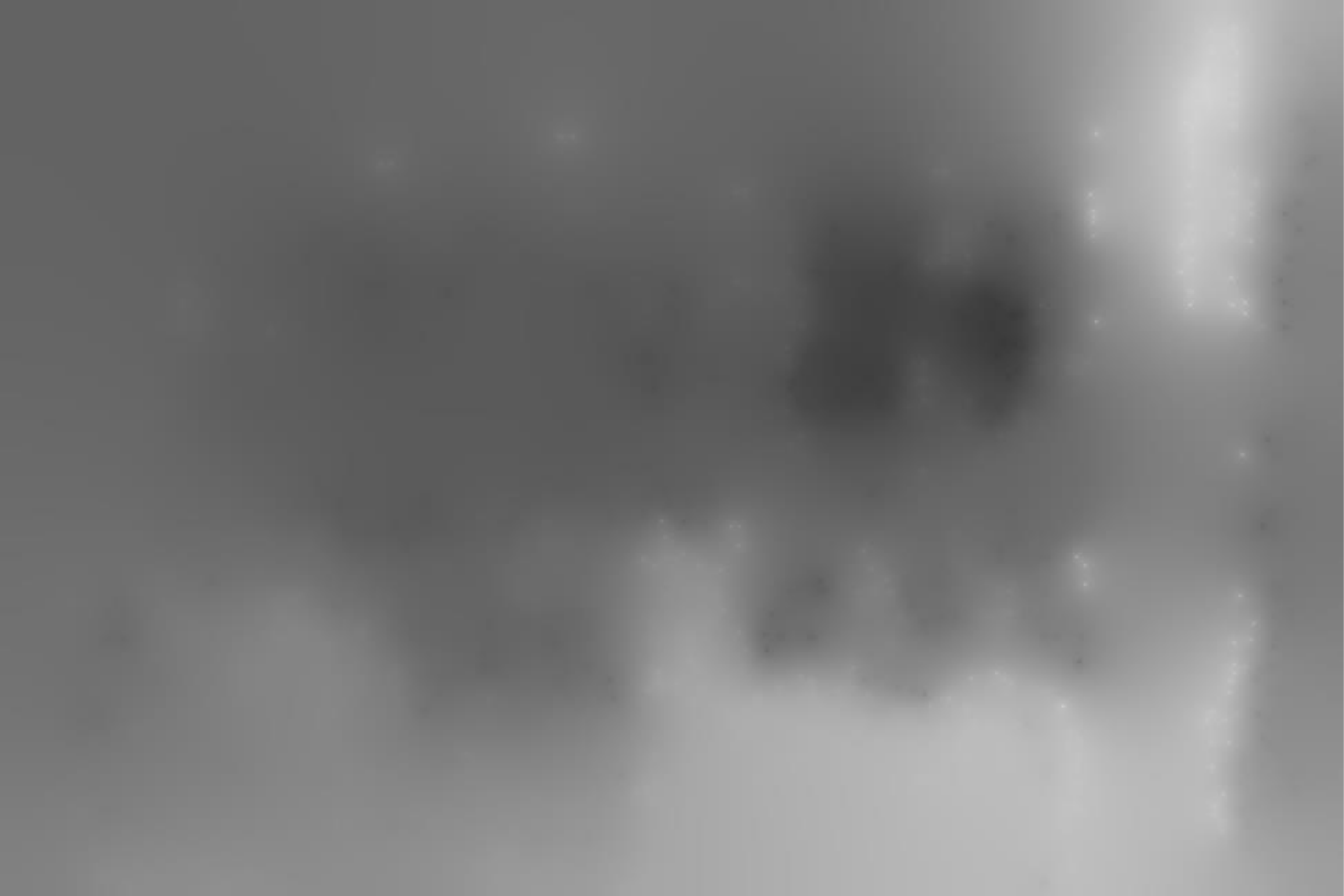}}
		{\includegraphics[height=0.128\linewidth]{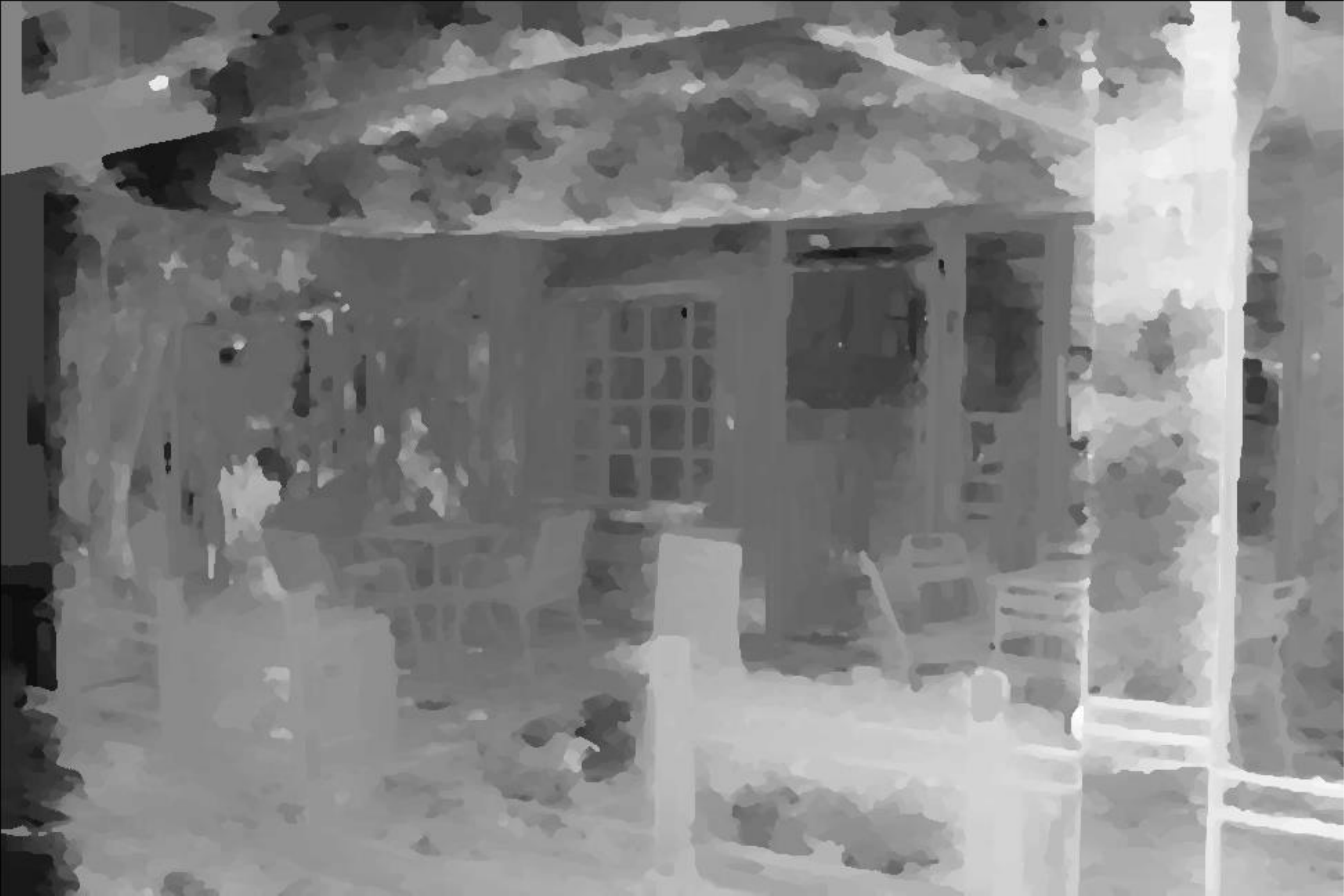}}
		{\includegraphics[height=0.128\linewidth]{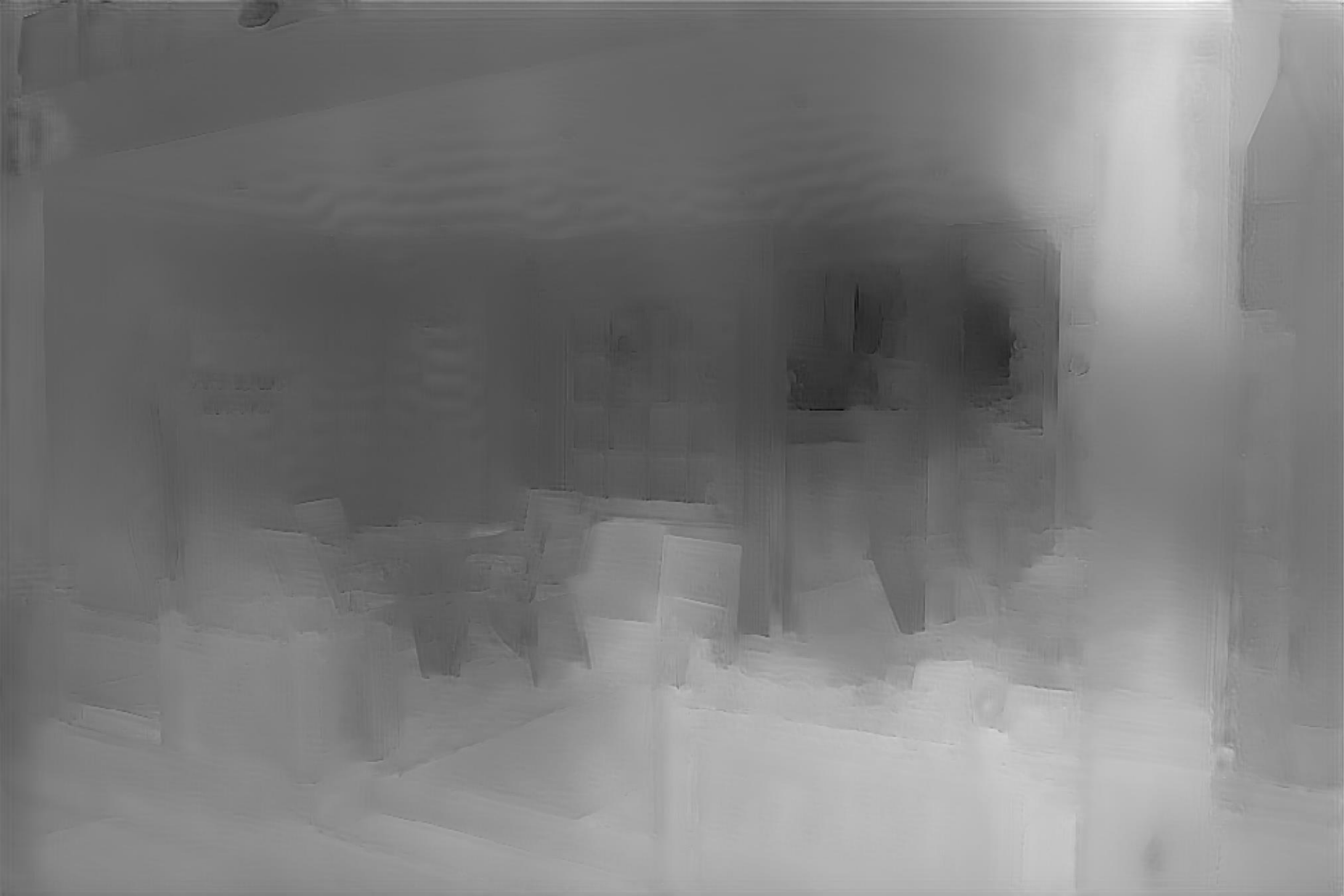}} \\
		\subcaptionbox{\label{real_input} Reference images}{\includegraphics[height=0.128\linewidth]{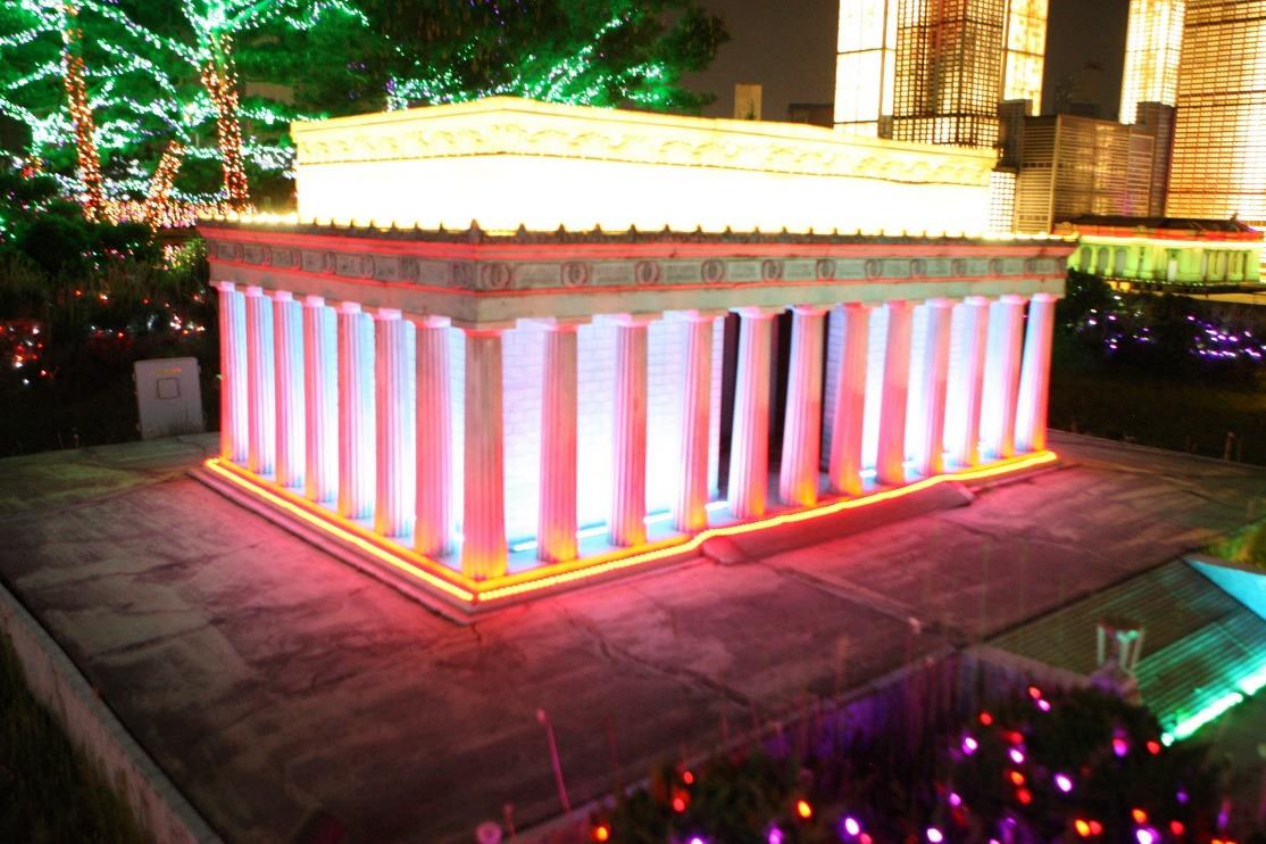}}
		\subcaptionbox{\label{real_iccv} Im \etal~\cite{Im15}}{\includegraphics[height=0.128\linewidth]{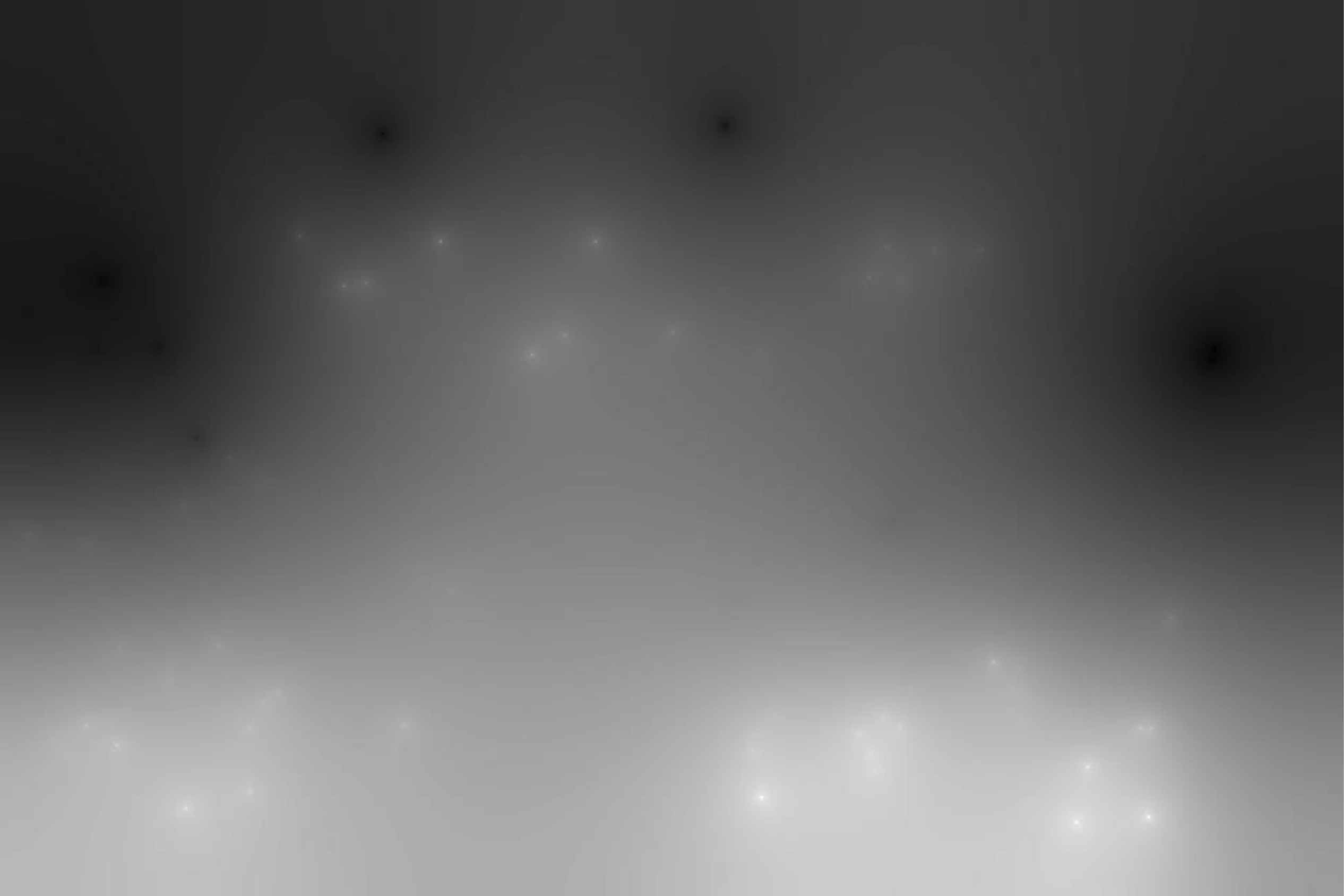}}	
		\subcaptionbox{\label{real_cvpr} Ha \etal~\cite{Ha16}}{\includegraphics[height=0.128\linewidth]{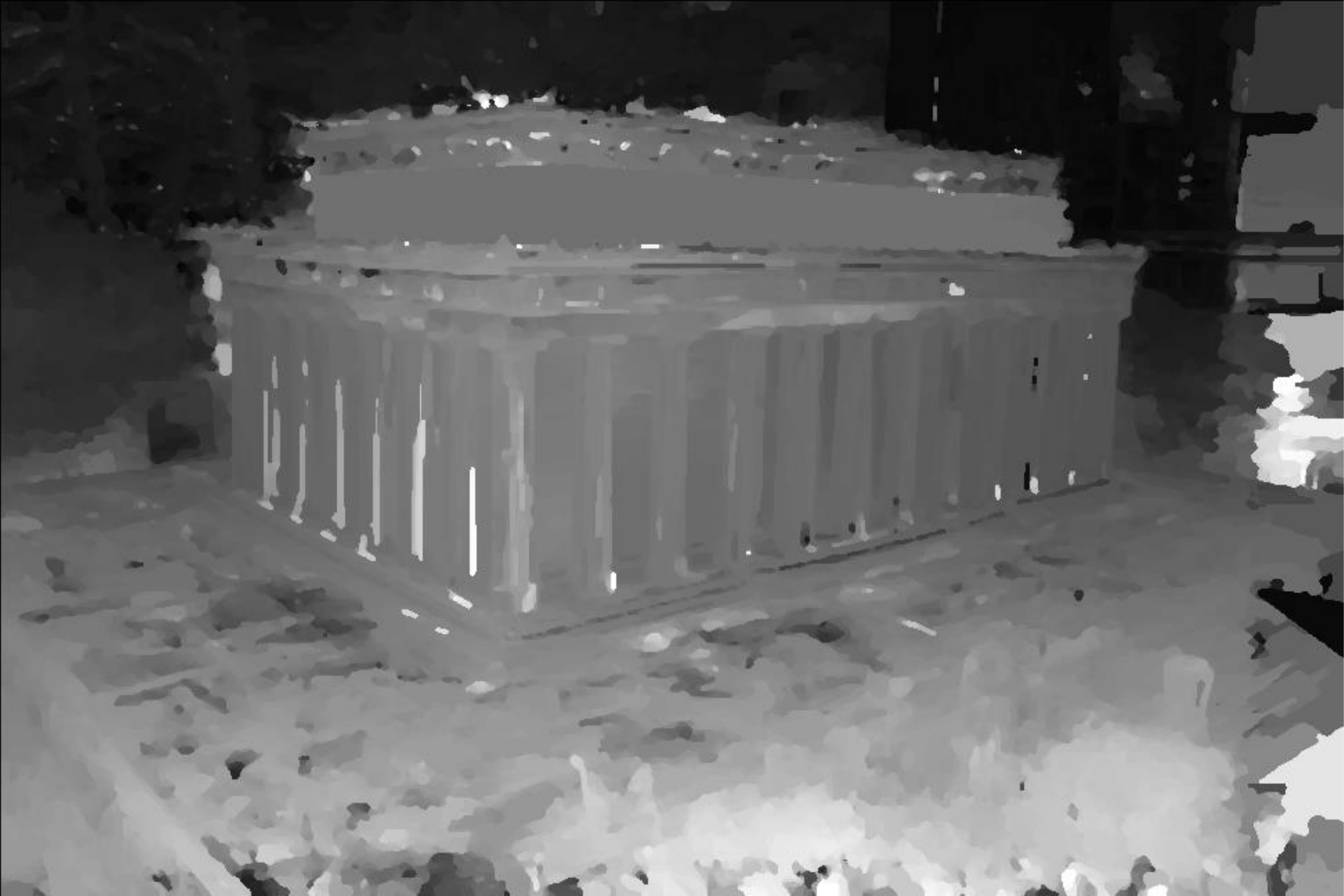}}	
		\subcaptionbox{\label{real_final} Our depths}{\includegraphics[height=0.128\linewidth]{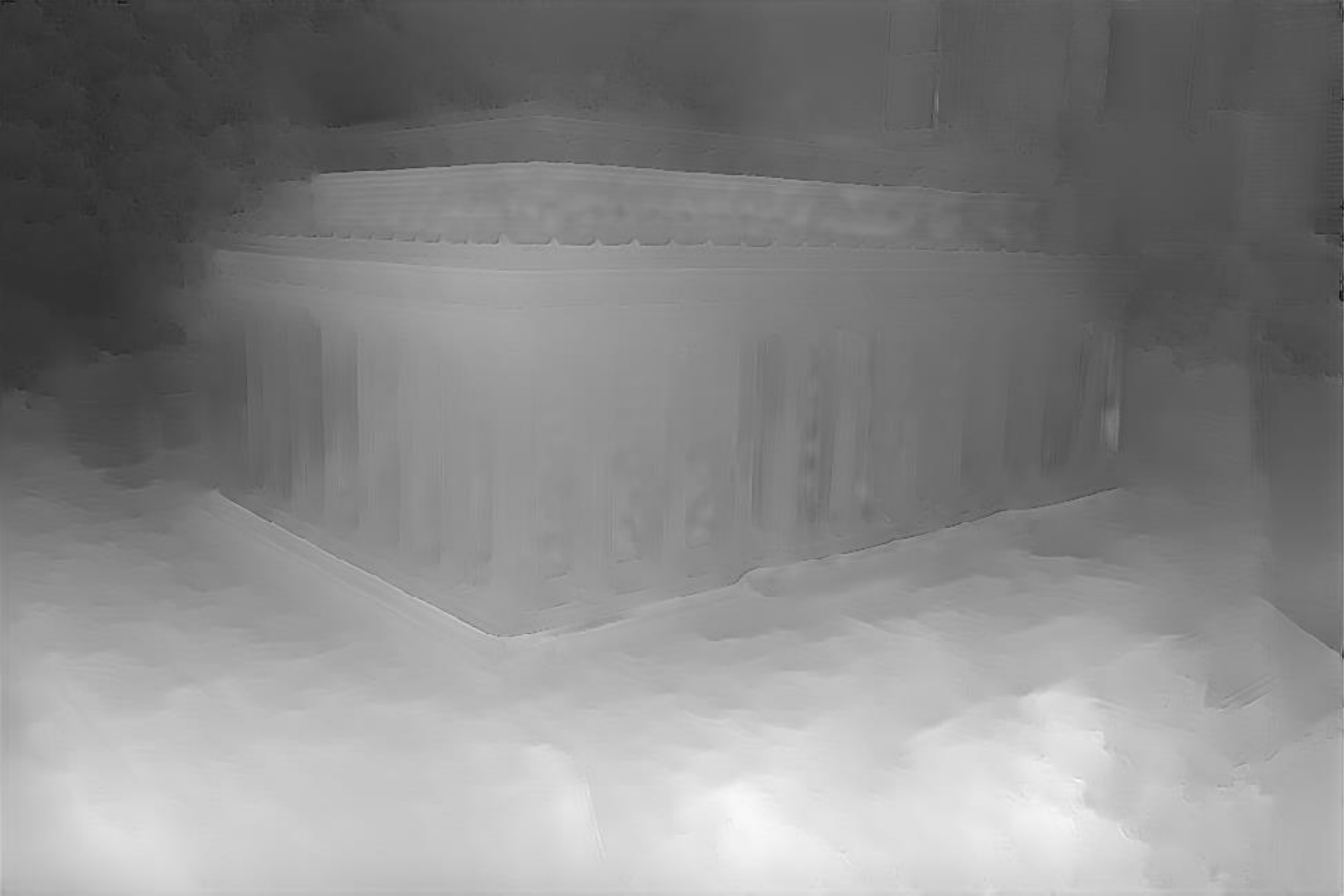}}	
	\end{tabular}
	\caption{Comparison of depth estimation results with state-of-the-art methods captured by Canon 1D Mark \RN{3}. (a) Reference images. (b) Depth maps from propagation~\protect\cite{Im15}. (c) Depth maps from plane sweeping~\protect\cite{Ha16}. (d) Our final depth maps.}
	\label{fig:real}
\end{figure*}

\figref{fig:real} shows the results of real-world datasets captured at night. All the comparative methods produce reasonable results; however, we found that our method achieves more reliable results. The propagation method~\cite{Im15} results in over-smoothing effect in~\figref{real_iccv}, and the plane sweep method~\cite{Ha16} exhibits the speckle artifacts and quantization errors in~\figref{real_cvpr}.
%
Although brightness-adjusted images were used for the competing methods, over or under-saturation regions might exist, which causes severe artifacts. 
Despite intensity changes on images, our results in~\figref{real_final} show an immunity towards the changes, similar to the result of the synthetic datasets in~\secref{sec:synthetic}.

\begin{figure*}[t]
	\centering
	
	\begin{tabular}{c@{\hspace{1mm}}c@{\hspace{1mm}}c@{\hspace{1mm}}c@{\hspace{1mm}}}
		{\includegraphics[height=0.154\linewidth]{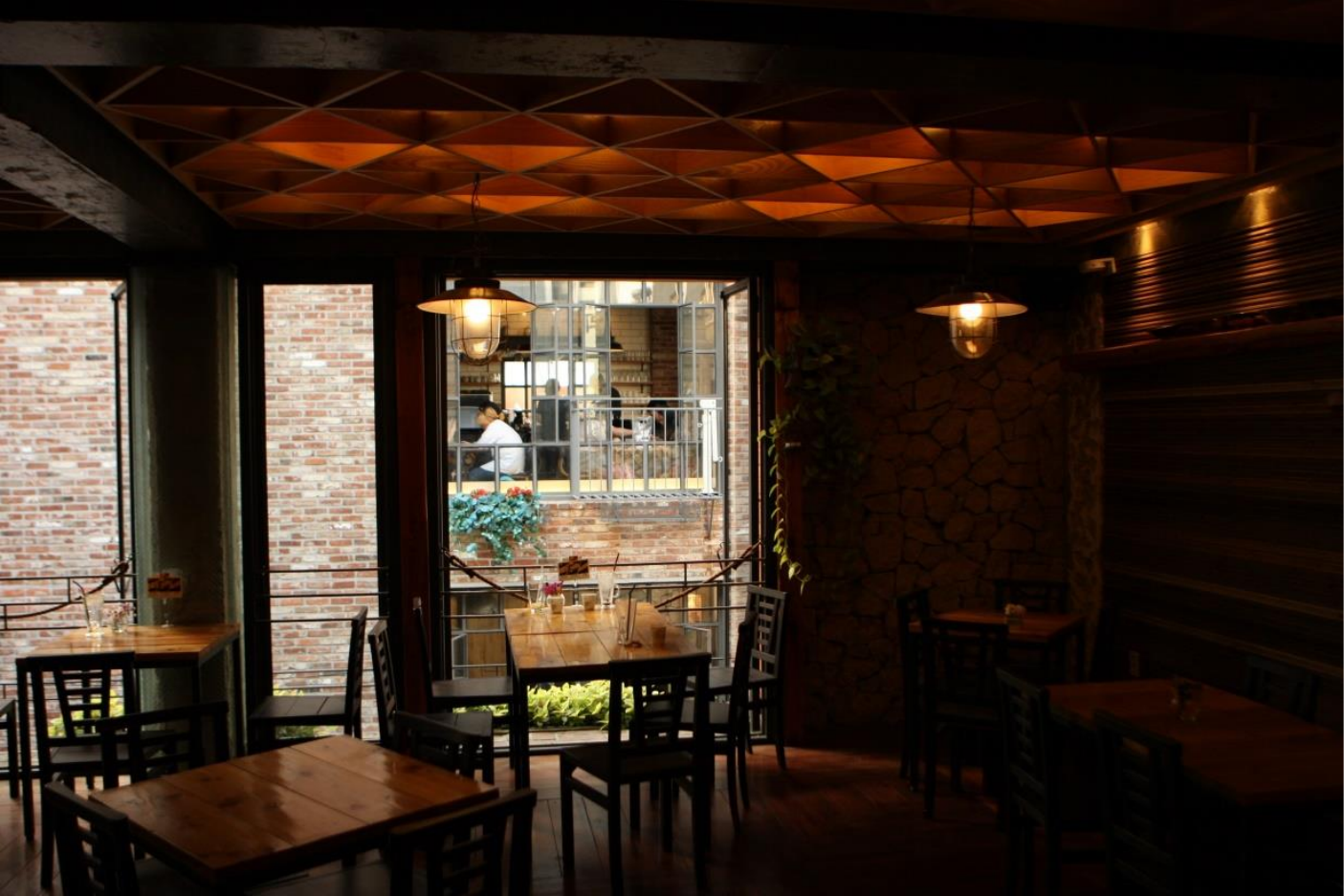}}
		{\includegraphics[height=0.154\linewidth]{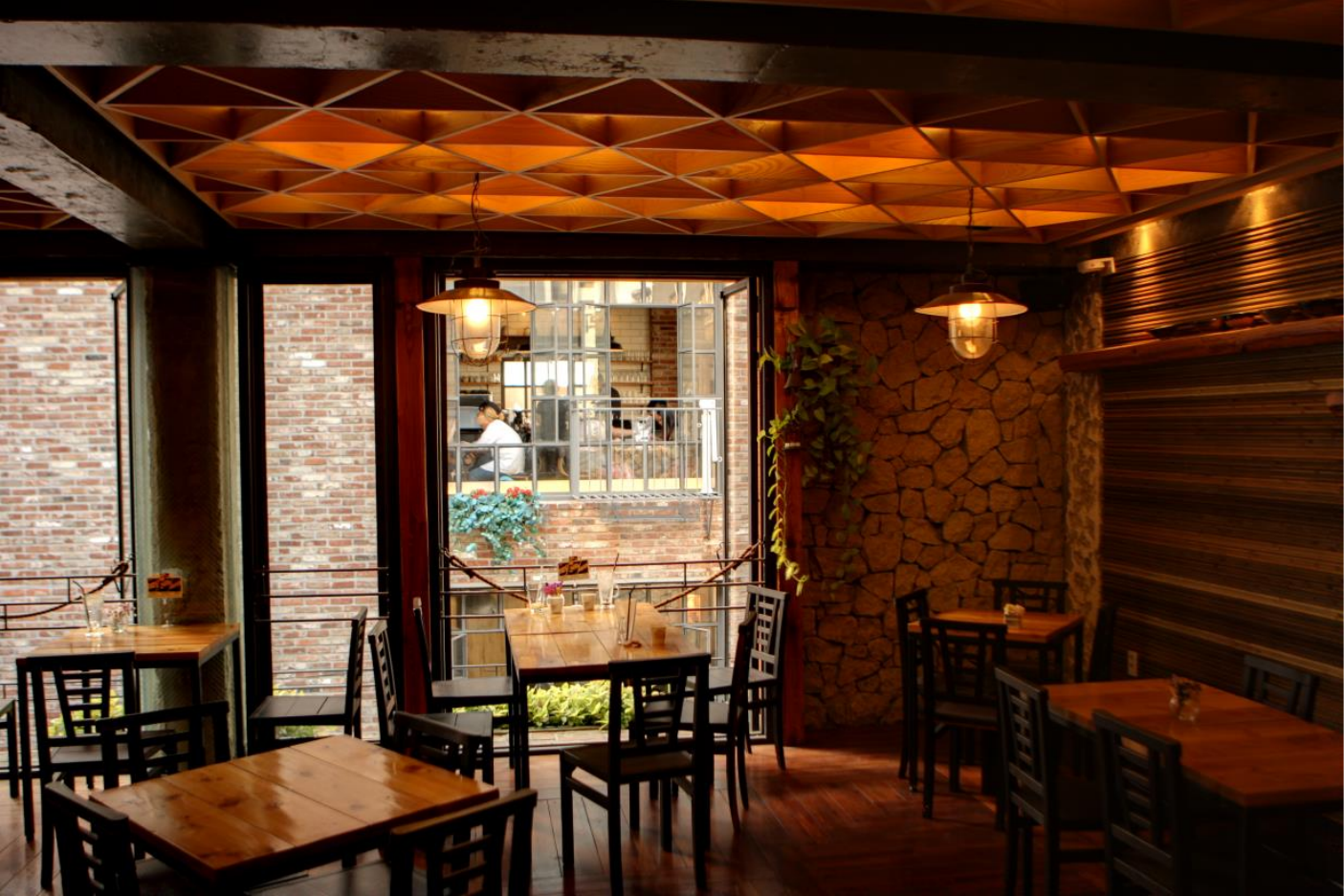}}
		{\includegraphics[height=0.154\linewidth]{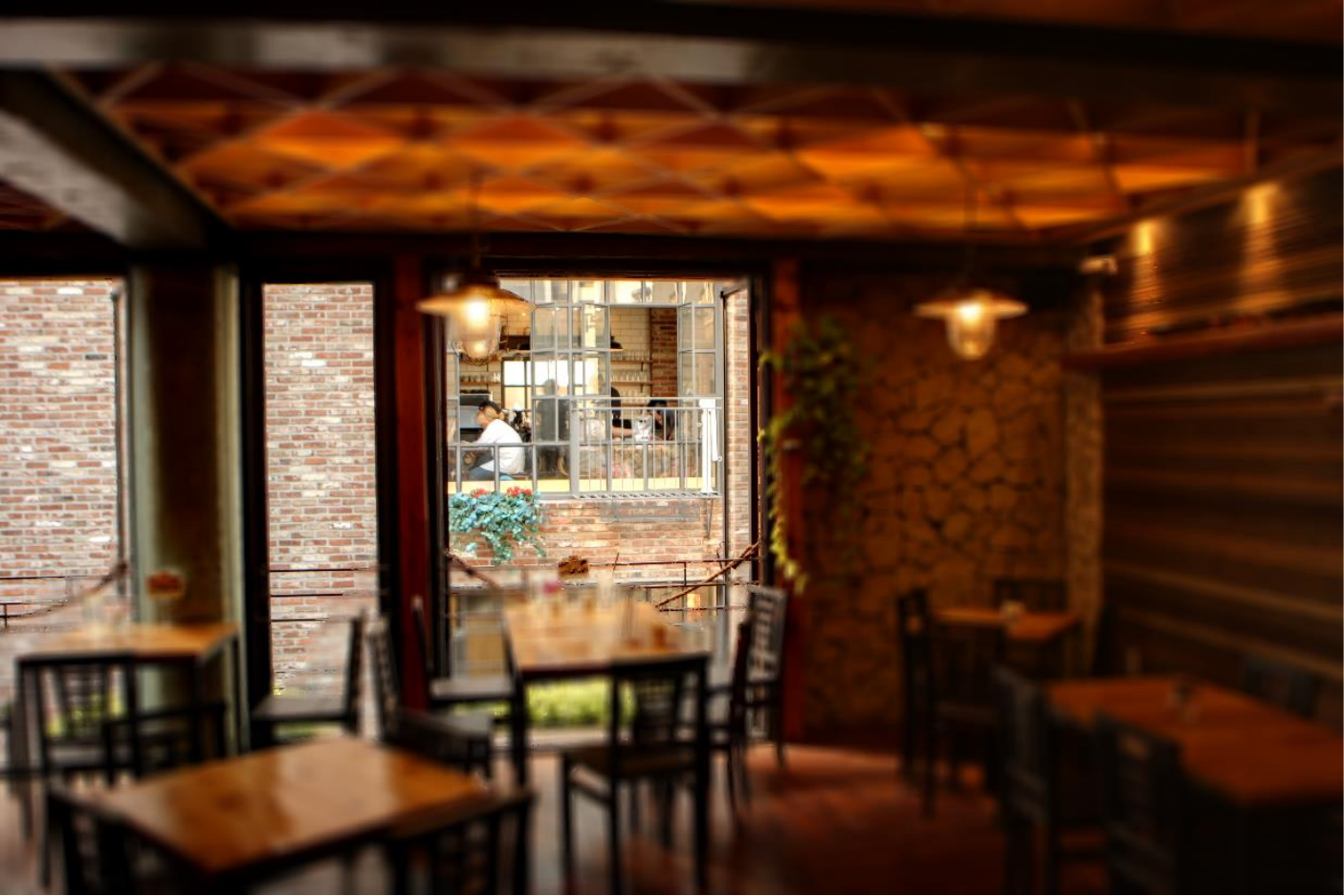}}
		{\includegraphics[height=0.154\linewidth]{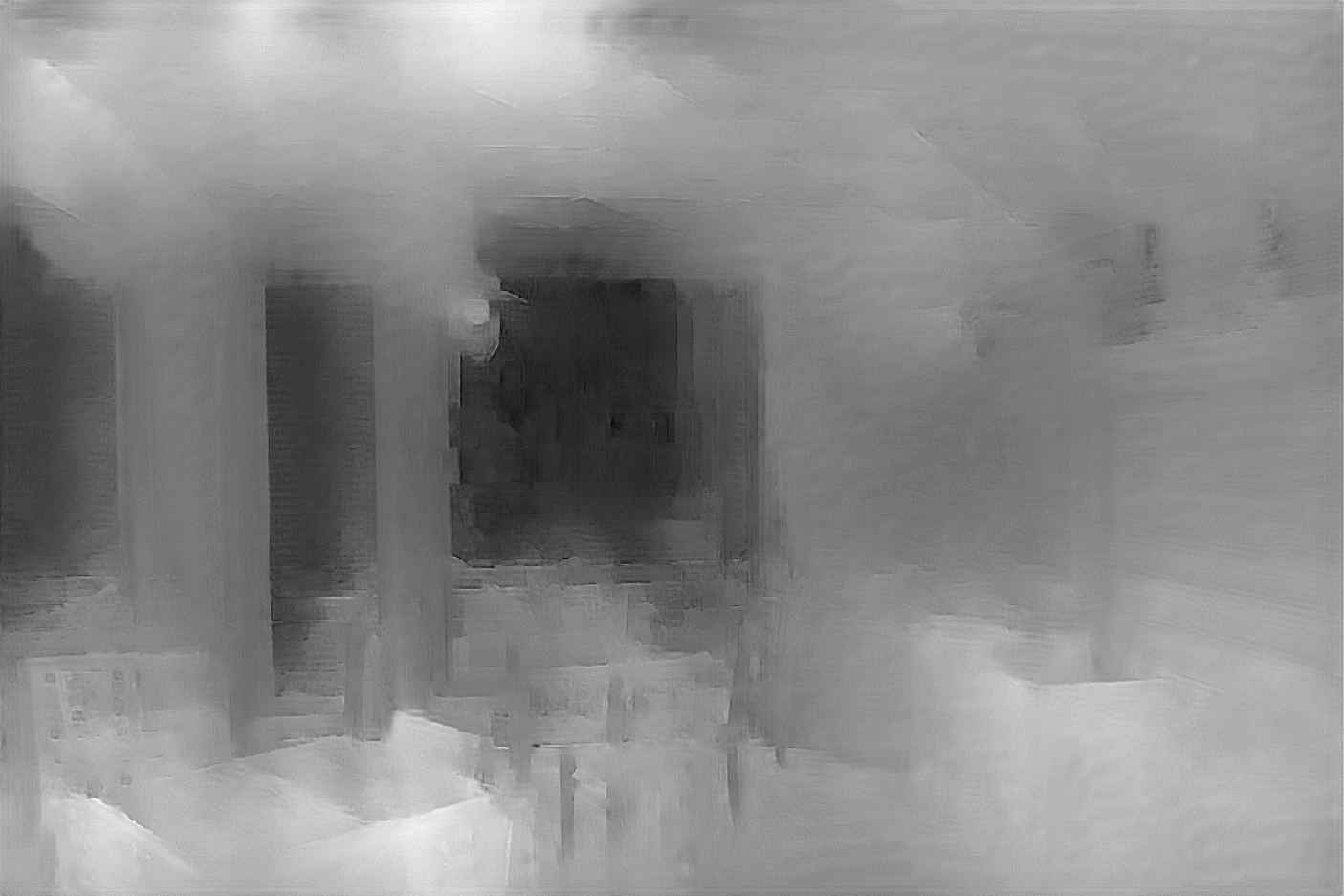}} \\
		\subcaptionbox{\label{final1_input} Reference images}{\includegraphics[height=0.154\linewidth]{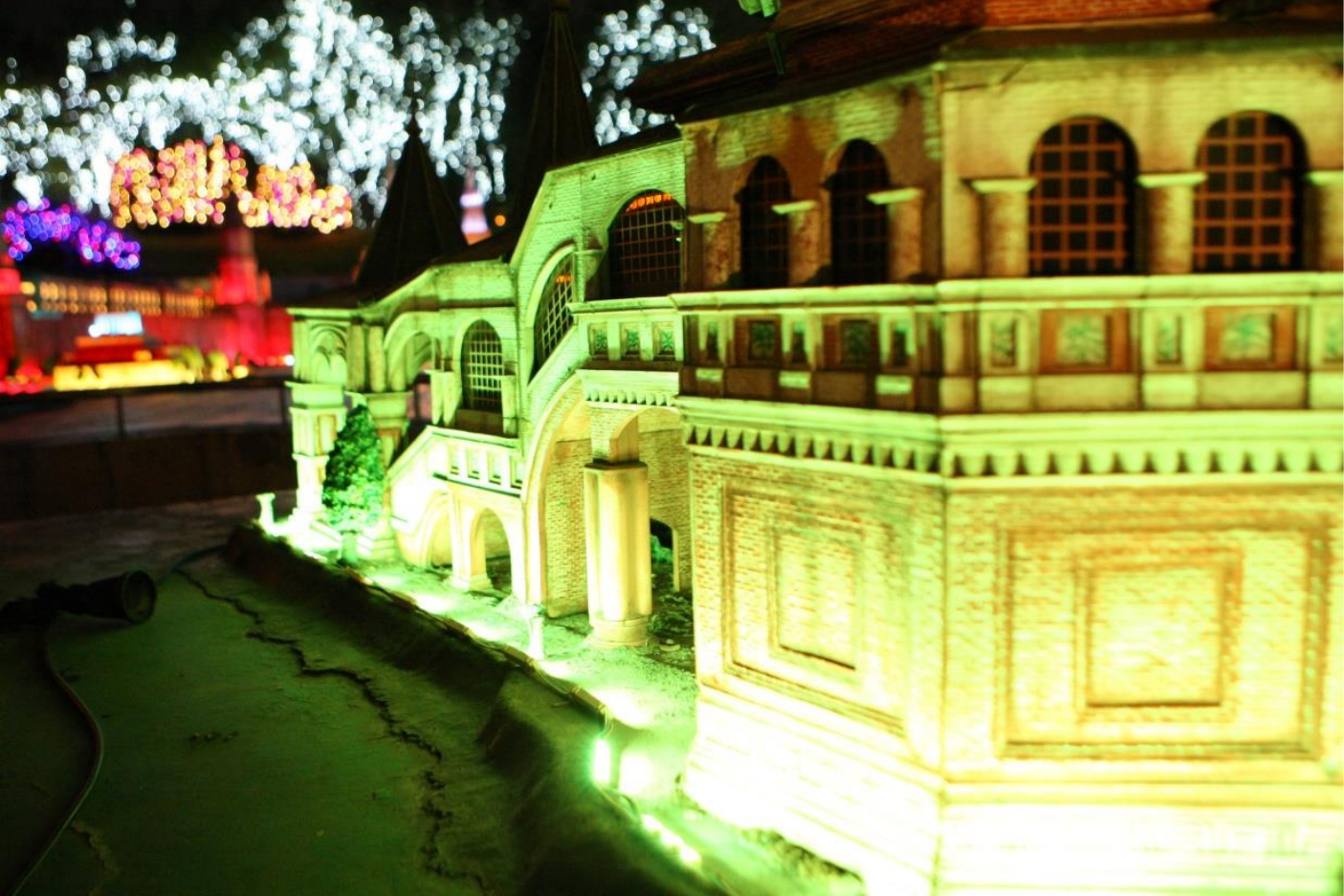}}	
		\subcaptionbox{\label{final1_dn} Exposure fusion }{\includegraphics[height=0.154\linewidth]{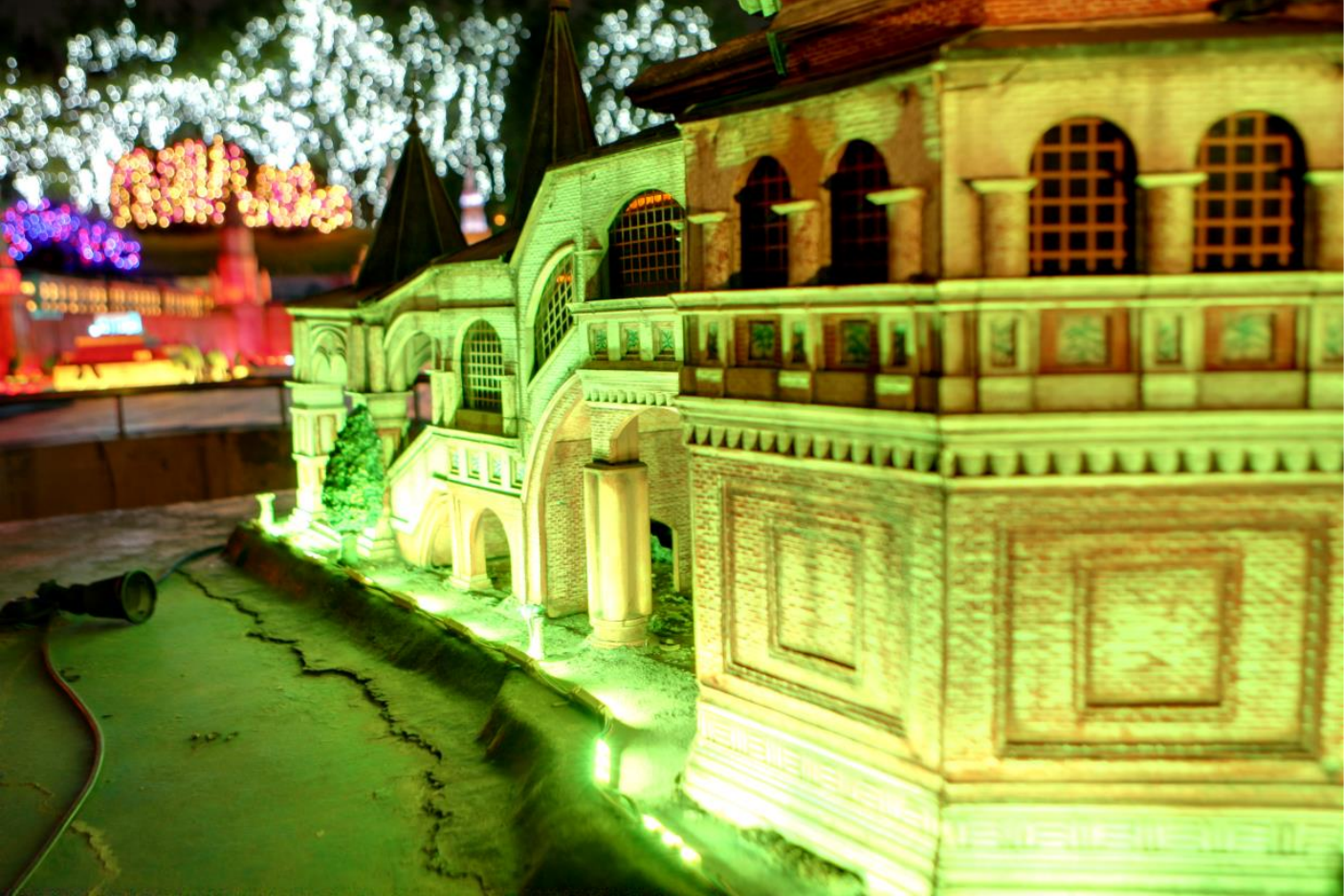}}	
		\subcaptionbox{\label{final1_ef} Photographic editing}{\includegraphics[height=0.154\linewidth]{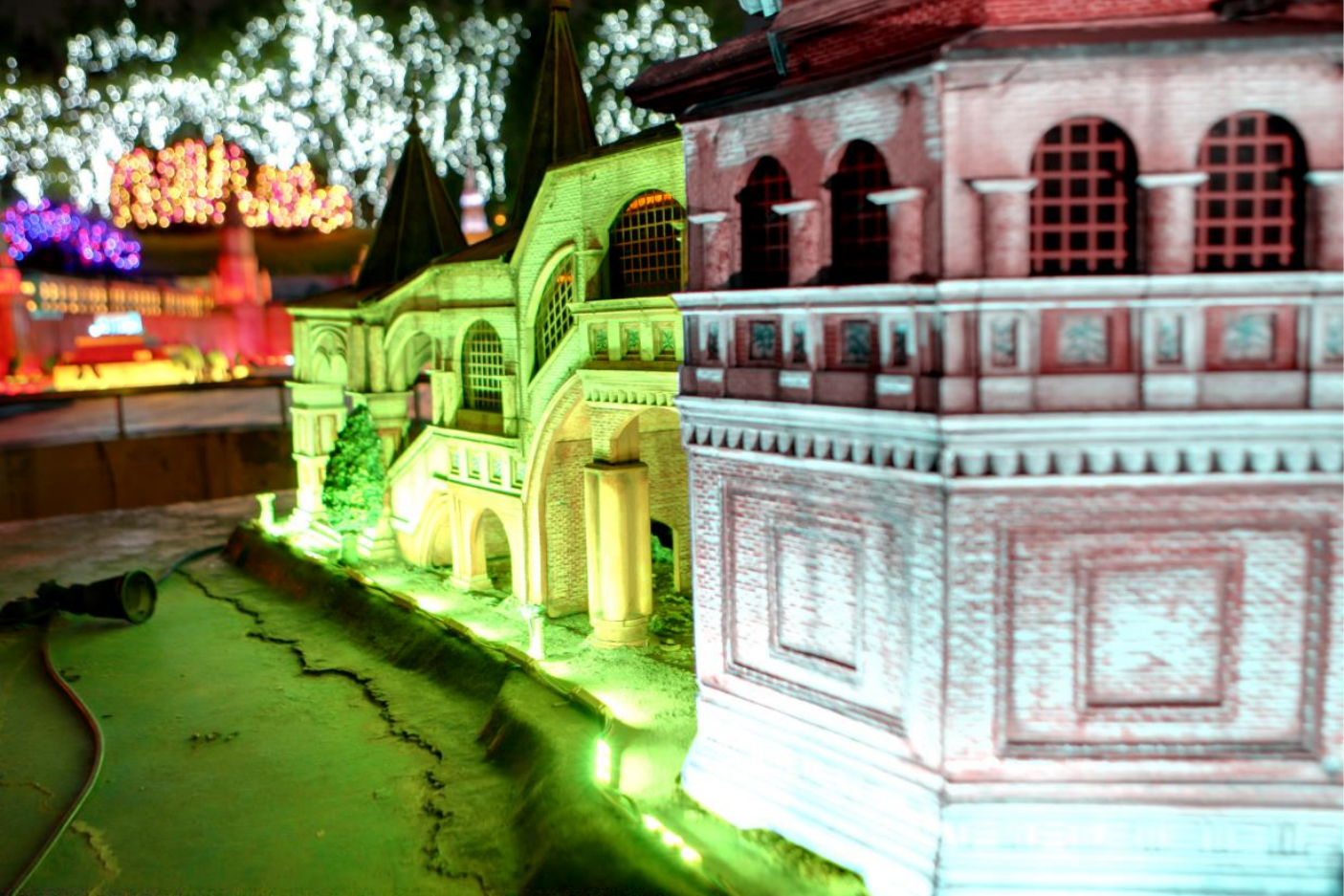}}	
		\subcaptionbox{\label{final1_rf} Our depths}{\includegraphics[height=0.154\linewidth]{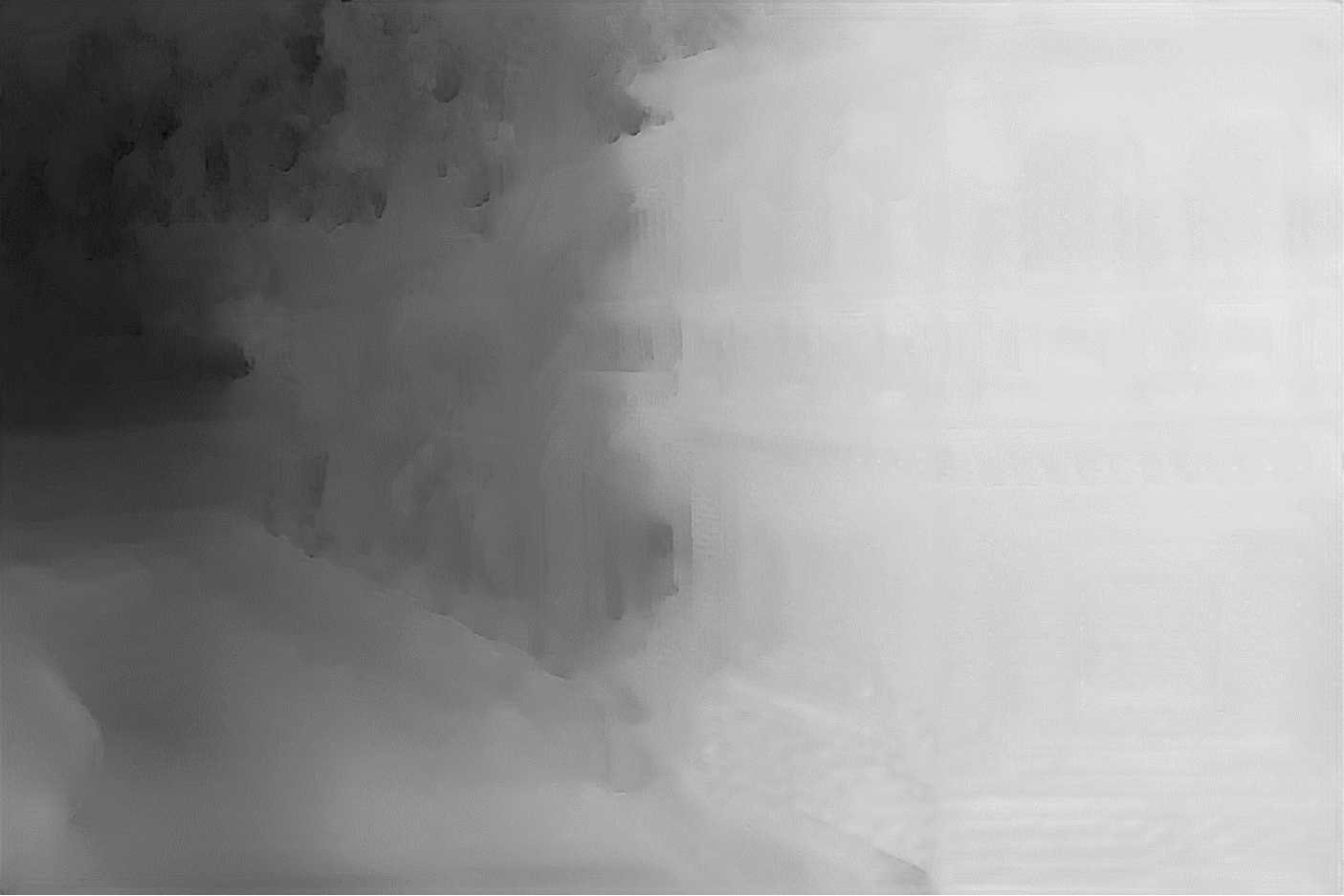}}	
	\end{tabular}
	\caption{Depth-aware photographic editing applications to Synthetic refocusing (top), Image stylization (bottom) and our depths captured by Canon 1D Mark~\RN{3}}
	\label{fig:final1}
	
	\vspace*{\floatsep}	
	
	\centering
	\begin{tabular}{c@{\hspace{1mm}}c@{\hspace{1mm}}c@{\hspace{1mm}}c@{\hspace{1mm}}}
		{\includegraphics[height=0.34\linewidth]{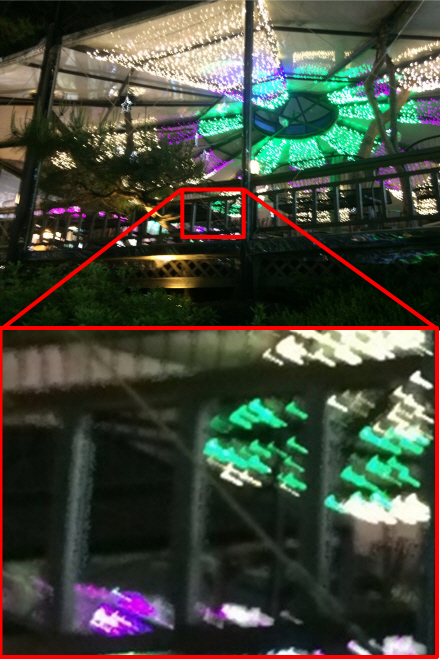}}
		{\includegraphics[height=0.34\linewidth]{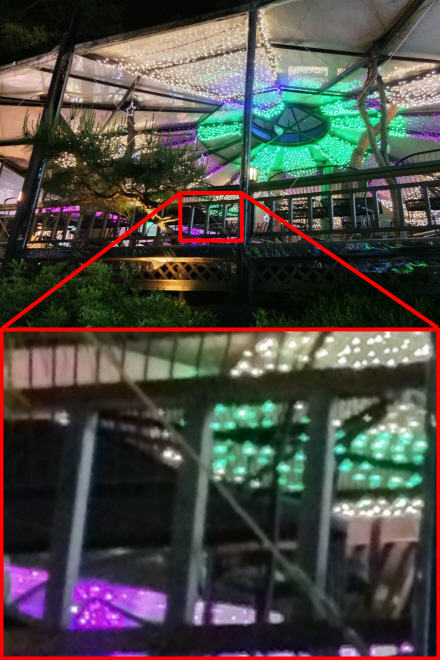}}
		{\includegraphics[height=0.34\linewidth]{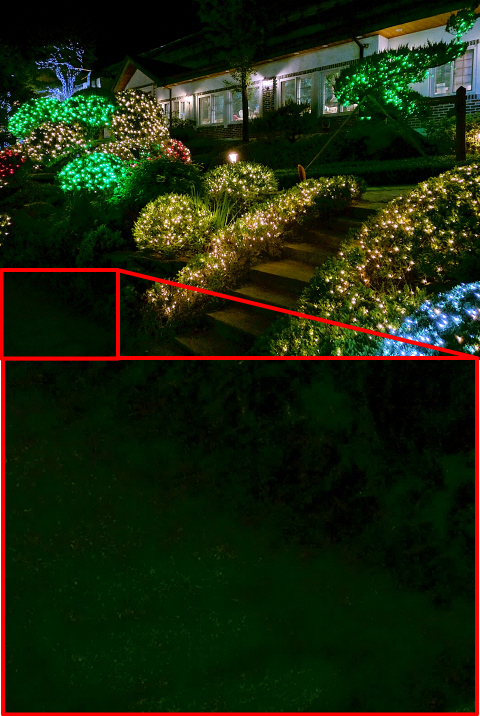}}
		{\includegraphics[height=0.34\linewidth]{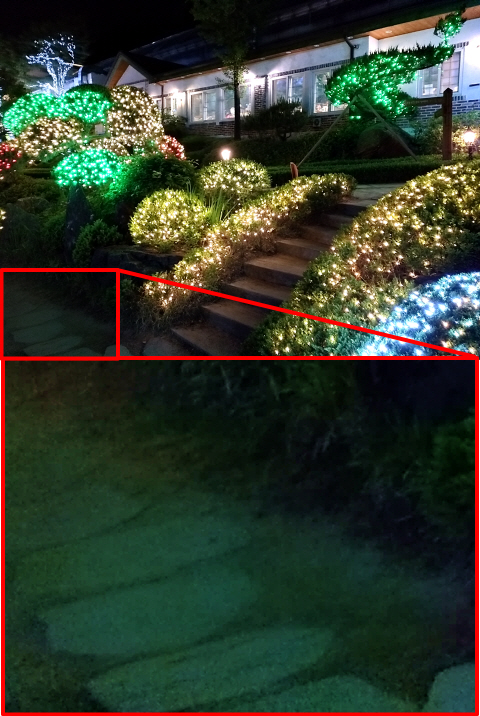}}	\\
		\subcaptionbox{\label{iphone2} Microsoft selfie (iPhone)}{\includegraphics[height=0.34\linewidth]{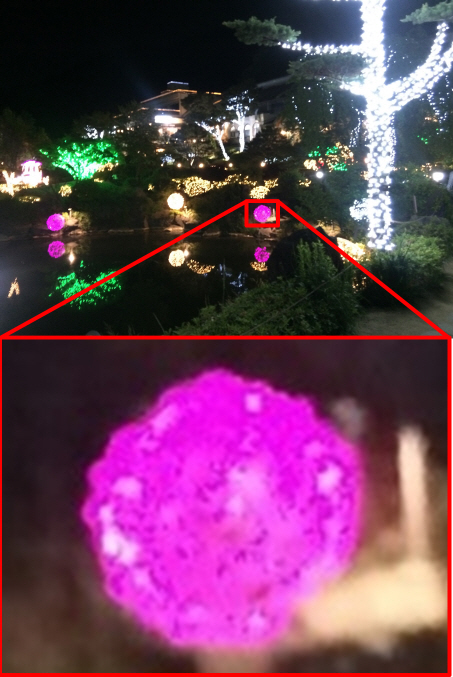}}
		\subcaptionbox{\label{iphone1} Ours (iPhone)}{\includegraphics[height=0.34\linewidth]{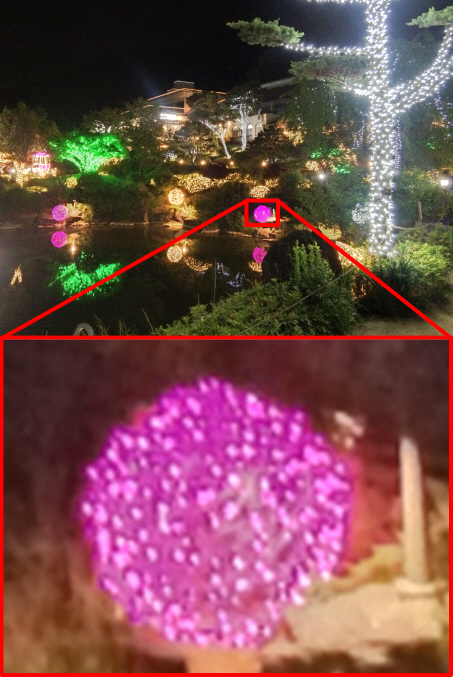}}
		\subcaptionbox{\label{nexus2} Google camera  (Nexus)}{\includegraphics[height=0.34\linewidth]{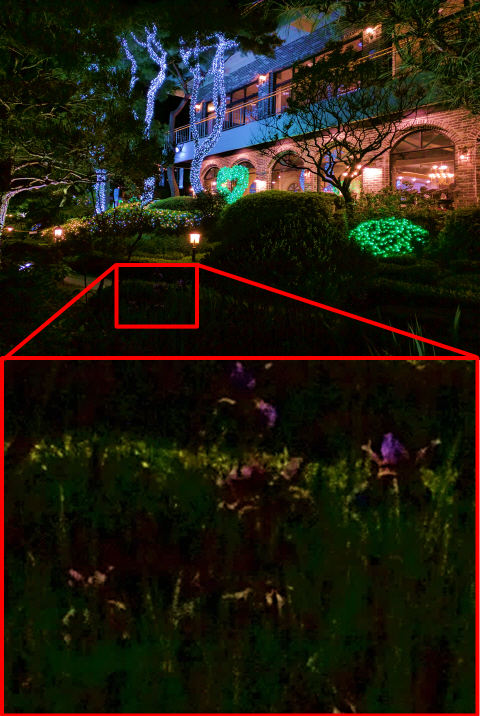}}
		\subcaptionbox{\label{nexus1} Ours (Nexus)}{\includegraphics[height=0.34\linewidth]{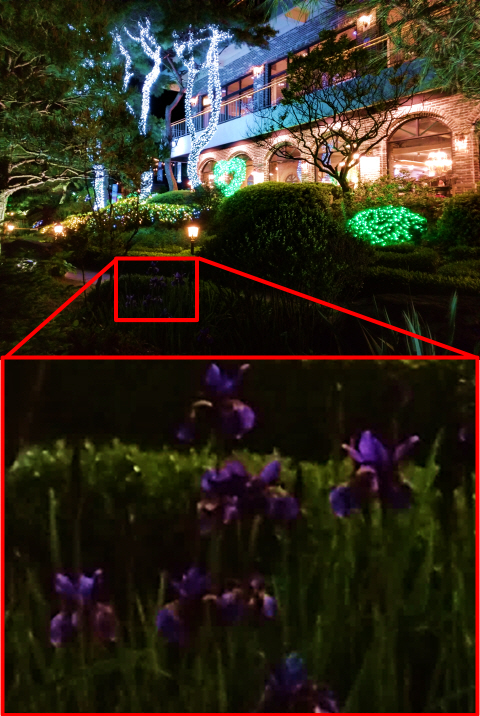}}
	\end{tabular}
	\caption{Qualitative comparison with the state-of-the-art methods~\protect\cite{Liu14,Hasinoff16}. (a) Burst images Denoising results from Microsoft selfie app~\protect\cite{Liu14}. (b) Our noise-free exposure fusion results. (c) HDR+ results from Google camera app~\protect\cite{Hasinoff16}.  (d) Our noise-free exposure fusion results. (a), (b) are captured by an iPhone 5S and (c), (d) are captured by a Nexus 6.}
	\label{fig:iphone}
\end{figure*}

We also found that our accurate depth can be additionally useful for exposure fusion and depth-aware photographic editing applications, such as digital refocusing and image stylization in~\figref{fig:final1}. 
Exposure fusion assembles the multi-exposure sequence into a high quality image using a weighted blending of the input images~\cite{mertens2009exposure}. To obtain a desirable result, the set of images should be well-aligned. The final results in~\figref{final1_dn} demonstrated that our depth can accurately align the set of images.
Digital refocusing, which shifts the in-focus region after taking a photo~\cite{HTC,Dell}, is one of the most popular depth-aware applications. 
For a realistic refocused image, accurate depth information is necessary. 
We added synthetic blurs to the images and produced a shallow depth of field image using our depth in~\figref{final1_ef} (top).
Another interesting application is image stylization, which photographically changes an image at a certain depth range in~\figref{final1_ef} (bottom).
These results demonstrate that our depth is enough to be utilized on real-world images for various photographic applications.

\noindent{\bf{Comparison to state-of-the-art burst photography}} \quad 
Finally, we compared the proposed method to state-of-the-art burst image photographic approaches; Burst Image Denoising~\cite{Liu14} and HDR+~\cite{Hasinoff16}. 
\textit{Microsoft selfie app} and \textit{Google camera app} pioneered the use of Burst Image Denoising and HDR+ on \textit{iOS8} and \textit{Android}, respectively. 
We took the image sequences from each phone to use them as the input of our algorithm, and compared them with Burst Image Denoising and HDR+. We obtained independent results from an iPhone5S and Nexus6, as shown in~\figref{fig:iphone}.

Burst Image Denoising~\cite{Liu14} aligns the input image sequences using the local homography, then merges them with the weighted average. 
The denoising results in~\figref{iphone2} shows that Burst Image Denoising outputs blurred results, while our method preserves image boundaries and fine detail in~\figref{iphone1}. (Note that the blurred frame is not our selection, but is the result of image alignment~\cite{Liu14}.)
The local homography might sometimes fail to handle the user's inevitable motion during burst mode.

The HDR+~\cite{Hasinoff16} generates synthetic exposures by applying gain and gamma corrections to multiple images using a constant exposure, then fuses the synthetic images as if they had been captured using bracketing.
Although HDR+ shows promising results in well exposed areas, the constant exposure does not help to recover some badly exposed areas due to lack of light, as shown in~\figref{nexus2}.
On the other hand, exposure fusion with real bracketing can cover all of the areas of the input image, as shown in~\figref{nexus1}. 
Our depth estimation method enables the fusing of bracketed images with exposures that are not aligned, and results in brighter images than the original images.

\section{Discussion}
We have presented a robust narrow-baseline multi-view stereo matching method for noise or intensity changes.
We determined an important clue that the baseline of the inevitable motion can be used for depth estimation, and the depth enables accurate image alignment leading to image quality enhancement. 
Both depth and image enhancement results were compared against state-of-the-art methods with a variety of datasets, and demonstrated considerable improvement over existing methods. 

The main advantage of our method is its fast computational time and small size network, which are important features for implementation in a mobile platform. 
Compared to state-of-the-art DfSM methods~\cite{Im15,Ha16}, which take about a few minute, our method takes only a few second.
Our DNN plays a key role in reducing computational complexity in dense matching which is the most time-consuming part of conventional DfSM.
In addition, our network is much lighter than the DNN-based fast depth or optical flow estimation methods~\cite{Zbontar15,Mayer16,Luo16,dosovitskiy2015flownet} (Flownet: 32M vs Ours: 240K). 
This significant reduction without performance degradation is achieved by training the residual flow, and iteratively updating optical flow.		
We expect that the proposed framework will become popular as a mobile phone application.

On the other hand, there are still rooms for improvements:
1) when there is large camera rotation, inaccurate camera poses might be obtained, which can cause an error in our DMVS;
2) our method requires the pre-calibrated intrinsic parameters to estimate the camera poses;
3) the performance of our method is not guaranteed for datasets with fast moving objects, since the scene flow contains additional flow on the object;
4) various fields, such as AR/VR, require metric scale depth, but the estimated depth is not represented in the metric scale.

As future works, we have plan to address these issues.
In particular, an idea of the uncalibrated DfSM in~\cite{Ha16} is expected to provide a solution to the calibration issue.
The scale problem can also be addressed if we directly measure the camera motion during taking photos by introducing additional hardware such as inertial sensors.

\small{\noindent{\bf{Acknowledgement}}\quad This work was supported by the Technology Innovation Program (No. 2017-10069072) funded By the Ministry of Trade, Industry \& Energy (MOTIE, Korea). Sunghoon Im was partially supported by Global Ph.D. Fellowship Program through the National Research Foundation of Korea (NRF) funded by the Ministry of Education (NRF-2016907531).}

{\small
	\bibliographystyle{ieee}
	\bibliography{egbib}
}

\end{document}